\documentclass[runningheads]{llncs}

\usepackage[mobile]{eccv}

\usepackage{eccvabbrv}

\usepackage{graphicx}
\usepackage{booktabs}
\usepackage{amsmath,amssymb}
\usepackage{subcaption}
\usepackage{algorithm}
\usepackage{algorithmic}
\usepackage{subcaption}
\usepackage{multirow}
\usepackage{enumitem}
\usepackage{inconsolata}

\usepackage[accsupp]{axessibility}  % Improves PDF readability for those with disabilities.

\usepackage[pagebackref,breaklinks,colorlinks,citecolor=eccvblue]{hyperref}
\usepackage{orcidlink}

\begin{document}

\title{Fragile Reconstruction: Adversarial Vulnerability of Reconstruction-Based Detectors for Diffusion-Generated Images\vspace{-2mm}} 



\author{Haoyang Jiang\inst{1}, 
Mingyang Yi\inst{1}\textsuperscript{,$\dagger$}, 
Shaolei Zhang\inst{1}, 
Junxian Cai\inst{2}, 
Qingbin Liu\inst{2}, 
Xi Chen\inst{2}, 
Ju Fan\inst{1}}

\authorrunning{H.~Jiang et al.}

\institute{
\textsuperscript{1}Renmin University of China, Beijing, China \\
\texttt{\{jianghaoyang233, yimingyang, zhangshaolei98, fanj\}@ruc.edu.cn} \\
\vspace{2mm} % 强制空开一点，层次分明
\textsuperscript{2}Tencent Inc., Shenzhen, China \\
\texttt{\{jasoncjxcai, qingbinliu, jasonxchen\}@tencent.com} \\
\vspace{2mm}
\textsuperscript{$\dagger$} {\scriptsize Corresponding author.}
}

\titlerunning{Fragile Reconstruction}

\maketitle

\begin{figure}[ht]
\vspace{-8mm}
\centering
\includegraphics[width=0.9\linewidth]{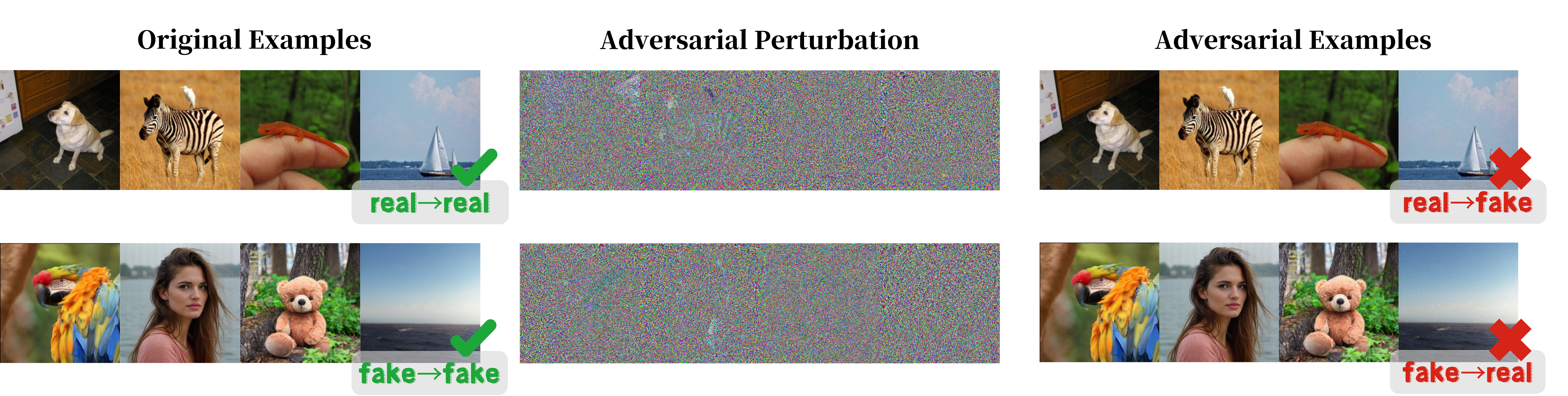}
\captionsetup{width=0.9\linewidth}
\caption{Adversarial attack on reconstruction-based detectors. \textbf{Left}: The detector correctly classifies real images (top row) and generated images (bottom row). \textbf{Middle}: Imperceptible adversarial perturbations ($\times 15$ magnified). \textbf{Right}: Adversarial samples obtained by adding the perturbations, the detector's predictions are completely reversed.}
\vspace{-10mm}
\label{fig:teaser}
\end{figure}

\begin{abstract}
Recently, detecting AI-generated images produced by diffusion-based models has attracted increasing attention due to their potential threat to safety. Among existing approaches, reconstruction-based methods have emerged as a prominent paradigm for this task. However, we find that such methods exhibit severe security vulnerabilities to adversarial perturbations; that is, by adding imperceptible adversarial perturbations to input images, the detection accuracy of classifiers collapses to near zero. To verify this threat, we present a systematic evaluation of the adversarial robustness of three representative detectors across four diverse generative backbone models. First, we construct adversarial attacks in white-box scenarios, which degrade the performance of all well-trained detectors. Moreover, we find that these attacks demonstrate transferability; specifically, attacks crafted against one detector can be transferred to others, indicating that adversarial attacks on detectors can also be constructed in a black-box setting. Finally, we assess common countermeasures and find that standard defense methods against adversarial attacks provide limited mitigation. We attribute these failures to the low signal-to-noise ratio (SNR) of attacked samples as perceived by the detectors. Overall, our results reveal fundamental security limitations of reconstruction-based detectors and highlight the need to rethink existing detection strategies. The code is available at \url{https://github.com/atrijhy/Fragile-Reconstruction}.
%: diffusion purification fails due to its structural similarity to the attack and the erasure of discriminative artifacts, while adversarial training is hindered by the intrinsically low signal-to-noise ratio (SNR) of reconstruction residuals
%By applying To enable end-to-end white-box attacks through diffusion inversion, we adopt the Adjoint Sensitivity Method to compute gradients with constant memory. Leveraging this capability, we demonstrate that these detectors exhibit catastrophic vulnerability: under imperceptible perturbations, detection accuracy collapses to near zero. Beyond white-box settings, we also observe strong black-box transferability: adversarial examples crafted on one generator or detector generalize to unseen generators and detection architectures. We then assess common countermeasures and find that standard defenses offer limited mitigation: diffusion purification fails due to its structural similarity to the attack and the erasure of discriminative artifacts, while adversarial training is hindered by the intrinsically low signal-to-noise ratio (SNR) of reconstruction residuals. Overall, our results reveal fundamental security limitations of reconstruction-based detectors and highlight the need for more robust detection strategies.

\end{abstract}

\section{Introduction}
\label{sec:intro}
%=====================================================================

The rapid development of diffusion models~\cite{ho2020denoising} has completely reshaped visual content generation, ranging from images~\cite{dhariwal2021diffusion,gu2022vector,rombach2022high,blackforest2024flux} to videos~\cite{ho2022video,ma2024latte,brooks2024video}. As these models evolve, synthesized content has achieved a level of realism in texture and semantic consistency that is virtually indistinguishable from real data. However, this remarkable capability also poses severe security threats, as photorealistic fakes can be easily generated to spread misinformation~\cite{mirsky2021creation,vaccari2020deepfakes} or infringe copyrights~\cite{somepalli2023diffusion}, necessitating reliable detection methods~\cite{wang2020cnn,corvi2023detection}.

To address these threats, various detectors have been developed to identify diffusion-generated images~\cite{wang2020cnn,corvi2023detection}. Among these, reconstruction-based methods~\cite{wang2023dire,luo2024lare,ricker2024aeroblade} stand out for their high accuracy and strong transferability. Their core mechanism exploits the fact that after an ``inversion-reconstruction'' cycle (e.g., DDIM inversion~\cite{song2020denoising}), synthesized images exhibit distinctly different reconstruction errors from real ones, which can serve as discriminative features for classification.

However, we demonstrate that while performing reliably on standard benchmarks, these classifiers are highly vulnerable to adversarial perturbations ~\cite{szegedy2013intriguing,goodfellow2014explaining}. As shown in Fig. \ref{fig:teaser}, adding tiny, imperceptible perturbations can induce a drastic shift during the diffusion reconstruction process, ultimately rendering the detector ineffective. This exposes a critical issue: \textbf{detectors designed to safeguard security are themselves susceptible to security breaches.}

Concretely, by efficiently crafting adversarial examples via an ODE solver, we comprehensively investigate the adversarial vulnerability and transferability of reconstruction-based detectors across various methods and generative backbones, revealing that:
\begin{enumerate}
\item \textbf{Attack Effectiveness:} The proposed adversarial attacks effectively dismantle reconstruction-based detection methods in white-box scenarios. Concretely, under APGD~\cite{croce2020reliable} with an $\ell_\infty$ budget of $\varepsilon=8/255$, the robust accuracy of all explored detectors plummets to nearly 0\%.

\item \textbf{Attack Transferability:} The attacks exhibit strong transferability across both generative models and detection methods. Since a detector's behavior is determined by its training data and methodology, we find our adversarial perturbations successfully transfer across different training datasets and detection frameworks, demonstrating the severe threat of black-box attacks.

\item \textbf{Defense Vulnerability:} Unfortunately, standard defense mechanisms, namely adversarial training \cite{madry2017towards} and diffusion-based purification \cite{nie2022diffusion}, fail to mitigate our attacks. Crucially, we identify Signal-to-Noise Ratio (SNR) collapse as the root cause: the reconstruction residuals are structurally too weak to withstand perturbations. 
\end{enumerate}

In summary, our investigation reveals a critical paradox: while reconstruction-based detectors excel on benign data, they collapse to near-random accuracy under adversarial pressure. This vulnerability is intrinsic, highly transferable, and renders standard defenses ineffective. These findings challenge the viability of reconstruction error as a sole metric, necessitating a rethinking of detector robustness for real-world deployment.

%=====================================================================
\section{Related Work}
\label{sec:related}
%=====================================================================

\paragraph{Diffusion Models.}
The emergence of diffusion models~\cite{ho2020denoising,song2020score} has transformed the landscape of image generation. Building upon the iterative denoising foundation laid by DDPMs~\cite{ho2020denoising,wang2025improved,yi2023generalization,yi2024towards}, subsequent milestones rapidly expanded their capabilities. Specifically, DDIM~\cite{song2020denoising} accelerated sampling via non-Markovian processes, while ADM~\cite{dhariwal2021diffusion} first achieved generation quality surpassing GANs on ImageNet~\cite{deng2009imagenet}. To further improve efficiency, Latent Diffusion Models (LDMs)~\cite{rombach2022high} shifted the process into a compressed latent space. More recently, transformer-based architectures like DiT~\cite{peebles2023scalable} and flow-matching frameworks like FLUX~\cite{blackforest2024flux} have pushed the boundaries of photorealism. These remarkable advances, however, significantly amplify the risks of malicious misuse~\cite{mirsky2021creation,vaccari2020deepfakes,somepalli2023diffusion}, making robust detection methods increasingly urgent~\cite{wang2020cnn,corvi2023detection}.

\paragraph{Reconstruction-Based Detection.}
DIRE~\cite{wang2023dire} first establishes this paradigm by observing that diffusion-generated images exhibit lower reconstruction errors than real ones. SeDID~\cite{ma2023exposing} further explores pixel-space reconstruction by accumulating errors across diffusion trajectories. To improve efficiency, LaRE$^2$~\cite{luo2024lare} and LATTE~\cite{vasilcoiu2025latte} shift reconstruction to the latent space, while AEROBLADE~\cite{ricker2024aeroblade} bypasses diffusion entirely via VAE reconstruction error, enabling training-free deployment. Other methods incorporate complementary signals such as frequency-domain features (FIRE~\cite{chu2025fire}), semantic information (SARE~\cite{kang2025semantic}), and color-channel priors (GRRE~\cite{he2026grre}). Despite their diversity, all share a common principle: measuring discrepancies between inputs and their reconstructions.
% More importantly, the adversarial security of these methods has yet to be fully explored. 

\paragraph{Adversarial Attacks and Defenses.}
Adversarial examples~\cite{szegedy2013intriguing,goodfellow2014explaining,yi2021reweighting,yi2021improved} are crafted by adding imperceptible perturbations to fool neural networks. To defend against such attacks, adversarial training~\cite{madry2017towards} trains models on worst-case perturbations, later improved by~\cite{zhang2019theoretically,wang2019improving}. As an alternative, diffusion-based purification~\cite{nie2022diffusion} removes adversarial noise via a ``noising then reconstruction'' process before classification. However, while these attack and defense mechanisms have been extensively studied for standard classification~\cite{carlini2017towards,croce2020reliable}, the adversarial robustness of AIGC detectors themselves remains largely unexplored.
% and frequency-based forensic methods~\cite{}, reconstruction-based detectors remain unexplored under adversarial settings.
%=====================================================================
\section{Background}
\label{sec:background}
%=====================================================================
\subsection{Continuous-time Diffusion Model}

Diffusion models can be formulated within a continuous-time framework~\cite{song2020score}. The forward process gradually transforms the data distribution $x_0$ into Gaussian noise $x_T$ via a stochastic differential equation (SDE):
\begin{equation}\label{eq:sde}
dx_{t} = f(x_{t},t)dt + g(t)dw_{t},
\end{equation}
where $f(\cdot,t)$ is the properly chosen drift coefficient (e.g., variance preserving or variance exploding \cite{song2020score}), $g(t)$ is the diffusion coefficient, and $w_{t}$ denotes a standard Wiener process. 

Song et al.~\cite{song2020score} demonstrate that by solving the following reverse-time deterministic probability-flow ODE:
\begin{equation}\label{eq:ode}
\frac{dx_{t}}{dt} = f(x_{t}, t) - \tfrac{1}{2}g(t)^2\nabla_x\log p_t(x),
\end{equation}
starting from a standard Gaussian, the obtained $x_{0}$ follows the original data distribution. Here, the score function $\nabla_x\log p_t(x)$ is parameterized by a learned network $\epsilon_\theta$ satisfying $\epsilon_{\theta}(x) \approx -\sqrt{1 - \bar{\alpha}_{t}}\nabla_{x}\log{p_{t}}(x)$, where $\bar{\alpha}_{t}$ is a constant depending on $f$ and $g$. This deterministic formulation naturally aligns with DDIM~\cite{song2020denoising} sampling, which enables an exact inversion process \cite{mokady2023null,hertz2022prompt}: given an input $x_0$, one can reverse the ODE \eqref{eq:ode} to obtain a latent representation $x_T$, and subsequently integrate the ODE \eqref{eq:ode} forward to reconstruct $\hat{x}_0$. Importantly, this invertible property is the foundation of reconstruction-based detection methods.

\subsection{Adjoint Method}
\label{subsec:adjoint_method}

The adjoint method, originating from Neural ODEs~\cite{chen2018neural}, serves as a critical tool for computing gradients efficiently. Concretely, for the ODE in \eqref{eq:ode}, given an objective loss function $\mathcal{L}(x_{t})$, our goal is to compute the gradient $\nabla_{x_0}\mathcal{L}(x_{0})$. Naive backpropagation through the ODE solver requires storing all intermediate states, incurring an $\mathcal{O}(N)$ memory footprint for $N$ integration steps.

The adjoint method overcomes this bottleneck by introducing an adjoint state $a_{t} = \frac{\partial \mathcal{L}(x_{t})}{\partial x_{t}}$, which satisfies the following backward ODE:
\begin{equation}
\frac{da_{t}}{dt} = -a_{t}^\top \nabla_{x}\left(f(x_{t}, t) - \tfrac{1}{2}g(t)^2\nabla_x\log p_t(x_{t})\right).
\end{equation}
By integrating this adjoint ODE backward from $t=T$ to $t=0$ using numerical ODE solvers \cite{chen2018neural} to get $a_{0} = \nabla_{x_0}\mathcal{L}(x_{0})$, the gradient can be evaluated with $\mathcal{O}(1)$ memory complexity—entirely independent of the number of forward integration steps. As we will elaborate later, such an adjoint method is indispensable for crafting adversarial examples against generative detectors, owing to its remarkable memory efficiency in computing the input gradient $\nabla_{x_0}\mathcal{L}(x_{0})$.
% This constant-memory property is crucial for our attack: it enables efficient gradient-based optimization through the entire DDIM inversion-reconstruction pipeline, which would otherwise be memory-prohibitive.

\subsection{Detection Method}
\label{sec:background_detection}

% Reconstruction-based detectors exploit that generative models can reconstruct synthesized images (e.g., via DDIM inversion) with higher fidelity than real images. Leveraging this property, reconstruction errors can be utilized as discriminative features to train detectors of synthetic images. 
We briefly review three representative detection methods explored in our study:

\paragraph{DIRE (Pixel-Space Reconstruction).}
DIRE~\cite{wang2023dire} operates directly in the pixel domain. It employs DDIM inversion to reconstruct the input image and defines the feature as the pixel-level $L_1$ reconstruction error, capturing diffusion artifacts lost during the re-generation cycle:
\begin{equation}
    \phi_{\text{DIRE}}(x) = | x - \Psi_{\text{rec}}(\Psi_{\text{inv}}(x))|,
    \label{eq:loss_dire}
\end{equation}
where $\Psi_{\text{inv}}(x)$ maps $x_{0} = x$ to $x_{T}$ via \eqref{eq:ode}, and $\Psi_{\text{rec}}(\Psi_{\text{inv}}(x))$ reconstructs $x_{0}$ by solving \eqref{eq:ode} in reverse under a given diffusion model $\epsilon_{\theta}$. By training a classifier on the extracted feature $\phi_{\text{DIRE}}(x)$, a robust detector is obtained.

% \paragraph{LaRE$^2$ (Latent-Space Reconstruction).} 
% To further improve efficiency, LaRE$^2$~\cite{luo2024lare} computes the reconstruction error within the compressed latent space of Latent Diffusion Models (LDMs) \cite{rombach2022high}. Instead of a full inversion-reconstruction loop, it employs a single-step forward diffusion to inject noise $\epsilon_{i}$ into the VAE-encoded latent code $\mathcal{E}(x)$. 
% %i.e., obtaining $x_{t}$ (for a fixed timestep $t$) via $\sqrt{\bar{\alpha}_{t}}\mathcal{E}(x) + \sqrt{1 - \bar{\alpha}_{t}}\epsilon_{i}$. 
% The reconstruction error score is the averaged squared gap between the injected Gaussian noise and the noise predicted by the model:

\paragraph{LaRE$^2$ (Latent-Space Reconstruction).} 
To further improve efficiency, LaRE$^2$~\cite{luo2024lare} computes reconstruction errors in the LDM latent space~\cite{rombach2022high}. Furthermore, instead of a full inversion-reconstruction loop, it employs a single-step forward diffusion to inject noise ${\epsilon}_{i}$ into the VAE-encoded latent code $\mathcal{E}(x)$ to compute the feature:
\begin{equation}
    \phi_{\text{LaRE}^2}(x) = \frac{1}{e} \sum_{i = 1}^e ({L}_{\epsilon_i} \odot {L}_{\epsilon_i}), \quad \text{where } {L}_{\epsilon_i} = {\epsilon}_i - {\epsilon}_\theta(\sqrt{\bar{\alpha}_t}\mathcal{E}(x) + \sqrt{1-\bar{\alpha}_t}{\epsilon}_i, t).
    \label{eq:loss_lare2}
\end{equation}
This single-step error serves as an efficient proxy that captures discriminative signals without the computational overhead of iterative sampling.

\paragraph{AEROBLADE (Autoencoding-based Reconstruction).}
AEROBLADE~\cite{ricker2024aeroblade} bypasses the diffusion process entirely, focusing solely on the VAE component of LDMs. It passes the image through the encoder $\mathcal{E}$ and the decoder $\mathcal{D}$, measuring the reconstruction discrepancy via the LPIPS distance~\cite{zhang2018unreasonable}:
\begin{equation}
    \phi_{\text{AEROBLADE}}(x) = d_{\text{LPIPS}}(x, \mathcal{D}(\mathcal{E}(x))).
    \label{eq:loss_aero}
\end{equation}
This indicator is then combined with a data-dependent threshold to distinguish generated content based on spectral alignment artifacts.

These reconstruction-based approaches represent three prominent methods for synthetic image detection: pixel-space (DIRE), latent-space (LaRE$^2$), and training-free VAE-based (AEROBLADE) reconstruction. Crucially, all three are differentiable, enabling the construction of adversarial attacks via standard backpropagation.
\footnote{This differentiability also applies to other reconstruction-based methods in our related work, as they rely on differentiable backbones or feature transformations.}
% \paragraph{Shared Adversarial Vulnerability.} 
% Despite their differences in operational space and complexity, these methods share a critical characteristic: full differentiability. Whether minimizing pixel distance (Eq.~\eqref{eq:loss_dire}), noise prediction error (Eq.~\eqref{eq:loss_lare2}), or perceptual loss (Eq.~\eqref{eq:loss_aero}), the entire detection pipeline constitutes a differentiable computation graph. This allows an adversary to backpropagate gradients from the detection output directly to the input image, enabling the optimization of perturbations that can effectively minimize these reconstruction errors and bypass the detectors.

%=====================================================================
\section{Adversarial Attack Analysis}
\label{sec:attack}
%=====================================================================
\begin{figure}[t]
\centering
\includegraphics[width=\linewidth]{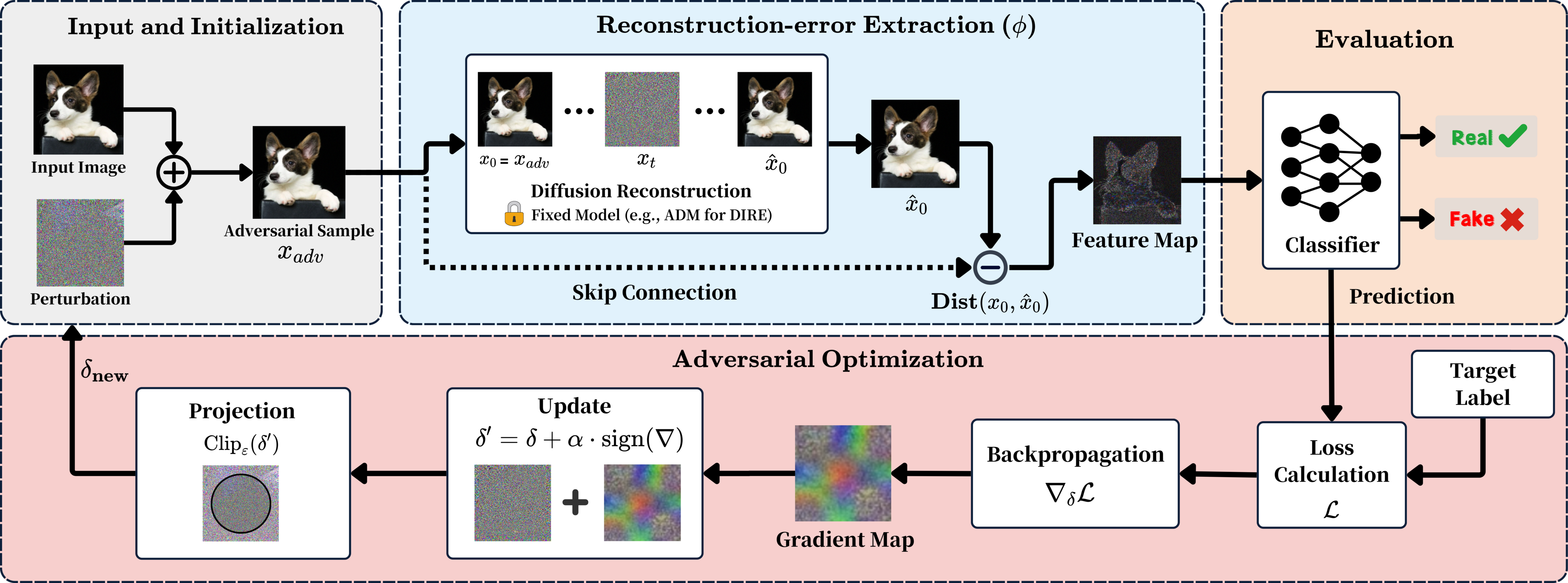}
\caption{Overview of the adversarial attack pipeline. Perturbations are optimized by backpropagating gradients through the coupled reconstruction and detection modules, treating the entire pipeline as a differentiable function.}
\label{fig:adversarial_attack}
\vspace{-3mm}
\end{figure}

% While the benchmark results in Appendix~\ref{sec:benign_performance} demonstrate the high detection efficacy of the existing detectors discussed in Section~\ref{sec:background_detection} under benign conditions, these evaluations rely on the premise of unperturbed images. However, in security-critical deployment scenarios, this assumption is often invalid. In such regimes, detectors are required to maintain accurate classification in the presence of adversaries who actively manipulate inputs to evade detection. Consequently, evaluating worst-case robustness is a prerequisite for establishing the practical reliability of reconstruction-based methods. In this section, we verify that these detectors are highly vulnerable to adversarial attacks. 

While detectors discussed in Section~\ref{sec:background_detection} perform well on unperturbed benchmarks (Appendix~\ref{sec:benign_performance}), their robustness to potential adversarial attacks is unexplored. In this section, we verify that these detectors are highly vulnerable to adversarial attacks. 

% \subsection{Threat Model and Attack Formulation}

% To evaluate worst-case robustness, we adopt the strict white-box threat model~\cite{carlini2017towards}. We assume the adversary possesses complete knowledge of the target detectors and the specific diffusion backbone used for reconstruction. This setup allows for end-to-end gradient computation ($\nabla_x$) across the entire inference pipeline.

% Formally, given an $\ell_\infty$ perturbation budget $\varepsilon$, an adversary targeting a binary detector $f_{\theta}(\phi(x))$ seeks a perturbation $\delta$ satisfying $\|\delta\|_\infty \leq \varepsilon$ to achieve:
% \begin{itemize}
%     \setlength\itemsep{0em} 
%     \item \textbf{False Negative Attack:} $f_\theta(\phi(x_{\text{fake}} + \delta)) \to \text{Real}$
%     \item \textbf{False Positive Attack:} $f_\theta(\phi(x_{\text{real}} + \delta)) \to \text{Fake}$
% \end{itemize}
% where $x_{\text{real}}$ and $x_{\text{fake}}$ represent real and synthetic images, respectively, $\phi$ denotes the feature extractor (e.g., the reconstruction error module in DIRE~\cite{wang2023dire}). To construct such an adversarial example, we solve the following constrained optimization problem that maximizes the classification loss $\mathcal{L}$ (e.g., cross-entropy loss)~\cite{goodfellow2014explaining}:
% \begin{equation}
%     \delta^* = \mathop{\arg\max}_{\|\delta\|_\infty \leq \varepsilon} \mathcal{L}\big(f_\theta(\phi(x + \delta)), y\big),
%     \label{eq:attack_objective}
% \end{equation}
% where $y$ is the ground-truth label of $x$.

\subsection{Threat Model and Attack Formulation}
To evaluate worst-case robustness, we adopt a strict white-box threat model~\cite{carlini2017towards}, assuming the adversary possesses complete knowledge of the target detector and its specific diffusion backbone. Formally, given an $\ell_\infty$ perturbation budget $\varepsilon$, an adversary targeting a binary detector, comprising a feature extractor $\phi$ and a classifier $f_\theta$, seeks a perturbation $\delta$ ($\|\delta\|_\infty \le \varepsilon$) to flip the prediction (e.g., $f_\theta(\phi(x_{\text{fake}} + \delta)) \rightarrow \text{Real}$ for a fake image $x_{\rm{fake}}$). Notably, as in \cite{wang2023dire,luo2024lare,ricker2024aeroblade}, the feature extractor $\phi$ of a detection method is usually tied to a specific diffusion backbone model, e.g., the pixel-space DDIM inversion residual of ADM for DIRE. %, the latent-space noise prediction discrepancy of SDv1.5 for LaRE$^2$, and the VAE reconstruction error of SDv1.5 for AEROBLADE. 
To construct such an adversarial example, we solve the following constrained optimization problem that maximizes the classification loss $\mathcal{L}$ (e.g., cross-entropy)~\cite{goodfellow2014explaining}:
\begin{equation}
    \delta^* = \mathop{\arg\max}_{\|\delta\|_\infty \le \varepsilon} \mathcal{L}\big(f_\theta(\phi(x + \delta)), y\big),
    \label{eq:attack_objective}
\end{equation}
where $y$ is the ground-truth label of $x$.

\subsection{Gradient Estimation via Adjoint ODE}
A significant challenge in optimizing Eq.~(\ref{eq:attack_objective}) for diffusion-based detectors lies in backpropagating through the feature extractor $\phi$, which often involves a multi-step DDIM inversion (solving an ODE). Standard backpropagation incurs an impractical $\mathcal{O}(T)$ memory footprint. As detailed in Section~\ref{subsec:adjoint_method}, we overcome this bottleneck by formulating the inversion as a Neural ODE~\cite{chen2018neural} and applying the Adjoint Sensitivity Method to compute gradients with $\mathcal{O}(1)$ memory. Equipped with this efficient gradient estimation, we solve the constrained optimization problem via $T$-step Projected Gradient Descent (PGD)~\cite{madry2017towards}:
\begin{equation}
    \delta^{(t+1)} = \Pi_{\|\cdot\|_\infty \le \varepsilon} \Big( \delta^{(t)} + \alpha \cdot \text{sign}\big(\nabla_{\delta^{(t)}} \mathcal{L}(f_\theta(\phi(x + \delta^{(t)})), y)\big) \Big),
\end{equation}
starting from a randomized initialization $\delta^{(0)}$, where $\Pi$ denotes the projection operator onto the $\ell_\infty$ $\varepsilon$-ball, $\alpha$ is the step size, and $t$ denotes the iteration step.

\subsection{Experimental Results}
\label{sec:iid experiments}

To evaluate detectors, we construct four datasets, each containing 5,000 real ImageNet~\cite{deng2009imagenet} images and 5,000 synthetic images from one of four representative diffusion backbones: ADM~\cite{dhariwal2021diffusion}, SDv1.5~\cite{rombach2022high}, FLUX~\cite{blackforest2024flux}, or VQDM~\cite{gu2022vector}. On these datasets, we evaluate three representative detectors based on diverse feature spaces: DIRE~\cite{wang2023dire}, LaRE$^2$~\cite{luo2024lare}, and AEROBLADE~\cite{ricker2024aeroblade}. As in Appendix~\ref{sec:benign_performance}, all detectors (3 $\times$ 4 = 12 models in total) are trained on each group of corresponding benign data for optimal performance prior to adversarial evaluation. Then, to attack these detectors, we employ 100-step adaptive PGD~\cite{croce2020reliable} with an $\ell_\infty$ budget of $\varepsilon = 8/255$, which adapts the step size to avoid manual tuning bias.

%Refer to Appendix~\ref{subsec:appendix_setup} for further implementation details.

\begin{figure}[t]
\centering
\begin{subfigure}{0.32\linewidth}
    \includegraphics[width=\linewidth]{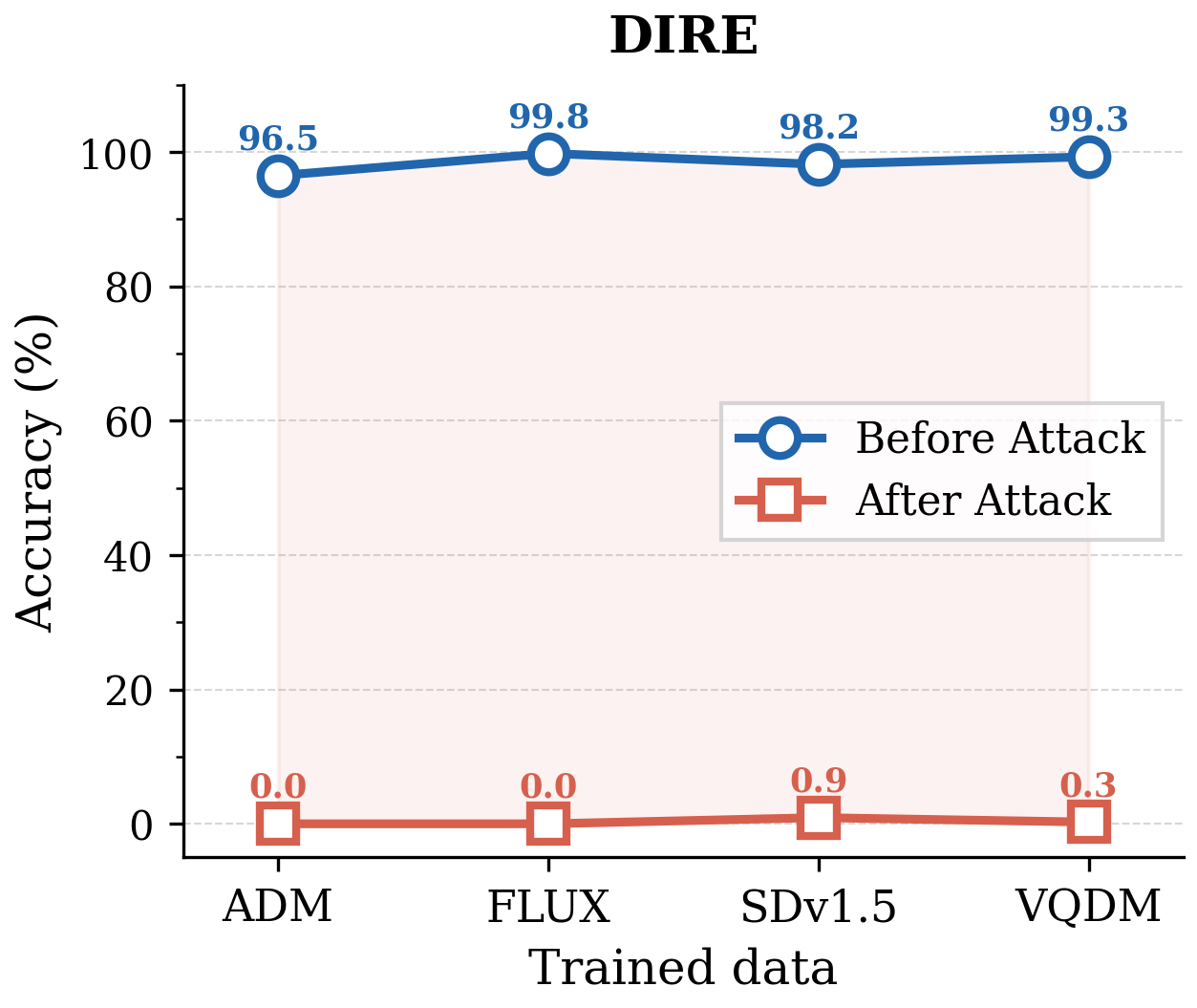}
    \caption{DIRE}
\end{subfigure}
\hfill
\begin{subfigure}{0.32\linewidth}
    \includegraphics[width=\linewidth]{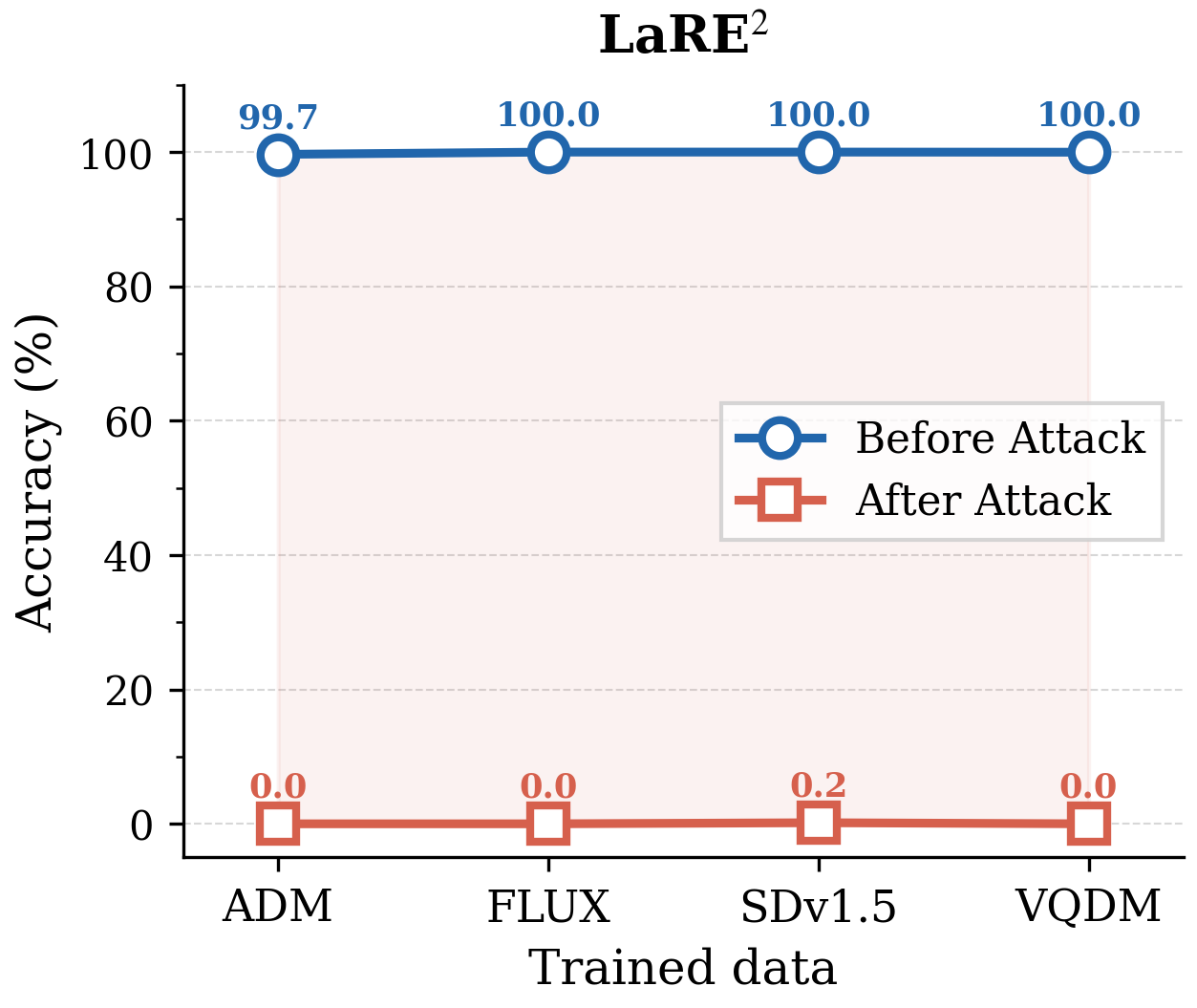}
    \caption{LaRE$^2$}
\end{subfigure}
\hfill
\begin{subfigure}{0.32\linewidth}
    \includegraphics[width=\linewidth]{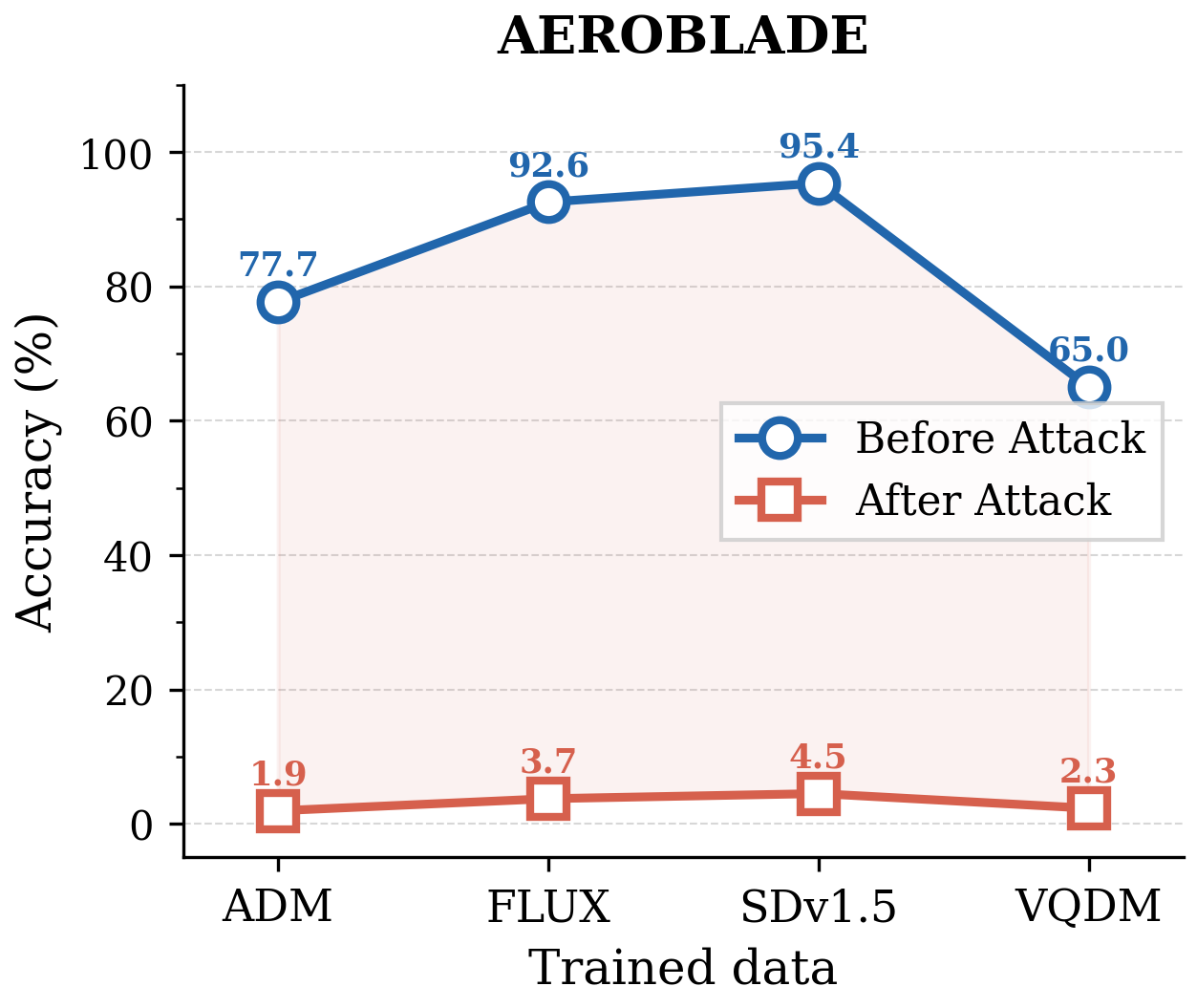}
    \caption{AEROBLADE}
\end{subfigure}
\caption{Detection performance under white-box attack. Blue lines indicate baseline accuracy on benign samples; red lines denote robust accuracy on adversarial examples.}
\label{fig:adversarial_attack_result}
\vspace{-3mm}
\end{figure}

% Following the evaluation protocol in \cite{wang2023dire}, the results presented in Fig.~\ref{fig:adversarial_attack_result}(with adversarial samples visualized in the supplementary material) 
% reveal a foundational vulnerability of various detectors to adversarial examples. As illustrated, while reconstruction-based detectors perform reliably on clean images, their performance degrades catastrophically under adversarial settings. Specifically, the robust accuracy of all three detectors drops to nearly $0\%$ across all evaluated generative models. This suggests that an adversary does not require complex optimization to detach the input from the generative manifold, rendering the reconstruction-based detection paradigm highly vulnerable to white-box attacks. This phenomenon highlights a fundamental limitation of existing detectors, exposing a critical irony: systems designed to ensure security are inherently vulnerable themselves.

Following the evaluation in DIRE~\cite{wang2023dire}, the results in Fig.~\ref{fig:adversarial_attack_result} reveal a foundational vulnerability of all detectors. As illustrated, while reconstruction-based detectors perform reliably on clean images, their robust accuracy drastically drops to nearly $0\%$ across all evaluated diffusion models under adversarial settings. %This suggests that the reconstruction-based detection paradigm is highly vulnerable to adv attacks. 
This highlights a fundamental limitation of existing detectors, exposing a critical irony: systems designed for security are inherently vulnerable themselves.

% We refer readers to Appendix~\ref{sec:appendix_adv_extended} for extensive ablation studies on these adversarial attacks. The results demonstrate that the proposed attacks remain highly effective against these detectors across a wide range of hyperparameters, including attack steps and perturbation budget $\varepsilon$.

Furthermore, extensive ablation studies in Appendix~\ref{sec:appendix_adv_extended} demonstrate that these attacks remain highly effective across a wide range of hyperparameters, including attack steps and perturbation budget $\varepsilon$.

% Following standard protocols~\cite{wang2023dire}, Fig.~\ref{fig:adversarial_attack_result} reveals a catastrophic vulnerability across all evaluated detectors. While performing reliably on benign images, their robust accuracy universally plummets to near $0\%$ across all generative models under adversarial settings. This indicates that adversaries can easily detach inputs from the target generative manifold without complex optimization, rendering the reconstruction-based paradigm fundamentally insecure against white-box attacks. Furthermore, extensive ablations in Appendix~\ref{sec:appendix_adv_extended} confirm that this high attack efficacy persists across diverse perturbation budgets ($\varepsilon$) and iteration steps.

%=====================================================================
\section{Transferability Analysis}
\label{sec:transfer}
%=====================================================================
\begin{figure}[h]
\centering
\begin{subfigure}{0.32\linewidth}
    \includegraphics[width=\linewidth]{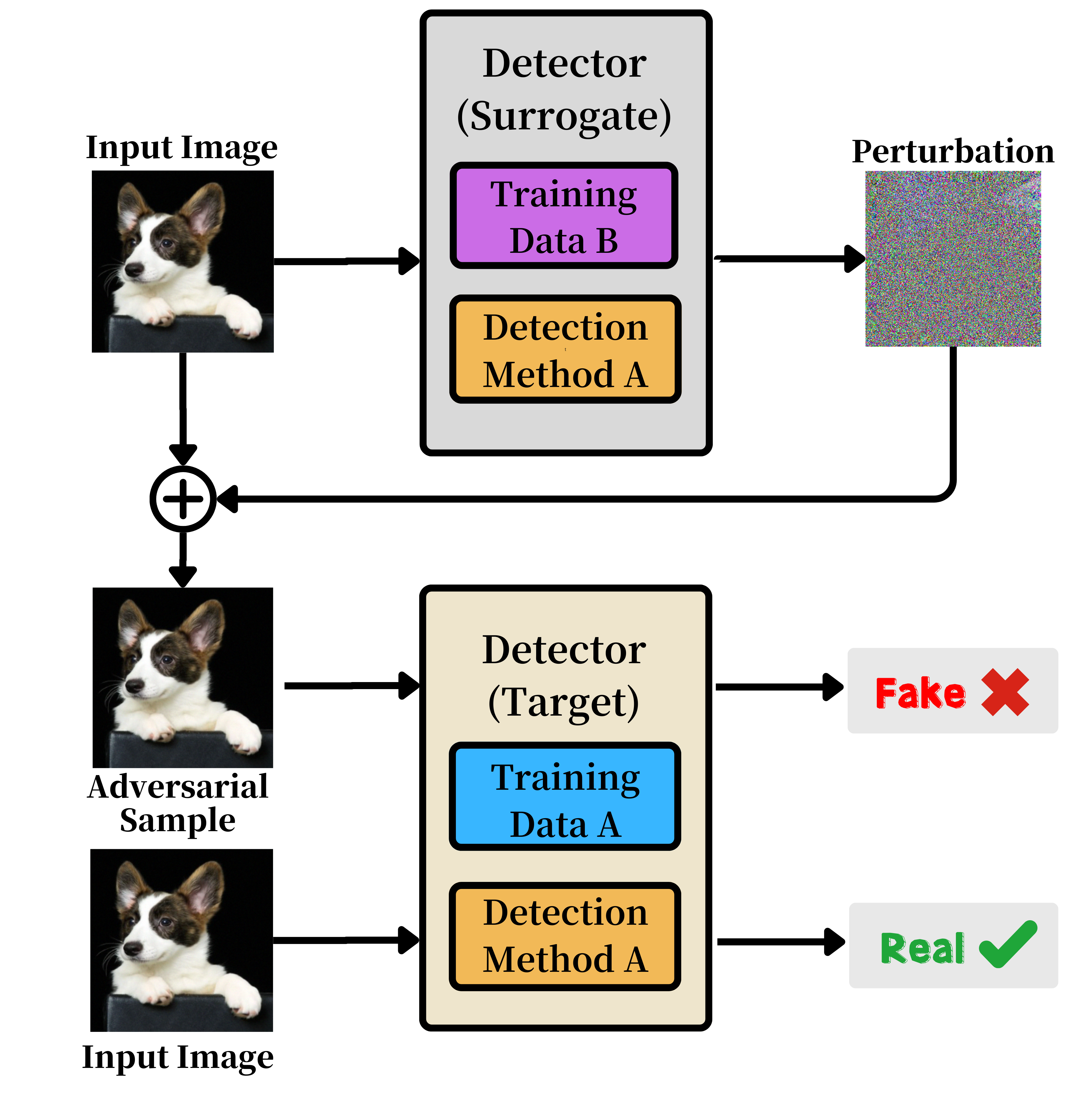}
    \caption{Generator Transferability}
\end{subfigure}
\hfill
\begin{subfigure}{0.32\linewidth}
    \includegraphics[width=\linewidth]{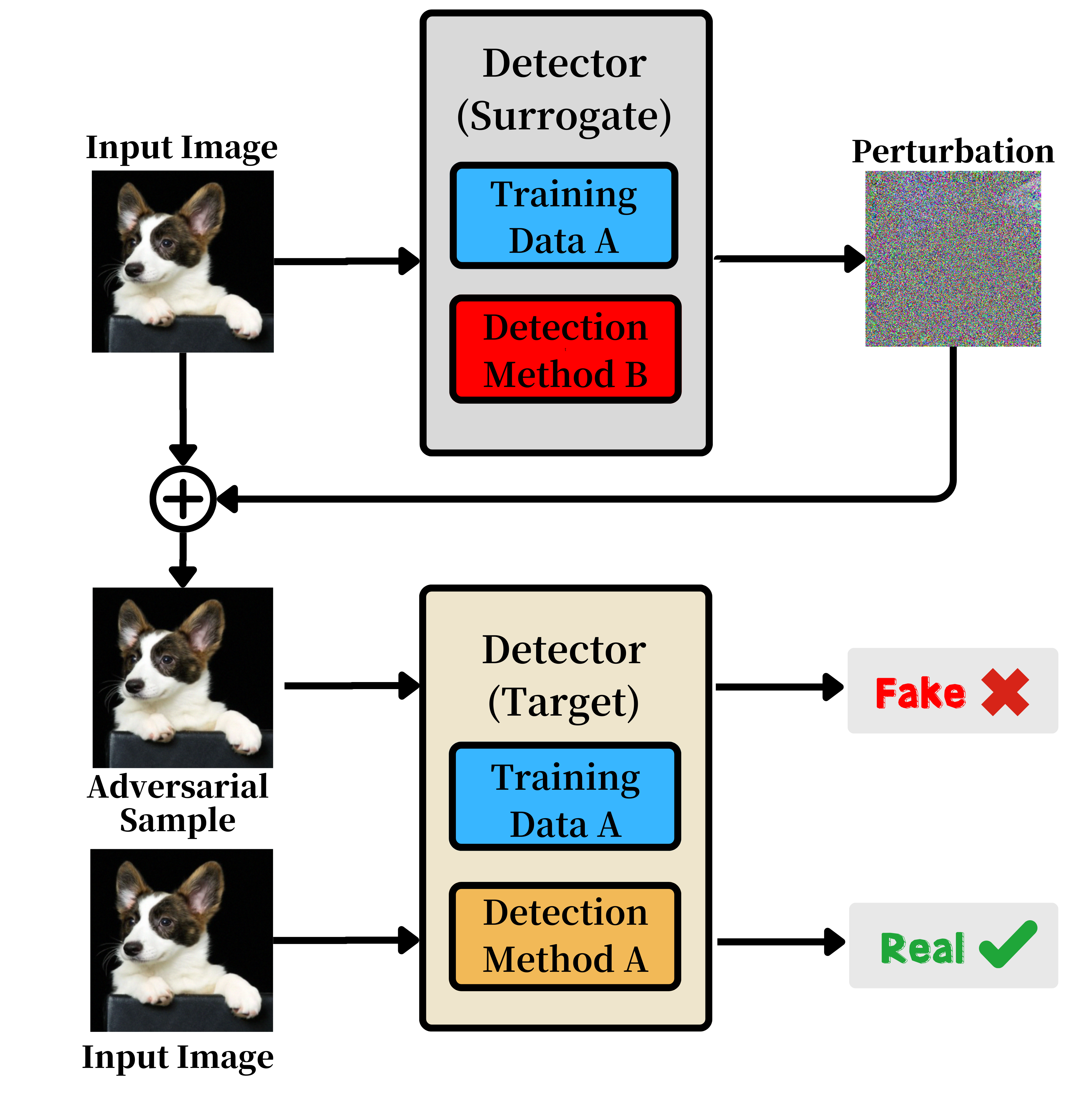}
    \caption{Method Transferability}
\end{subfigure}
\hfill
\begin{subfigure}{0.32\linewidth}
    \includegraphics[width=\linewidth]{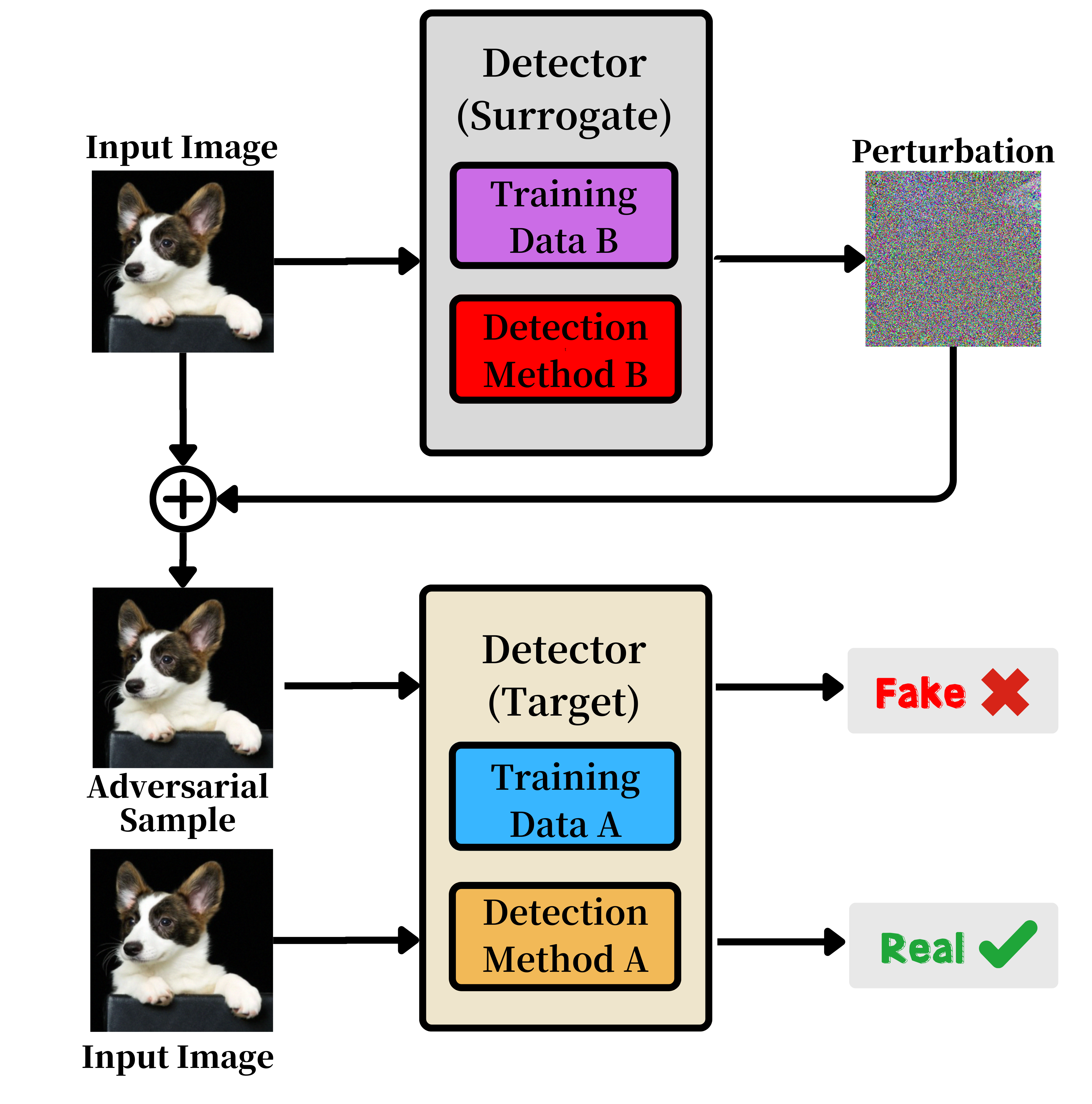}
    \caption{Both Transferability}
\end{subfigure}
\caption{\textbf{Adversarial transferability threat models.} The pipeline shows perturbations $\delta$ generated on a \textbf{Surrogate} detector (top, gray) and transferred to a \textbf{Target} detector (bottom, beige). Each detector's configuration is denoted by two color-coded blocks: \textbf{Training Data} (top) and \textbf{Detection Method} (bottom).}
\label{fig:cross_overview}
\vspace{-3mm}
\end{figure}

The attacks above are in a white-box scenario, establishing the worst-case vulnerability of detectors. However, real-world deployments often operate under black-box regimes where the training data sources or detection methodologies of detectors are unknown. This raises a question regarding the practical threat level: \textbf{Can adversarial perturbations transfer across unknown systems?}

Formally, given an input $x$ with true label $y$, an attacker optimizes a perturbation $\delta^*$ on an accessible surrogate detector (comprising a feature extractor $\phi_{src}$ and a classifier $f_{\theta_{src}}$) and evaluates it on an unseen target detector (comprising $\phi_{tgt}$ and $f_{\theta_{tgt}}$). A successful transfer occurs when:
\begin{equation}
    \delta^* = \mathop{\arg\max}_{\|\delta\|_\infty \leq \varepsilon} \mathcal{L}\big(f_{\theta_{src}}(\phi_{src}(x + \delta)), y\big) \quad \Longrightarrow \quad f_{\theta_{tgt}}(\phi_{tgt}(x + \delta^*)) \neq y
\end{equation}

% To rigorously assess this risk, we employ a hierarchical evaluation protocol, progressing from partial-knowledge scenarios to a fully black-box setting. Our empirical observations indicate that even under such restricted settings, detectors remain highly vulnerable to crafted perturbations, severely undermining their real-world security.

A target detector is primarily determined by two components: the synthetic training data source and the detection methodology. In practice, an attacker may lack knowledge of either or both. To rigorously assess the transferability risk, in the following subsections we systematically isolate these components, progressing from partial-knowledge scenarios to a fully black-box setting. Our empirical observations indicate that detectors remain highly vulnerable even under these restricted conditions, severely undermining their real-world security.

% This systematic analysis aims to disentangle whether the observed vulnerability is merely an artifact of specific generators or a more generalized characteristic of reconstruction-based detectors.

%---------------------------------------------------------------------
\subsection{Cross-Generator Transferability: Unknown Generative Source}
\label{subsec:cross_generator}
%---------------------------------------------------------------------
% Generally, a detector is primarily determined by two components: the synthetic training data source (e.g., generated by SDv1.5) and the detection methodology (e.g., DIRE). Next, 
First, we consider a threat model where the adversary knows the detection methodology (which determines the feature extractor $\phi$) but lacks knowledge of the specific generative source used to train the target detector. Consequently, the attacker crafts perturbations using a \emph{surrogate detector trained on one generator} and evaluates their transferability against a \emph{target detector trained on another generator}. The evaluation protocol follows Section~\ref{sec:iid experiments}, with the conceptual pipeline depicted in Fig.~\ref{fig:cross_overview}a.

% First, we consider cross-generator transferability, where the attacker knows the detection method (e.g., DIRE) but is unaware of the specific generative model (e.g., ADM, FLUX) used for the target's training data. Consequently, the adversary crafts perturbations using a surrogate detector trained on one generator and evaluates them against a target trained on a different one. Following the pipeline in Fig.~\ref{fig:cross_overview}a and the setup in Section~\ref{sec:iid experiments}, these results represent a partial-knowledge threat focused on the training source.

\begin{figure}[h]
\vspace{-3mm}
\centering
\begin{subfigure}{0.32\linewidth}
    \includegraphics[width=\linewidth]{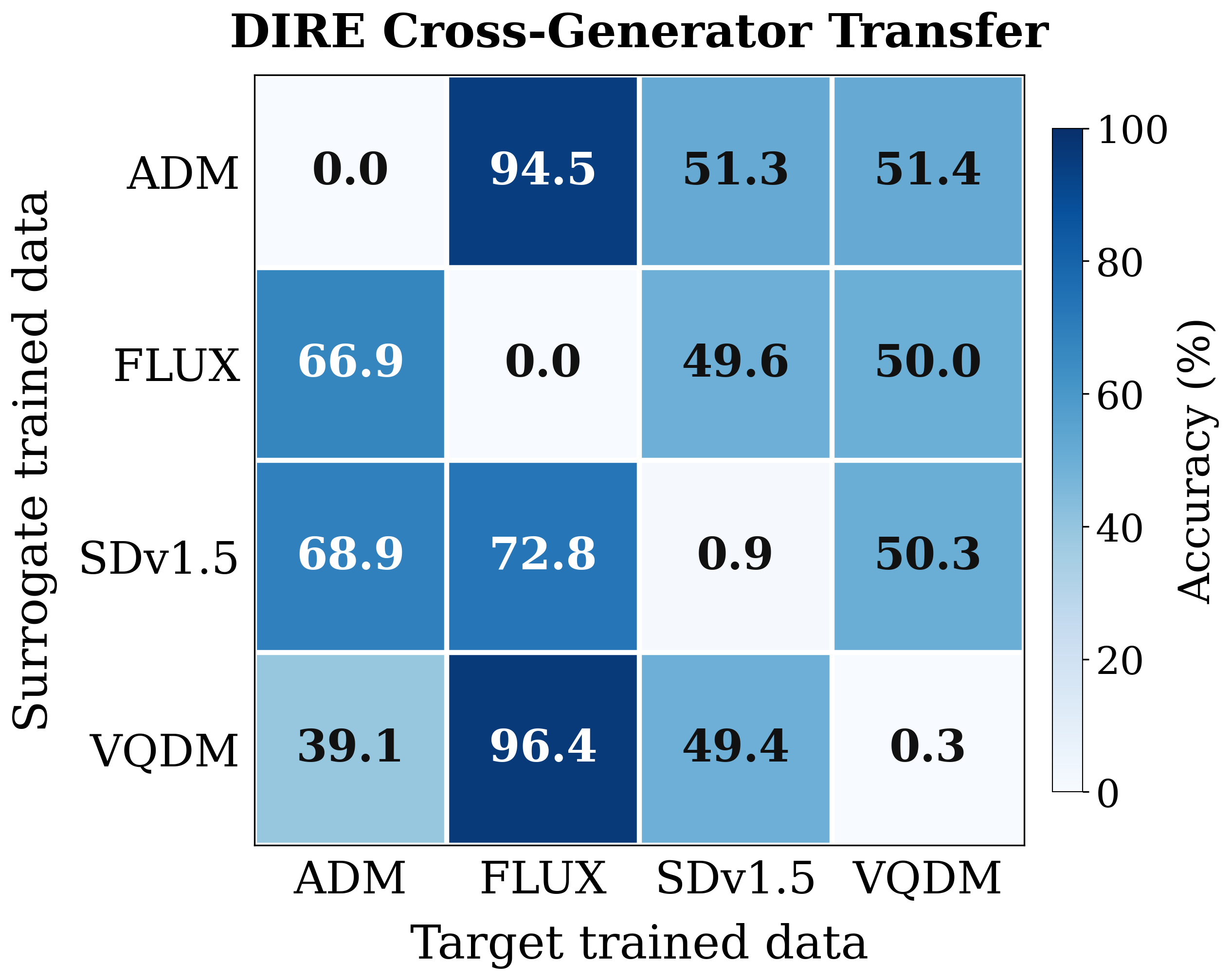}
    \caption{DIRE}
\end{subfigure}
\hfill
\begin{subfigure}{0.32\linewidth}
    \includegraphics[width=\linewidth]{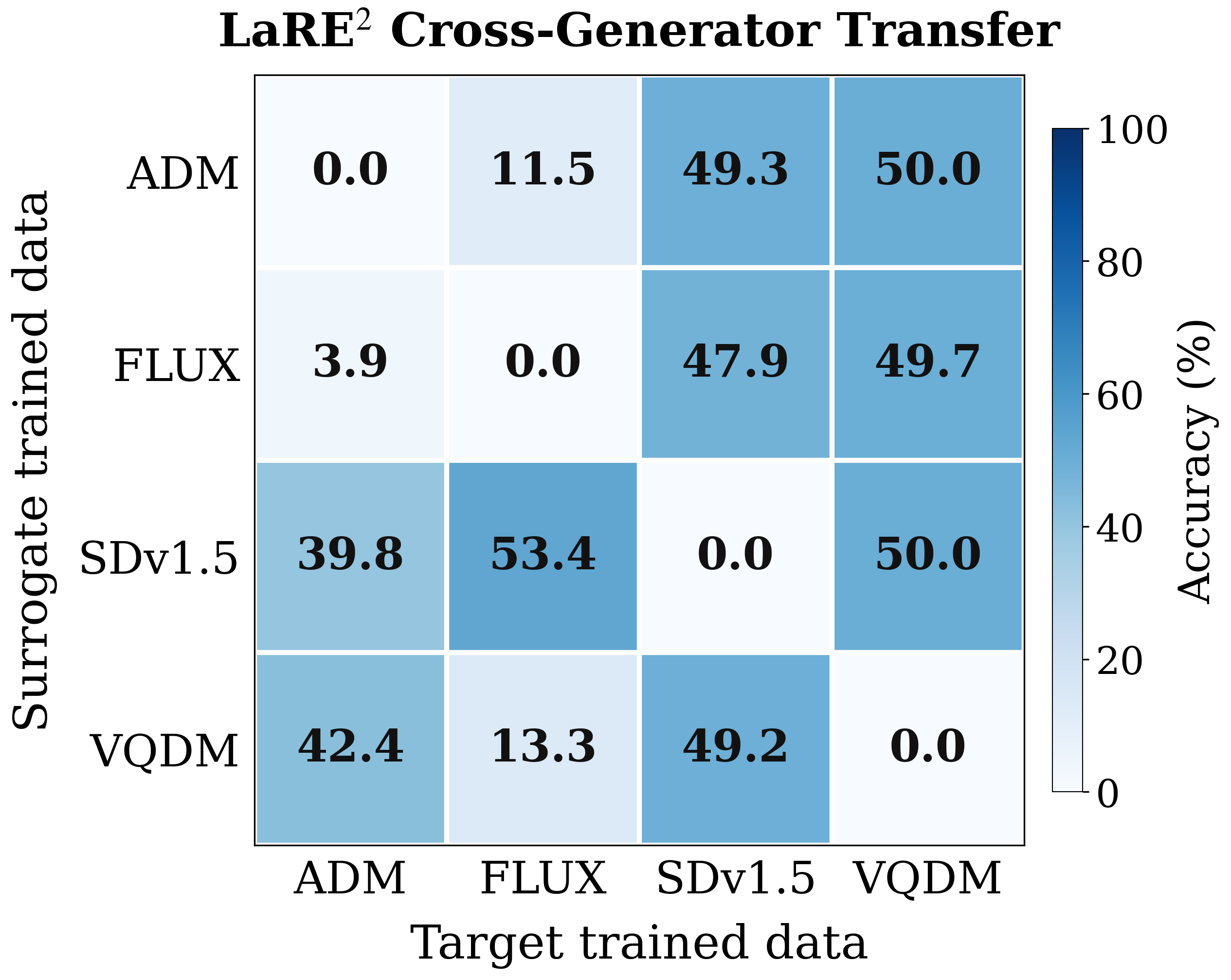}
    \caption{LaRE$^2$}
\end{subfigure}
\hfill
\begin{subfigure}{0.32\linewidth}
    \includegraphics[width=\linewidth]{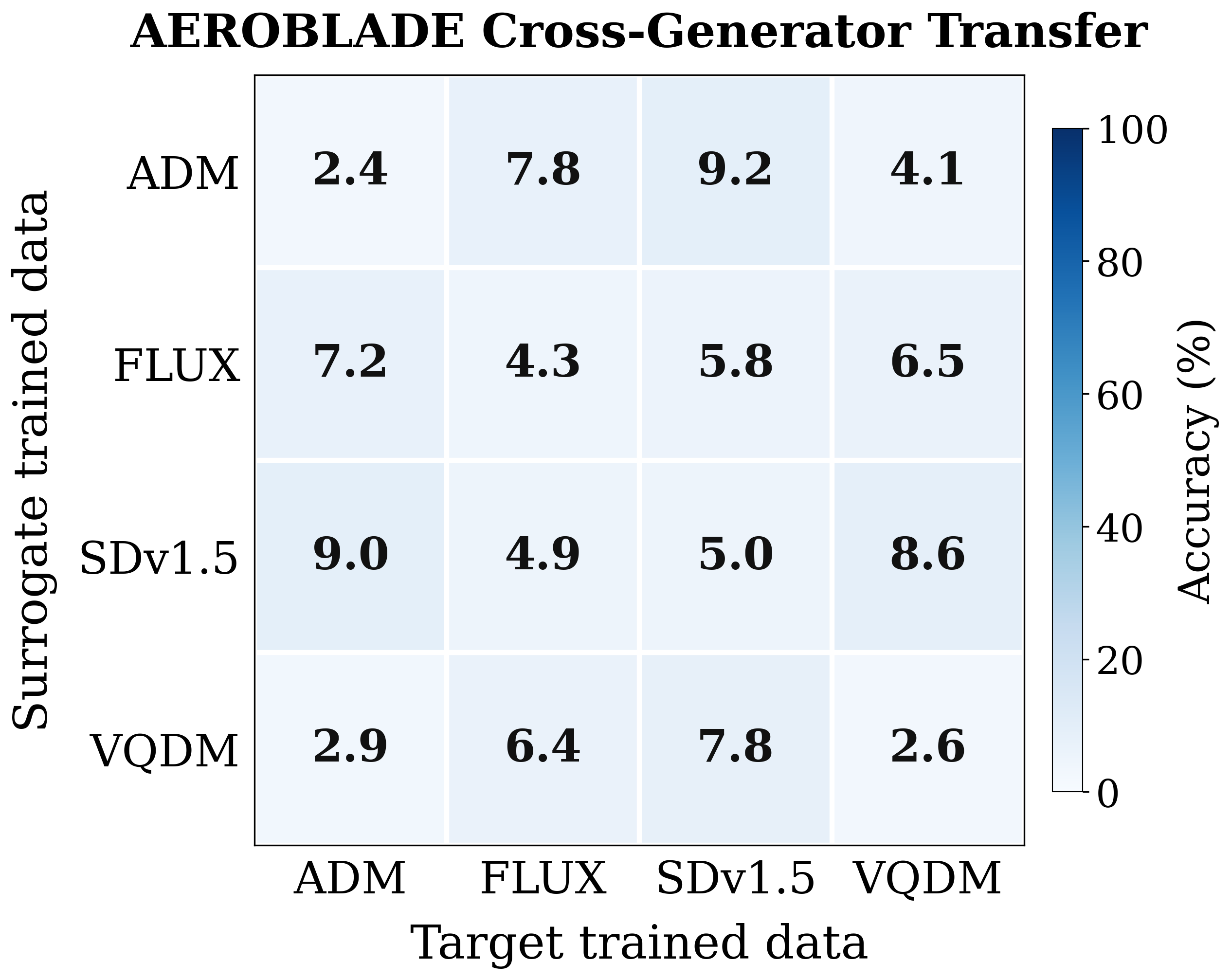}
    \caption{AEROBLADE}
\end{subfigure}
\caption{Cross-generator transfer attack results. Rows indicate the generator used to train the surrogate detector (source); columns indicate the generator used to train the target detector.}
\label{fig:cross_generator}
\vspace{-3mm}
\end{figure}
% The results, summarized in Fig.~\ref{fig:cross_generator}, reveal extensive transferability across a diverse spectrum of generative models. Concretely, in almost all evaluated cases, adversarial examples crafted using a surrogate detector trained on one source effectively deceive target detectors trained on entirely different sources. This widespread transferability suggests that the adversarial vulnerability is not tethered to the unique artifacts of a specific generator. Instead, the attack exploits a fundamental reconstruction discrepancy shared across all diffusion models. 

% We speculate that this transferability arises from the inherent similarity of the decision boundaries learned by these detectors, regardless of their training data sources. As shown in Fig.~\ref{fig:detection_performance} in the Appendix, these detectors exhibit strong generalization to out-of-distribution (OOD) \cite{yibreaking,wang2022out} synthetic data. Consequently, adversarial perturbations optimized against them inevitably inherit this generalization capability, allowing the attacks to transfer effectively.

The results, summarized in Fig.~\ref{fig:cross_generator}, reveal extensive transferability across generative models. In almost all evaluated cases, adversarial examples crafted on a surrogate detector trained on one source effectively deceive target detectors trained on entirely different sources. This suggests that the vulnerability is not tethered to artifacts of a specific generator, but instead exploits a fundamental reconstruction discrepancy shared across diffusion models. We attribute this to the inherent similarity of the learned decision boundaries: as shown in Fig.~\ref{fig:detection_performance} in the Appendix, these detectors generalize well to OOD~\cite{yibreaking,wang2022out} synthetic data, and adversarial perturbations inevitably inherit this generalization capability, allowing the attacks to transfer effectively.

%---------------------------------------------------------------------
\subsection{Cross-Method Transferability: Unknown Detection Method}
\label{subsec:cross_method}
%---------------------------------------------------------------------

% Next, we investigate cross-method transferability—specifically, whether adversarial perturbations crafted against one detection architecture (e.g., DIRE) can successfully deceive a fundamentally different one (e.g., LaRE$^2$). To evaluate this, we generate adversarial examples following the pipeline in Fig.~\ref{fig:cross_overview}b and the experimental setup in Section~\ref{sec:iid experiments}, our findings are summarized in Fig.~\ref{fig:cross_method}.

Next, we investigate cross-method transferability, where the attacker knows the training source of detectors but is unaware of the detection methodology. In this setting, we evaluate whether adversarial perturbations crafted against one detection method can successfully deceive a fundamentally different architecture trained on the same data. This configuration is illustrated in Fig.~\ref{fig:cross_overview}b, maintaining the same experimental setup in Section~\ref{sec:iid experiments}.

\begin{figure}[h]
\vspace{-3mm}
\centering
\begin{subfigure}{0.24\linewidth}
    \includegraphics[width=\linewidth]{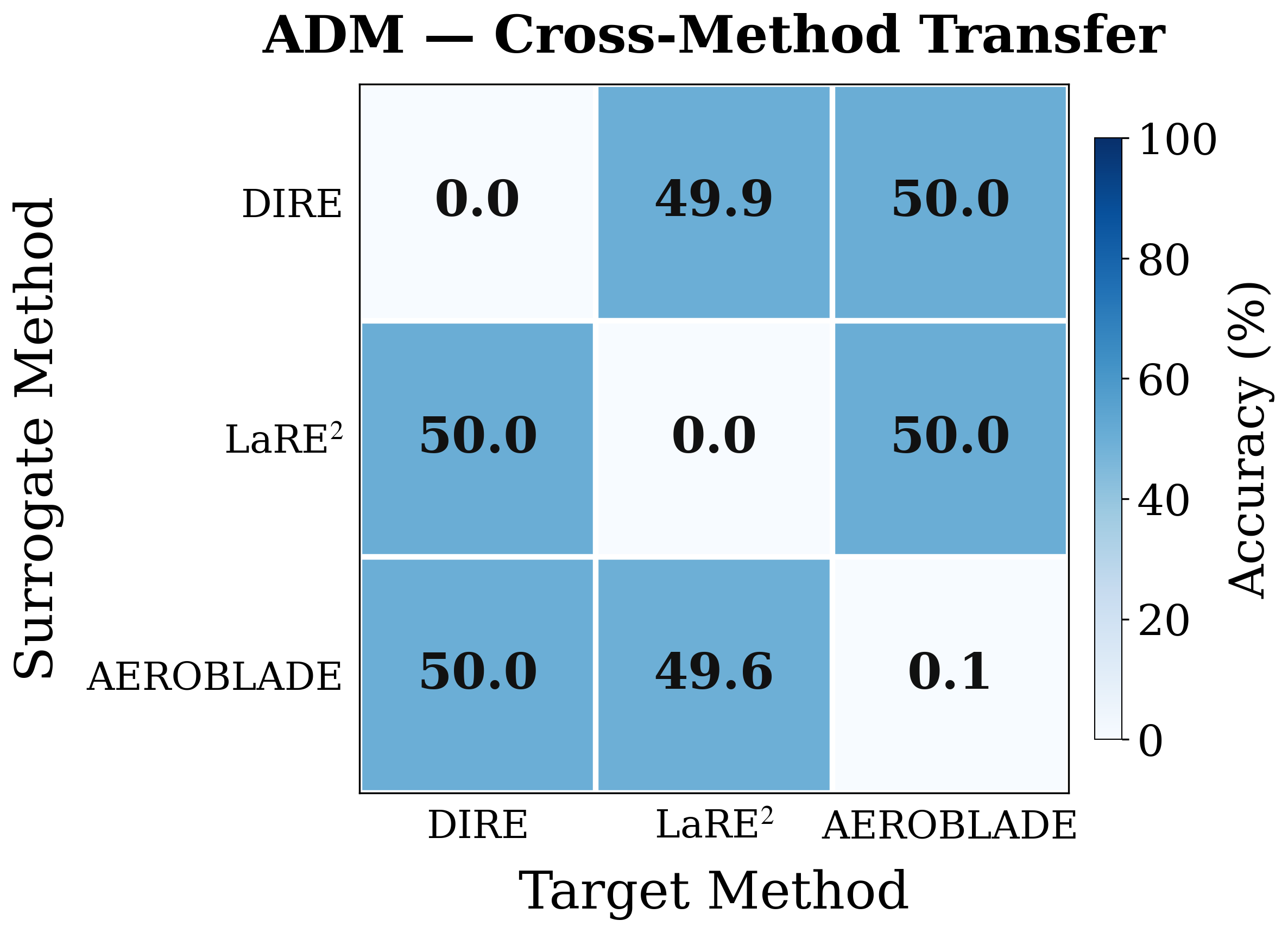}
    \caption{ADM Generator}
\end{subfigure}
\hfill
\begin{subfigure}{0.24\linewidth}
    \includegraphics[width=\linewidth]{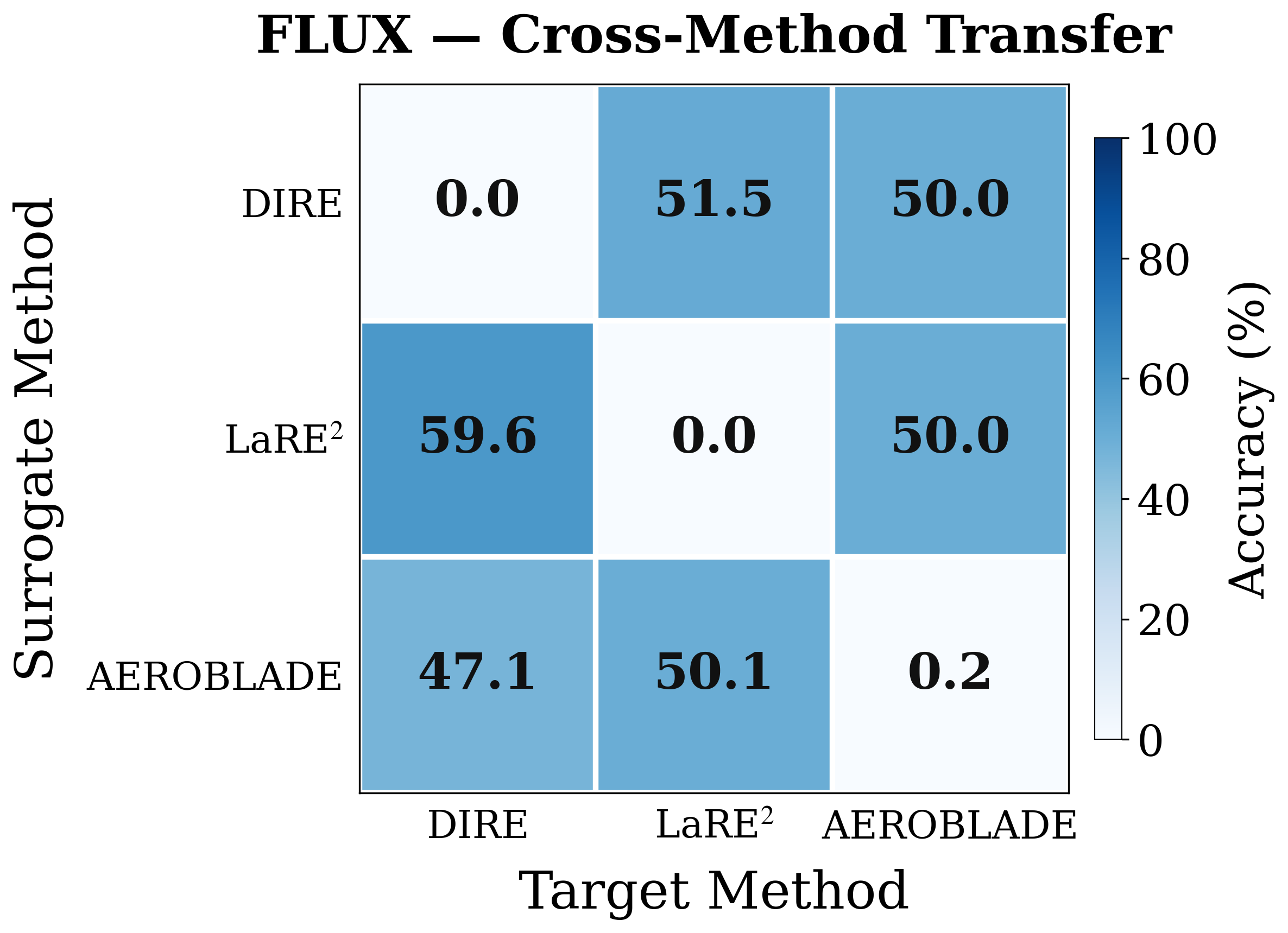}
    \caption{FLUX Generator}
\end{subfigure}
\hfill
\begin{subfigure}{0.24\linewidth}
    \includegraphics[width=\linewidth]{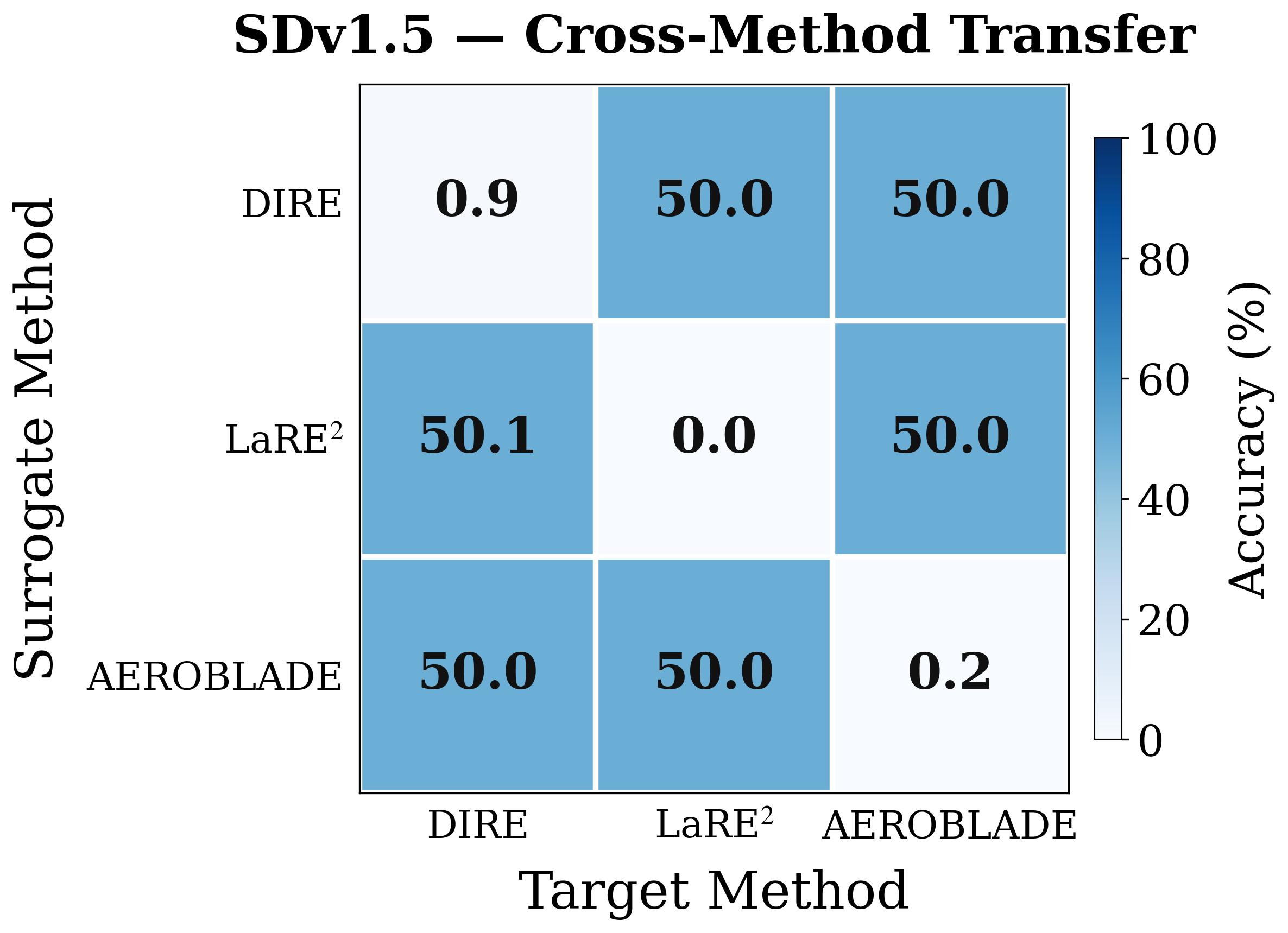}
    \caption{SD v1.5 Generator}
\end{subfigure}\hfill
\begin{subfigure}{0.24\linewidth}
    \includegraphics[width=\linewidth]{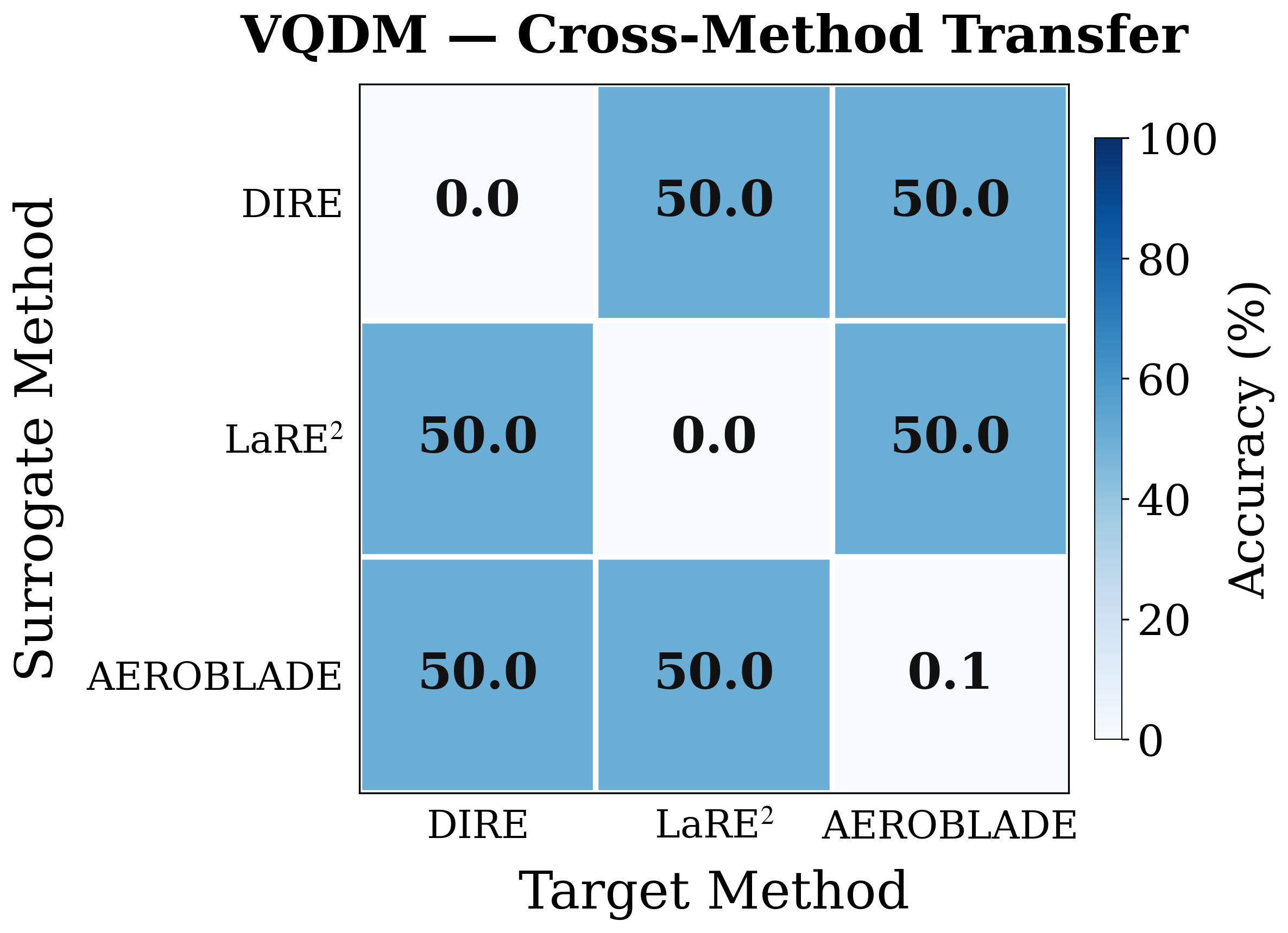}
    \caption{VQDM Generator}
\end{subfigure}
\caption{Cross-method transfer attack results. Heatmaps show post-attack accuracy (\%) where the source method (rows) attacks the target method (columns).}
\label{fig:cross_method}
\vspace{-3mm}
\end{figure}

% As illustrated in Fig.~\ref{fig:cross_method}, there is a striking uniformity in the vulnerability of these detectors to transferred attacks. Across all evaluated generators, perturbations optimized against one detector effectively deceive the others. This demonstrates that despite their distinct architectural implementations—ranging from reconstruction error in pixel-space discrepancies (DIRE) to latent-space (LaRE$^2$ and AEROBLADE)—these methods share an underlying structural similarity. They all fundamentally rely on the presence of specific reconstruction error patterns to distinguish real from synthetic images. Adversarial perturbations effectively disrupt these patterns at the signal level. Consequently, this disruption naturally transfers across any detection method that relies on extracting these features, regardless of whether its specific operating domain.

As illustrated in Fig.~\ref{fig:cross_method}, there is a striking uniformity in the vulnerability of these detectors to transferred attacks. Across all evaluated training data sources, perturbations optimized against one methodology effectively deceive the others. This demonstrates that despite their distinct operating domains---pixel-space (DIRE), latent-space (LaRE$^2$), and VAE-based (AEROBLADE)---these methods share an underlying structural similarity: they all rely on reconstruction error patterns to distinguish real from synthetic images. Adversarial perturbations effectively disrupt these patterns at the signal level, naturally transferring across any detection method that extracts similar features, regardless of its specific operating domain.

%---------------------------------------------------------------------
\subsection{Cross-Both Transferability: The Fully Black-Box Scenario}
\label{subsec:cross_both}
%---------------------------------------------------------------------
% In the previous two subsections, we evaluated transferability by keeping either the training data source or the detection method constant. In other words, although the surrogate detectors used to craft the attacks differed from the target models, the adversary still possessed partial knowledge about the downstream detectors (knowing detector training source or method). Thus, these evaluations do not represent a fully black-box threat model. To assess a more realistic threat, we now investigate transferability across both dimensions simultaneously. In this strictly black-box regime, the attacker possesses no knowledge of either the generative source or the detection architecture of the target system.

Finally, to assess a more realistic threat, we investigate transferability across both dimensions simultaneously. The previous settings each fixed one experimental variable, granting the attacker partial knowledge—this does not constitute a true black-box threat. We therefore consider a strictly black-box regime where the attacker possesses no knowledge of either the training source or the detection method of the target detector. Following the pipeline in Fig.~\ref{fig:cross_overview}c and the setup in Section~\ref{sec:iid experiments}, we report results in Table~\ref{tab:cross_both} (full results in Appendix~\ref{subsec:appendix_cross_both}).
%Fig.~\ref{fig:cross_both}.

% \begin{table}[t]
%     \centering
%     \renewcommand{\arraystretch}{1.3} 
%     \caption{Fully black-box transfer attack results. We report the average post-attack robust accuracy (\%) across all 16 generator combinations for each method pair. The performance consistently collapses to near random-guessing ($\approx 50\%$).}
%     \label{tab:cross_both}
%     \begin{tabular}{l ccc}
%     \toprule
%     \multirow{2}{*}{\textbf{Surrogate}} & \multicolumn{3}{c}{\textbf{Target Method}} \\
%     \cmidrule(lr){2-4} 
%      & \textbf{DIRE} & \textbf{LaRE$^2$} & \textbf{AEROBLADE} \\
%     \midrule
%     \textbf{DIRE}      & --- & $50.39 \pm 0.71$ & $50.00 \pm 0.00$ \\
%     \textbf{LaRE$^2$}  & $52.04 \pm 4.30$ & --- & $50.00 \pm 0.00$ \\
%     \textbf{AEROBLADE} & $49.27 \pm 1.16$ & $49.93 \pm 0.18$ & --- \\
%     \bottomrule
%     \end{tabular}
% \end{table}

\begin{table}[t]
    \centering
    \caption{Fully black-box transfer attack. We report the average post-attack robust accuracy (\%) across all 4 $\times$ 4 training source combinations for each detection method pair. The performance consistently collapses to near random-guessing ($\approx 50\%$).}
    \label{tab:cross_both}
    \begin{tabular}{@{} l @{\hspace{2em}} l c @{}} 
    \toprule
    \textbf{Surrogate} & \textbf{Target} & \textbf{Avg. Robust Acc. (\%)} \\
    \midrule
    \multirow{2}{*}{DIRE}
    & LaRE$^2$   & $50.39 \pm 0.71$ \\
    & AEROBLADE  & $50.00 \pm 0.00$ \\
    \midrule 
    \multirow{2}{*}{LaRE$^2$}  
    & DIRE       & $52.04 \pm 4.30$ \\
    & AEROBLADE  & $50.00 \pm 0.00$ \\
    \midrule 
    \multirow{2}{*}{AEROBLADE} 
    & DIRE       & $49.27 \pm 1.16$ \\
    & LaRE$^2$   & $49.93 \pm 0.18$ \\
    \bottomrule
    \end{tabular}
\end{table}

% Following an experimental setup similar to that in Section~\ref{sec:iid experiments}, we report the fully black-box transferability results in Fig.~\ref{fig:cross_both}.
The results demonstrate that strong adversarial transferability persists across both training data sources and detection methodologies, with detection accuracy consistently dropping to the random-guessing level of $\approx 50\%$. This reveals that even when attackers have absolutely no information about the target detectors, they can still successfully craft adversarial examples to bypass them.

% Interestingly, a deeper examination reveals that this severe performance degradation stems from a specific phenomenon we term \emph{deterministic collapse}. Although our attack objective is inherently bidirectional, i.e., aiming to flip both Real $\to$ Fake and Fake $\to$ Real as in white-box attack in Section \ref{sec:iid experiments}, the transferred perturbations do not symmetrically flip the labels. Instead, they overwhelm the model's decision boundary, forcing the detector to classify all inputs—regardless of their ground truth—into a single category. On a balanced dataset, this unilateral prediction naturally yields a stable $50\%$ accuracy, signifying that the detector has lost all discriminative capability and is essentially dominated by the adversarial artifacts. We conduct a dedicated empirical analysis of this deterministic collapse in Appendix~\ref{subsec:collapse_analysis}, demonstrating that when classifiers are fed with unknown inputs, they tend to collapse into one output pattern (predicting either entirely ``Real'' or entirely ``Fake'').

Interestingly, a deeper examination reveals that this severe performance degradation stems from a specific phenomenon we term \emph{deterministic collapse}. Although our attack objective is inherently bidirectional (targeting both Real $\to$ Fake and Fake $\to$ Real), the transferred perturbations do not symmetrically flip the labels. Instead, they overwhelm the model's decision boundary, forcing the detector to unilaterally classify all inputs into a single category. On a balanced dataset, this unilateral prediction naturally yields a $50\%$ accuracy, signifying a complete loss of discriminative capability. 
We provide a detailed empirical analysis of this collapse in Appendix~\ref{subsec:collapse_analysis}, demonstrating that when classifiers are fed with unknown inputs, they tend to collapse into one output pattern.

Furthermore, similar to the discussion in Section \ref{subsec:snr_analysis}, we argue that this widespread transferability originates fundamentally from the inherent sensitivity of the feature extractors. In essence, minute adversarial perturbations injected into the input images cause overwhelmingly large deviations in the extracted feature space. 
Further details on this point are provided in Appendix~\ref{sec:appendix_ratio}.
%=====================================================================
\section{Defense}
\label{sec:defense}
%=====================================================================

Given the identified vulnerabilities, we investigate whether reconstruction-based detectors can be fortified via standard defense paradigms. We analyze two representative techniques: \textbf{Diffusion Purification}~\cite{nie2022diffusion} and \textbf{Adversarial Training}~\cite{madry2017towards}. Our evaluation reveals that these methods encounter fundamental bottlenecks when applied to the reconstruction-based paradigm.

\subsection{Diffusion-Based Purification}
\label{subsec:purification}

Diffusion purification~\cite{nie2022diffusion} exploits the generative priors of diffusion models to neutralize adversarial perturbations. Conceptually, adversarial noise is treated as high-frequency artifacts residing off the data manifold. By injecting partial noise and subsequently denoising the input, the process projects the adversarial example back onto the manifold of realistic images.

% Formally, we model purification as a stochastic trajectory governed by the generative dynamics. Given an adversarial example $x_{\text{adv}}$, the process consists of a forward diffusion phase followed by a reverse restoration phase. First, we diffuse $x_{\text{adv}}$ up to a truncation time $t^*$ by SDE \eqref{eq:sde} to obtain noisy $x_{t^*}$. The injected noise disrupts the specific structure of the adversarial perturbations. 
% %\begin{equation}
% %    x_{t^*} = \sqrt{\alpha_{t^*}}x_{\text{adv}} + \sqrt{1 - \alpha_{t^*}}\epsilon, \quad \epsilon \sim \mathcal{N}(0, I),
% %    \label{eq:purify_forward}
% %\end{equation}
% %where $\alpha_{t^*}$ denotes the noise schedule derived from the SDE coefficients. This partial noise injection disrupts the specific structure of the adversarial perturbations.
% Then, the purified estimate $\hat{x}_0$ is recovered by solving the reverse-time dynamics from $t^*$ to $0$. As in \cite{nie2022diffusion}, to maximize robustness for standard continuous-time diffusion models, we employ the stochastic reverse SDE rather than the deterministic ODE used for reconstruction~\cite{song2020score}:
% \begin{equation}
%     \mathrm{d}x_t = \left[ f(x_t, t) - g(t)^2 \nabla_x \log p_t(x_t) \right] \mathrm{d}t + g(t) \mathrm{d}\bar{w}_t.
%     \label{eq:purify_reverse}
% \end{equation}
% Here, the stochastic term $g(t) \mathrm{d}\bar{w}_t$ introduces beneficial randomness during generation, aiding in the removal of residual adversarial artifacts.

Formally, we model purification as a stochastic trajectory. Given an adversarial example $x_{\text{adv}}$, we first diffuse it to a truncation time $t^*$ via SDE \eqref{eq:sde}. This process injects noise to disrupt the specific structure of adversarial perturbations. The purified estimate $\hat{x}_0$ is then recovered by solving the reverse-time SDE \eqref{eq:purify_reverse} from $t^*$ to $0$. Following \cite{nie2022diffusion}, we employ the stochastic reverse SDE rather than the deterministic ODE \cite{song2020score} to maximize robustness:
\begin{equation}
    \mathrm{d}x_t = \left[ f(x_t, t) - g(t)^2 \nabla_x \log p_t(x_t) \right] \mathrm{d}t + g(t) \mathrm{d}\bar{w}_t,
    \label{eq:purify_reverse}
\end{equation}
where the stochastic term $g(t) \mathrm{d}\bar{w}_t$ introduces randomness to help remove residual adversarial artifacts.

% We adapt this formulation to accommodate diverse architectures where the standard VP-SDE is inapplicable. For FLUX~\cite{blackforest2024flux}, which is based on Rectified Flow~\cite{liu2022flow} rather than score matching~\cite{song2020score}, we apply a deterministic Euler ODE integrator aligned with its linear interpolation forward process. For discrete latent models like VQDM~\cite{gu2022vector}, the forward process is modeled as token masking, while the reverse process involves iterative categorical sampling~\cite{austin2021structured}.

% Crucially, to ensure a rigorous evaluation, we enforce a strict source-matched protocol: the model used for purification is identical to the one that generated the input image. For instance, if an image is generated by ADM, we utilize the ADM backbone to purify it. This setting eliminates the influence of domain shifts between models, ensuring that the measured robustness stems solely from the purification mechanism itself.

\begin{figure}[t]
\centering
\begin{subfigure}{0.24\linewidth}
    \includegraphics[width=\linewidth]{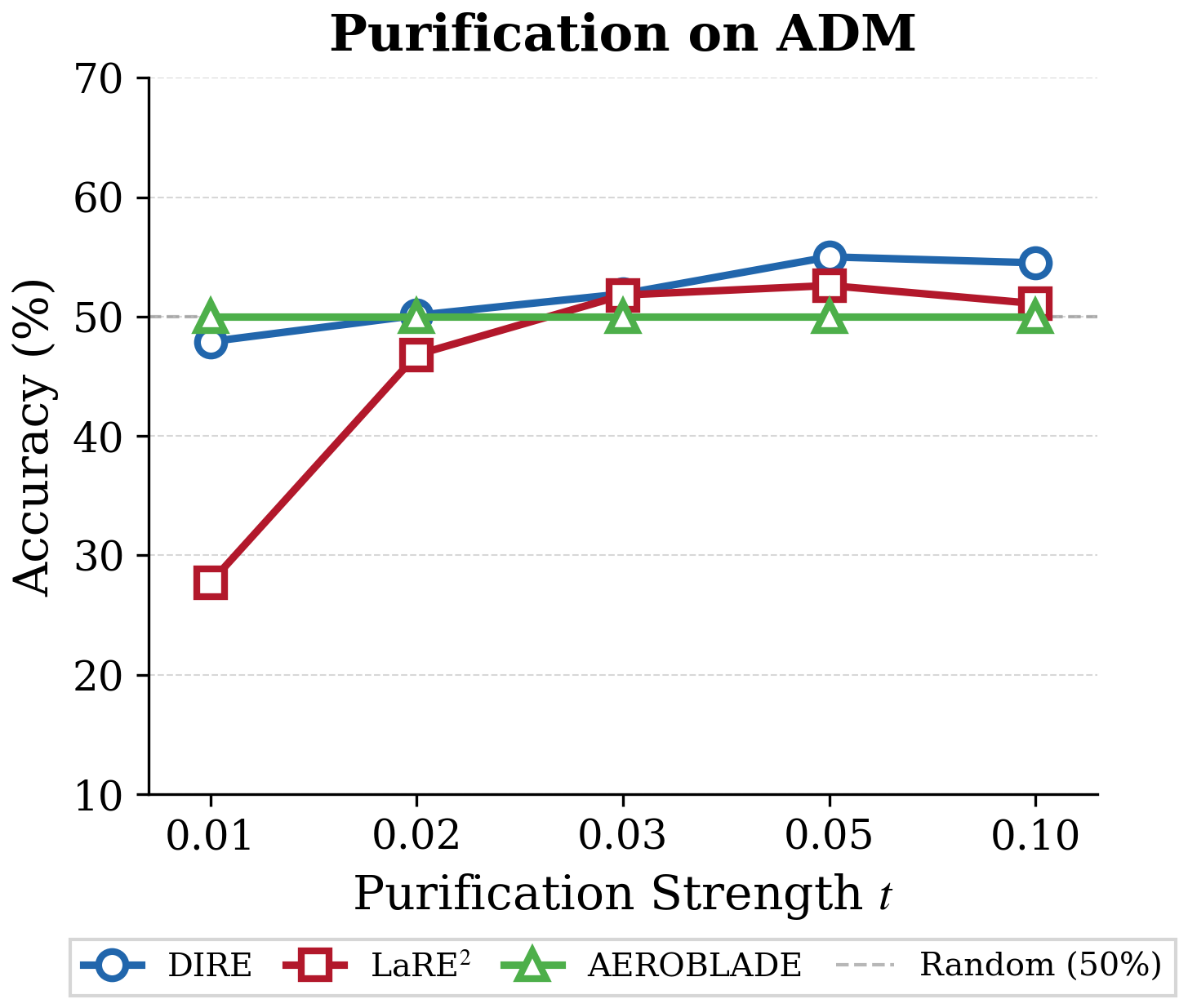}
    \caption{ADM}
\end{subfigure}
\hfill
\begin{subfigure}{0.24\linewidth}
    \includegraphics[width=\linewidth]{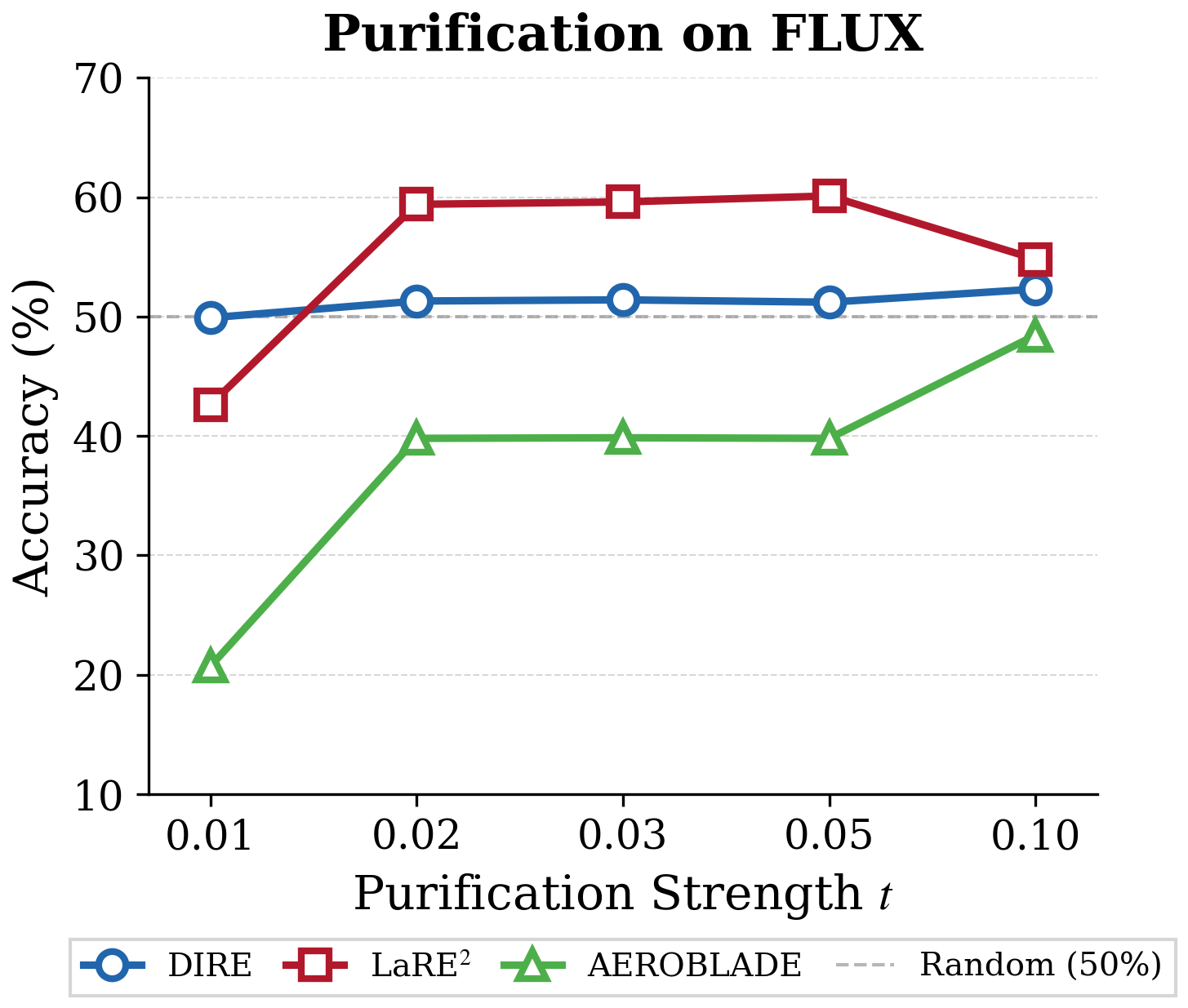}
    \caption{FLUX}
\end{subfigure}
\hfill
\begin{subfigure}{0.24\linewidth}
    \includegraphics[width=\linewidth]{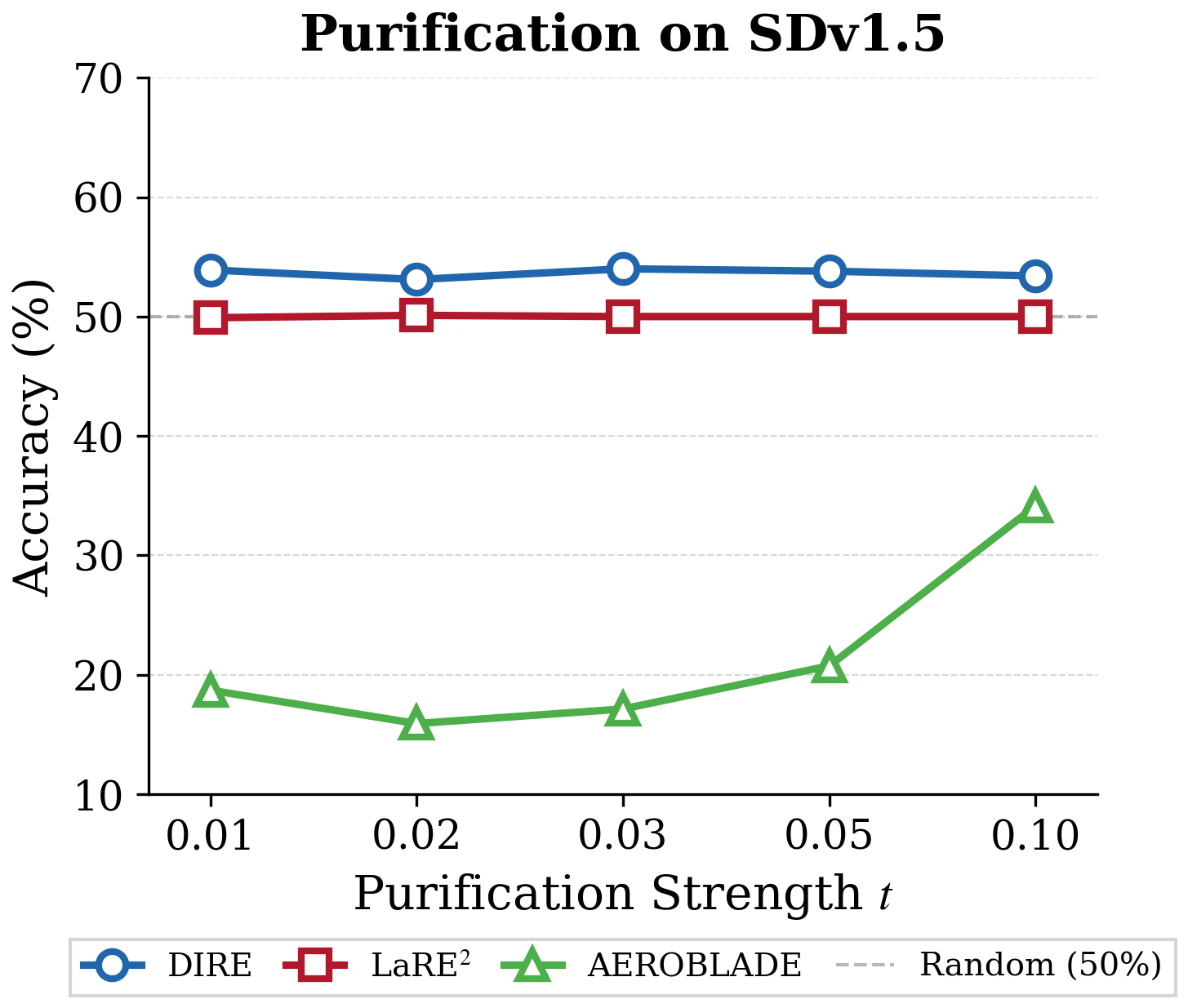}
    \caption{SD v1.5}
\end{subfigure}\hfill
\begin{subfigure}{0.24\linewidth}
    \includegraphics[width=\linewidth]{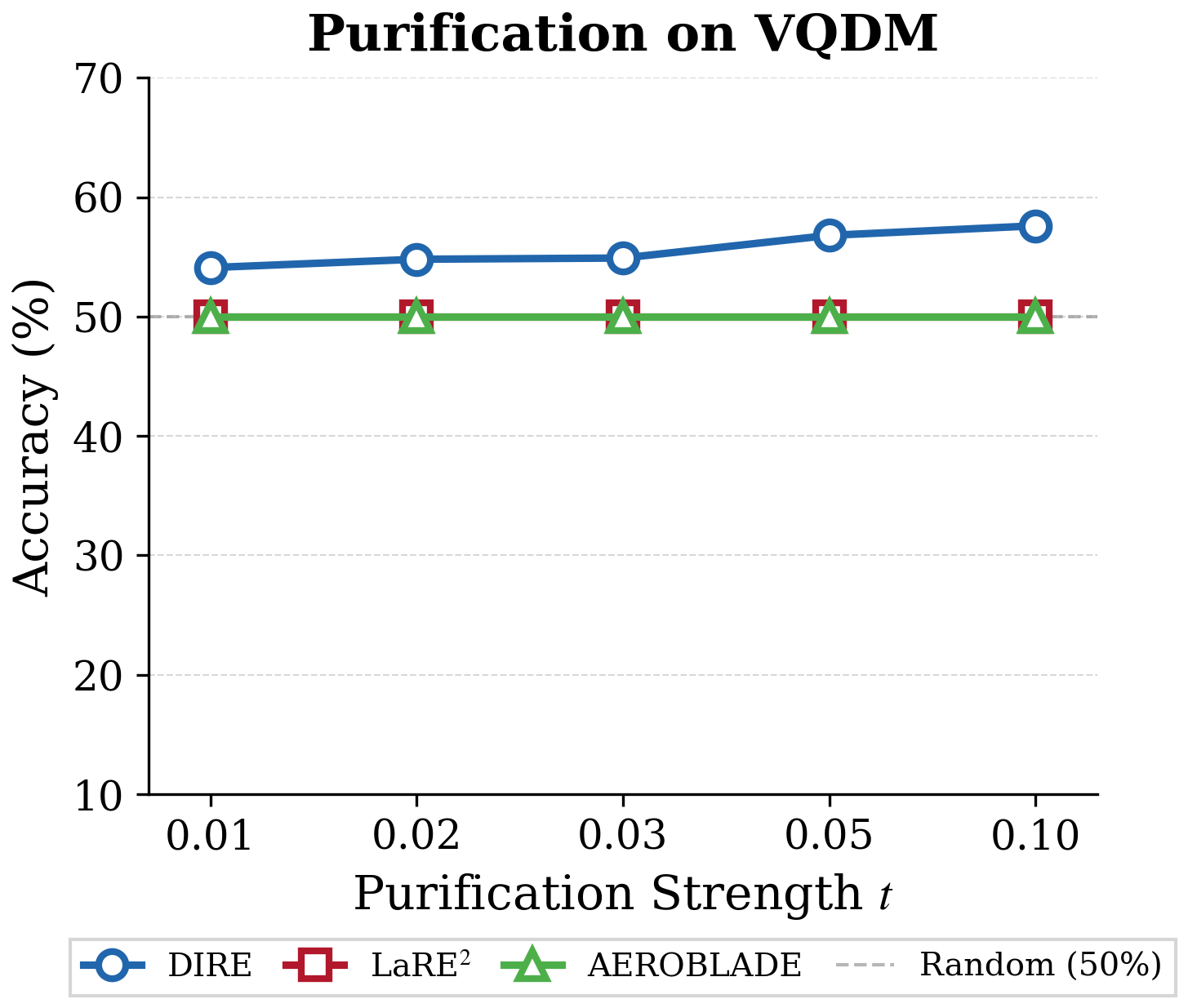}
    \caption{VQDM}
\end{subfigure}
\caption{Impact of Diffusion Purification on defense robustness. The plots illustrate detection accuracy (\%) across varying purification strengths $t$ for four generative sources. }
\label{fig:purification}
\vspace{-3mm}
\end{figure}

To fairly evaluate this defense, we enforce a source-matched protocol where adversarial examples are purified by their original generator (e.g., ADM images are purified by ADM)\footnote{For FLUX~\cite{blackforest2024flux} and VQDM~\cite{gu2022vector}, the noising then denoising process follows their own pattern, i.e., flow-based model and discrete diffusion model.}, eliminating the influence of domain shifts. The complete setup details are provided in Appendix~\ref{subsec:purification_setup}. We report accuracy trajectories in Fig.~\ref{fig:purification}.
% For a fair evaluation, 
%We sweep purification strength over $t/T \in \{0.01, 0.02, 0.03, 0.05, 0.1\}$ on 2,000 balanced images per generator-detector pair and report accuracy trajectories in Fig.~\ref{fig:purification}.

%

%=====================================================================
%=====================================================================
However, our empirical results demonstrate that this strategy is largely ineffective in the context of reconstruction-based detection. Across most settings (e.g., ADM and VQDM), the post-attack accuracy stagnates near the random-guessing level ($\approx 50\%$), failing to restore the detector's discriminative capability. 
% (refer to Appendix~\ref{subsec:purification_setup} for detailed experimental setups).
We attribute this failure to two primary factors:

\begin{enumerate}
\item \textbf{Robustness to Purification.}
This failure stems from the structural alignment between the attack and the defense mechanisms. The white-box attack optimizes the perturbation by backpropagating through the entire inference chain:
\begin{equation}
    x_{\rm{adv}} \xrightarrow{\text{Inversion}} x_T \xrightarrow{\text{Reconstruction}} \hat{x}_0 \xrightarrow{f_\theta} \text{Prediction}
\end{equation}
The purification process (Forward $\to$ Reverse) is functionally analogous to the reconstruction process used during the attack optimization (Inversion $\to$ Reconstruction). Consequently, the adversary explicitly optimizes the perturbation to survive this exact transformation logic, making successful perturbations inherently robust to additional diffusion processing.

\item \textbf{Incompatibility with Reconstruction Detection.}
More critically, we observe that applying purification to unperturbed images also causes severe performance degradation (e.g., accuracy drops to $62.8\%$, $51.1\%$, and $53.8\%$ for DIRE, LaRE$^2$, and AEROBLADE, respectively). By projecting images onto the data manifold, purification inadvertently eliminates the subtle reconstruction residuals required for detection. It effectively erases discriminative cues separating real from synthetic images, rendering the defense inherently counter-productive.
% More critically, the failure is not solely due to the attack's robustness but reflects a fundamental mismatch between purification and the detection paradigm itself. We observe that applying purification to \textbf{clean, unperturbed images} also causes severe performance degradation. For example, the detection accuracy on clean data drops to $62.8\%$, $51.1\%$, and $53.8\%$ for DIRE, LaRE$^2$, and AEROBLADE, respectively. This indicates that diffusion purification, by projecting images onto the data manifold, inadvertently eliminates the subtle reconstruction residuals required for detection. It effectively ``sanitizes'' the discriminative cues that separate real from synthetic images, rendering the defense counter-productive.

\end{enumerate}

\subsection{Adversarial Training}
\label{subsec:adv_training}

Adversarial training (AT)~\cite{madry2017towards} is a cornerstone defense mechanism designed to harden machine learning models against evasion attacks. Fundamentally, AT reformulates the standard training objective from minimizing the empirical risk on the natural distribution to solving a min-max saddle point problem:
\begin{equation}
    \min_\theta \mathbb{E}_{(x,y) \sim \mathcal{D}} \left[ \max_{\|\delta\|_\infty \leq \varepsilon} \mathcal{L}_{\text{ce}}(f_\theta(\phi(x+\delta)), y) \right].
    \label{eq:adv_train}
\end{equation}
% In this framework, the inner maximization simulates a potent adversary seeking the worst-case perturbation $\delta$ that maximizes the model's loss, while the outer minimization updates the model parameters $\theta$ to withstand these worst-case attacks. By incorporating such challenging adversarial examples directly into the training pipeline, the model is forced to learn features that are invariant to adversarial perturbations. 
By directly incorporating worst-case adversarial examples (inner maximization) into the optimization loop (outer minimization), AT forces the detector to learn features that are invariant to adversarial perturbations. 

Since AEROBLADE operates in a training-free manner, we apply this defense specifically to the trainable detectors, DIRE and LaRE$^2$. In our experiments (see Appendix~\ref{subsec:adv_training_efficient} for setup details), we directly optimize the robust objective defined in Eq.~\eqref{eq:adv_train}. We employ a multi-step PGD attack to approximate the inner maximization, systematically varying the perturbation budgets ($\varepsilon$) and iteration steps to ensure a rigorous and comprehensive evaluation of the defense's limits.

\begin{table}[t]
\centering
\caption{Impact of Adversarial Training across different generators. We observe that DIRE collapses on all subsets, while LaRE$^2$ maintains modest robustness.}
\label{tab:adv_training}
\begin{tabular}{clcc}
\toprule
\textbf{Detection Method} & \textbf{Generator} & \textbf{Clean Acc. (\%)} & \textbf{Robust Acc. (\%)} \\
\midrule
\multirow{4}{*}{DIRE} 
 & ADM & 64.4 & 0.0 \\
 & FLUX & 85.6 & 0.0 \\
 & SDv1.5 & 77.4 & 0.0 \\
 & VQDM & 76.4 & 3.4 \\
\midrule
\multirow{4}{*}{LaRE$^2$} 
 & ADM & 74.0 & 23.1 \\
 & FLUX & 98.1 & 90.3 \\
 & SDv1.5 &76.0 & 27.9 \\
 & VQDM & 75.5 & 38.5 \\
\bottomrule
\end{tabular}
\end{table}
% & SDv1.5 & 90.8 & 25.1 \\

As summarized in Table~\ref{tab:adv_training}, we observe a distinct divergence in defense effectiveness between the two methods. For DIRE, adversarial training results in a catastrophic optimization collapse. The robust accuracy drops to near zero across all generators, while clean accuracy degrades significantly. This indicates that in the pixel reconstruction space, the model fails to establish a valid decision boundary, effectively sacrificing clean performance without gaining any meaningful resistance to attacks.

% In contrast, LaRE$^2$ demonstrates a modest capacity for robustness. It achieves substantial robust accuracy on specific generators (e.g., FLUX). Although it also suffers from the clean accuracy trade-off \cite{tsipras2018robustness,zhang2019theoretically}, the key distinction is that LaRE$^2$ successfully converts this trade-off into effective defensive gains, whereas DIRE completely fails. These observations indicate that while adversarial training can mitigate attacks in specific latent-space scenarios, adversarial perturbations targeting reconstruction-based detectors remain fundamentally difficult to defend against in most cases.

In contrast, LaRE$^2$ demonstrates modest robustness, achieving substantial robust accuracy on specific generators (e.g., FLUX). Unlike DIRE, LaRE$^2$ successfully converts the clean accuracy trade-off \cite{tsipras2018robustness,zhang2019theoretically} into effective defensive gains. These observations indicate that while adversarial training mitigates attacks in specific latent-space scenarios, perturbations targeting reconstruction-based detectors remain fundamentally difficult to defend against.

We attribute this performance disparity primarily to the intrinsic Signal-to-Noise Ratio (SNR) of the respective feature spaces, which we quantitatively analyze in the following section.

\subsection{Mechanism Analysis: Low Signal-to-Noise Ratio}
\label{subsec:snr_analysis}

As discussed, we investigate the underlying cause of the performance divergence observed in Table~\ref{tab:adv_training}. We posit that the efficacy of adversarial training is fundamentally governed by the Signal-to-Noise Ratio (SNR) within the feature space—specifically, whether the magnitude of adversarial perturbations overwhelms the reconstruction-based features used for detection. To quantify this, we define the \emph{relative perturbation variation} $\rho(x)$ as:
\begin{equation}
    \rho(x) = \frac{\|\phi(x_{\rm{adv}}) - \phi(x)\|_2}{\|\phi(x)\|_{2}},
\end{equation}
which measures the magnitude of the feature deviation caused by the adversarial perturbation relative to the original feature norm. Here, $x_{\rm{adv}}$ denotes the crafted adversarial sample for the input image $x$, and $\phi(\cdot)$ is the feature extractor.

% \begin{figure}[h]
% \vspace{-3mm}
%     \centering
    
%     \begin{subfigure}[t]{0.32\textwidth}
%         \includegraphics[width=\linewidth]{image/SNR/dire/flux_ratio_l2.png}
%         \caption{proportion distribution of $\rho$ for DIRE (FLUX)}
%     \end{subfigure}
%     \hfill
%     \begin{subfigure}[t]{0.32\textwidth}
%         \includegraphics[width=\linewidth]{image/SNR/lare_decoder/flux_decoded_ratio_l2.png}
%         \caption{proportion distribution of $\rho$ for LaRE$^2$ (FLUX)}
%     \end{subfigure}
%     \hfill
%     \begin{subfigure}[t]{0.32\textwidth}
%         \includegraphics[width=\linewidth]{image/SNR/ratio_all_dl.png}
%         \caption{Mean relative perturbation ($\rho$) across datasets}
%     \end{subfigure}

%     \caption{Analysis of the relative perturbation variation $\rho(x)$. (a) and (b) illustrate the proportion distributions of $\rho$ for real and fake images evaluated on FLUX. (c) compares the mean $\rho$ across different generative models.} 
%     \label{fig:snr_ratio}
%     \vspace{-3mm}
% \end{figure}

\begin{figure}[t]
\vspace{-3mm}
    \centering
    
    \begin{subfigure}[t]{0.32\textwidth}
        \includegraphics[width=\linewidth]{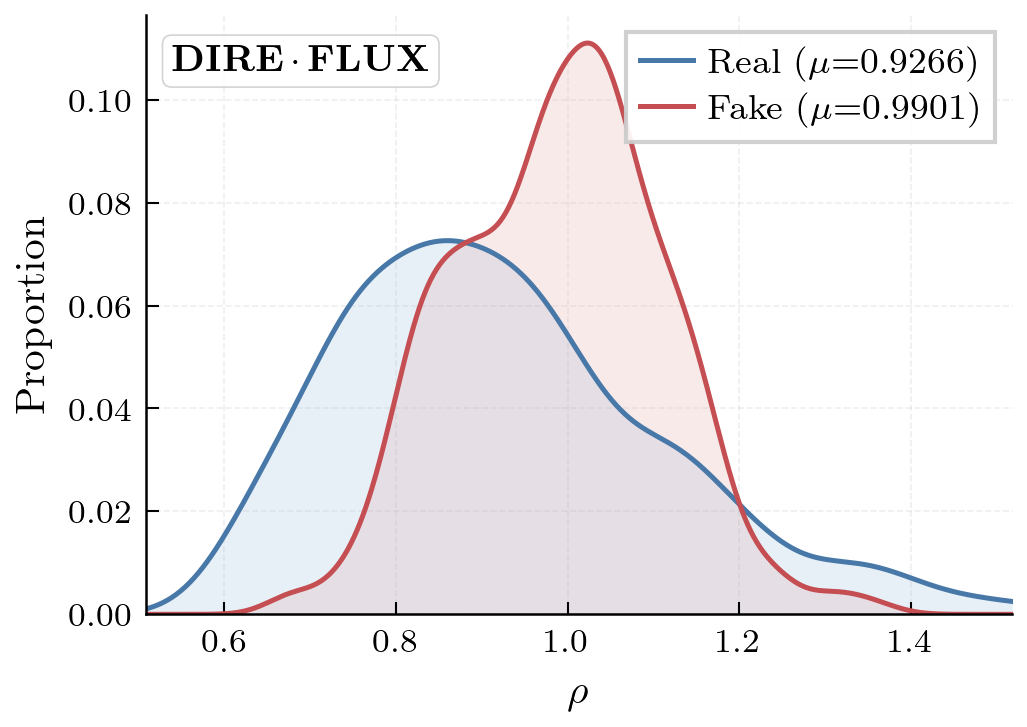}
        \caption{Proportion distribution of $\rho$ for DIRE (FLUX)}
    \end{subfigure}
    \hfill
    \begin{subfigure}[t]{0.32\textwidth}
        \includegraphics[width=\linewidth]{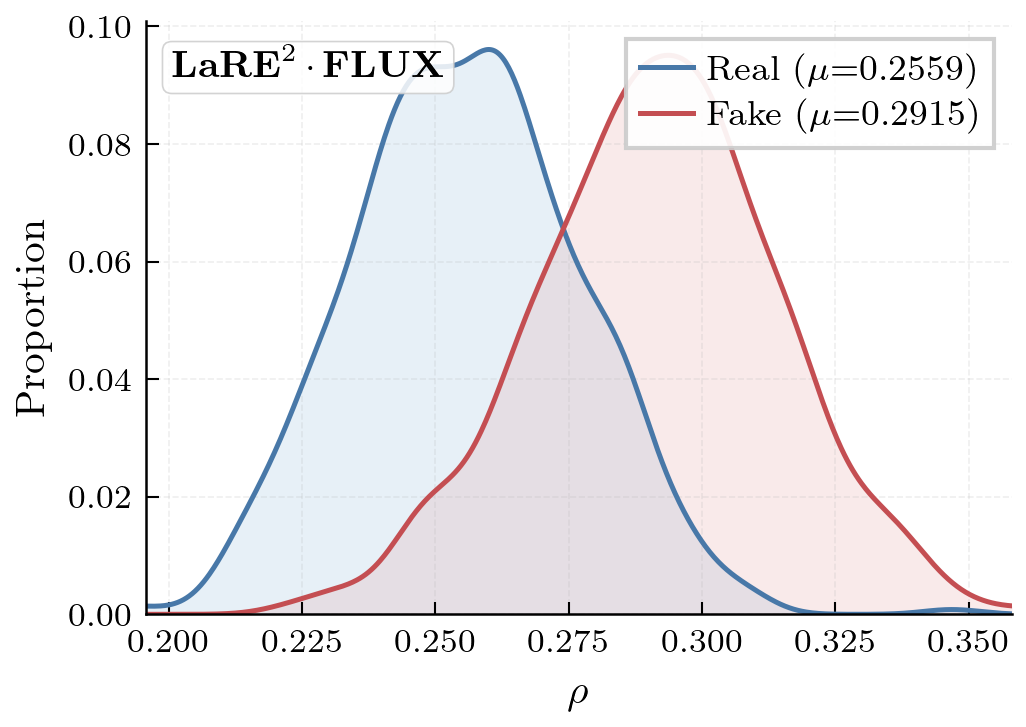}
        \caption{Proportion distribution of $\rho$ for LaRE$^2$ (FLUX)}
    \end{subfigure}
    \hfill
    \begin{subfigure}[t]{0.32\textwidth}
        \includegraphics[width=\linewidth]{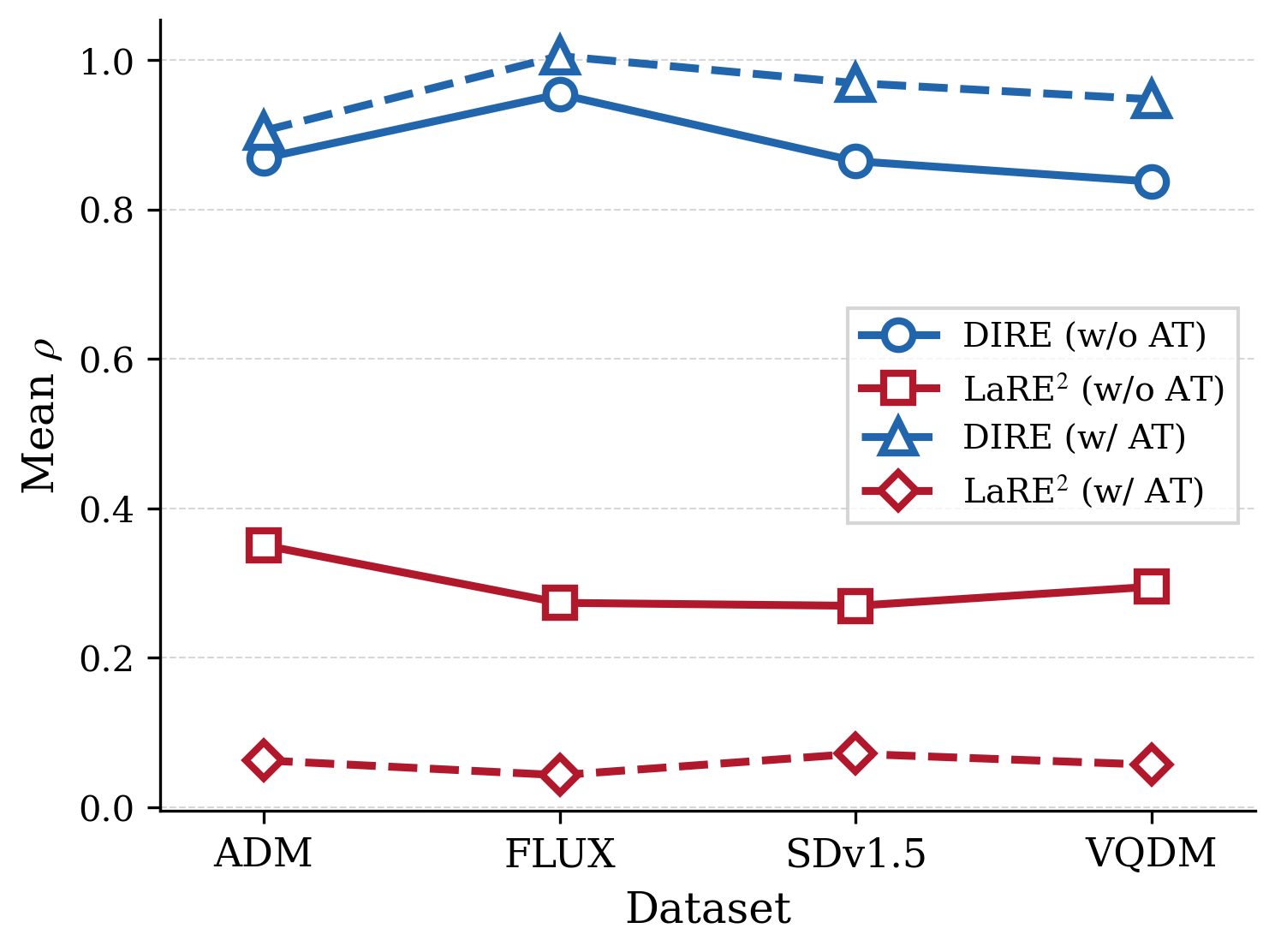}
        \caption{Mean relative perturbation ($\rho$) across datasets}
    \end{subfigure}

    \caption{Analysis of the relative perturbation variation $\rho(x)$. (a) and (b) illustrate the Proportion distributions of $\rho$ for real and fake images evaluated on FLUX. (c) compares the mean $\rho$ across different generative models.} 
    \label{fig:snr_ratio}
    \vspace{-3mm}
\end{figure}

Fig.~\ref{fig:snr_ratio} visualizes the distribution of $\rho(x)$ on the FLUX dataset, alongside the mean $\rho$ across all evaluated generators (additional distributions are deferred to the Appendix, which exhibit similar trends). To ensure a fair comparison, the latent representations of LaRE$^2$ are explicitly decoded into pixel space, perfectly aligning the evaluation domains. From these results, we derive two key observations:

% \begin{enumerate}
%     \item \textbf{A constrained relative perturbation is a prerequisite for viable adversarial training.} For DIRE, the mean relative perturbation ($\mu$) consistently approaches $0.9$. Given that the original reconstruction residuals are inherently small in magnitude, a $\rho$ value near $1.0$ indicates that the adversarial noise almost entirely masks the original discriminative signals. Forced to optimize over these corrupted representations, the classifier struggles to establish a reliable decision boundary, inevitably leading to training failure. In contrast, LaRE$^2$ maintains a sufficiently lower $\rho$, preserving enough structural integrity for the model to effectively learn and converge during adversarial training.

%     \item \textbf{Smaller relative deviations yield higher robust accuracy.} For models with better performance under adversarial training, such as LaRE$^2$, we observe a clear positive trend: lower relative deviations generally align with improved defense performance. For instance, on generators like FLUX, where the perturbation level remains naturally constrained, the model achieves superior robust accuracy. This suggests that minimizing feature deviation is a critical backbone to the efficacy of the defense (if success).
% \end{enumerate}

\begin{enumerate}
    \item \textbf{A constrained relative perturbation is a prerequisite for viable adversarial training.} For DIRE, the mean relative perturbation ($\mu$) consistently approaches $0.9$. Given that original reconstruction residuals are inherently small, a $\rho$ near $1.0$ indicates that adversarial noise almost entirely masks the discriminative signals. Forced to optimize over these corrupted representations, the classifier struggles to establish a reliable decision boundary, inevitably leading to training failure. In contrast, LaRE$^2$ maintains a sufficiently lower $\rho$, preserving enough structural integrity for the model to effectively learn and converge.

    \item \textbf{Smaller relative deviations yield higher robust accuracy.} For models like LaRE$^2$, we observe a clear positive trend: lower relative feature deviations ($\rho$) align directly with improved defense performance. For example, on FLUX, where perturbations are naturally constrained, the model achieves superior robust accuracy. Furthermore, adversarial training (AT) explicitly reinforces this mechanism. As shown in Fig.~\ref{fig:snr_ratio}c, perturbations optimized against the model with AT exhibit significantly smaller $\rho$ values compared to the standard model. This demonstrates that AT successfully restricts the adversary's ability to induce large feature-space variations, confirming that minimizing feature deviation in future work is central to defense efficacy.
\end{enumerate}

% Collectively, these findings highlight a fundamental constraint: defensive mechanisms are only viable when the underlying signal remains distinguishable from the adversarial noise. When the adversarial perturbation reaches a magnitude comparable to the natural reconstruction residuals, the semantic integrity of the features is severely compromised, rendering robust optimization unfeasible. This indicates that constraining feature deviation is a necessary foundation for robust learning. For a comprehensive analysis of the $\rho$ distributions across various generative priors, we refer the reader to Appendix~\ref{sec:appendix_ratio}.

Collectively, these findings highlight a fundamental constraint: defensive mechanisms are viable only when the underlying signal remains distinguishable from adversarial noise. Once the natural reconstruction residuals are corrupted by slight adversarial perturbations, the semantic integrity of the features is severely compromised, rendering robust optimization unfeasible. See Appendix~\ref{sec:appendix_ratio} for comprehensive $\rho$ distributions across generative priors.

\section{Conclusion}
\label{sec:conclusion}

In this paper, we present a comprehensive study on the adversarial robustness of reconstruction-based synthetic image detectors. By leveraging the adjoint method for memory-efficient gradient computation, we expose a fundamental vulnerability within this detection paradigm across both white-box and black-box regimes. We attribute this fragility to the extreme sensitivity of the extracted features, quantitatively characterized by a large relative perturbation variation. Furthermore, we demonstrate that standard defense mechanisms—namely, diffusion-based purification and adversarial training—fail to mitigate these attacks effectively, largely due to this overwhelming feature deviation. Collectively, these findings underscore a critical need to rethink the practical reliability and security of current reconstruction-based detection methods for diffusion models.

% \section*{Acknowledgements}
% TODO: Uncomment and complete this section ONLY for the camera-ready version.
% Please insert your acknowledgments here.

\bibliographystyle{splncs04}
\bibliography{reference}

\begin{thebibliography}{10}
\providecommand{\url}[1]{\texttt{#1}}
\providecommand{\urlprefix}{URL }
\providecommand{\doi}[1]{https://doi.org/#1}

\bibitem{blackforest2024flux}
{Black Forest Labs}: {FLUX}.1 (2024)

\bibitem{brooks2024video}
Brooks, T., Peebles, B., Holmes, C., DePue, W., Guo, Y., Jing, L., Schnurr, D., Taylor, J., Luhman, T., Luhman, E., et~al.: Video generation models as world simulators (2024)

\bibitem{carlini2017towards}
Carlini, N., Wagner, D.: Towards evaluating the robustness of neural networks. In: {IEEE} Computer Society (2017)

\bibitem{chen2018neural}
Chen, R.T., Rubanova, Y., Bettencourt, J., Duvenaud, D.K.: Neural ordinary differential equations. In: NeurIPS (2018)

\bibitem{chu2025fire}
Chu, B., Xu, X., Wang, X., Zhang, Y., You, W., Zhou, L.: {FIRE:} robust detection of diffusion-generated images via frequency-guided reconstruction error. In: CVPR (2025)

\bibitem{corvi2023detection}
Corvi, R., Cozzolino, D., Zingarini, G., Poggi, G., Nagano, K., Verdoliva, L.: On the detection of synthetic images generated by diffusion models. In: ICASSP (2023)

\bibitem{croce2020reliable}
Croce, F., Hein, M.: Reliable evaluation of adversarial robustness with an ensemble of diverse parameter-free attacks. In: ICML (2020)

\bibitem{deng2009imagenet}
Deng, J., Dong, W., Socher, R., Li, L.J., Li, K., Fei-Fei, L.: Imagenet: {A} large-scale hierarchical image database. In: CVPR (2009)

\bibitem{dhariwal2021diffusion}
Dhariwal, P., Nichol, A.: Diffusion models beat gans on image synthesis. In: NeurIPS (2021)

\bibitem{goodfellow2014explaining}
Goodfellow, I.J., Shlens, J., Szegedy, C.: Explaining and harnessing adversarial examples. In: ICLR (2015)

\bibitem{gu2022vector}
Gu, S., Chen, D., Bao, J., Wen, F., Zhang, B., Chen, D., Yuan, L., Guo, B.: Vector quantized diffusion model for text-to-image synthesis. In: CVPR (2022)

\bibitem{he2026grre}
He, S., Li, X., Yang, X., Xiong, Y., Li, K.: {GRRE:} leveraging g-channel removed reconstruction error for robust detection of ai-generated images. Preprint (2026)

\bibitem{hertz2022prompt}
Hertz, A., Mokady, R., Tenenbaum, J., Aberman, K., Pritch, Y., Cohen-Or, D.: Prompt-to-prompt image editing with cross-attention control. In: ICLR (2023)

\bibitem{ho2020denoising}
Ho, J., Jain, A., Abbeel, P.: Denoising diffusion probabilistic models. In: NeurIPS (2020)

\bibitem{ho2022video}
Ho, J., Salimans, T., Gritsenko, A., Chan, W., Norouzi, M., Fleet, D.J.: Video diffusion models. In: NeurIPS (2022)

\bibitem{kang2025semantic}
Kang, J.Y., Park, J., Kim, S., Yoon, J.W., Kim, N.S.: Semantic-aware reconstruction error for detecting ai-generated images. Preprint (2025)

\bibitem{luo2024lare}
Luo, Y., Du, J., Yan, K., Ding, S.: Lare\({}^{\mbox{2}}\): Latent reconstruction error based method for diffusion-generated image detection. In: CVPR (2024)

\bibitem{ma2023exposing}
Ma, R., Duan, J., Kong, F., Shi, X., Xu, K.: Exposing the fake: Effective diffusion-generated images detection. Preprint (2023)

\bibitem{ma2024latte}
Ma, X., Wang, Y., Chen, X., Jia, G., Liu, Z., Li, Y.F., Chen, C., Qiao, Y.: Latte: Latent diffusion transformer for video generation. TMLR  (2025)

\bibitem{madry2017towards}
Madry, A., Makelov, A., Schmidt, L., Tsipras, D., Vladu, A.: Towards deep learning models resistant to adversarial attacks. In: ICLR (2018)

\bibitem{mirsky2021creation}
Mirsky, Y., Lee, W.: The creation and detection of deepfakes: A survey. {ACM} Comput. Surv.  (2021)

\bibitem{mokady2023null}
Mokady, R., Hertz, A., Aberman, K., Pritch, Y., Cohen-Or, D.: Null-text inversion for editing real images using guided diffusion models. In: CVPR (2023)

\bibitem{nie2022diffusion}
Nie, W., Guo, B., Huang, Y., Xiao, C., Vahdat, A., Anandkumar, A.: Diffusion models for adversarial purification. In: ICML (2022)

\bibitem{peebles2023scalable}
Peebles, W., Xie, S.: Scalable diffusion models with transformers. In: ICCV (2023)

\bibitem{ricker2024aeroblade}
Ricker, J., Lukovnikov, D., Fischer, A.: {AEROBLADE:} training-free detection of latent diffusion images using autoencoder reconstruction error. In: CVPR (2024)

\bibitem{rombach2022high}
Rombach, R., Blattmann, A., Lorenz, D., Esser, P., Ommer, B.: High-resolution image synthesis with latent diffusion models. In: CVPR (2022)

\bibitem{somepalli2023diffusion}
Somepalli, G., Singla, V., Goldblum, M., Geiping, J., Goldstein, T.: Diffusion art or digital forgery? investigating data replication in diffusion models. In: CVPR (2023)

\bibitem{song2020denoising}
Song, J., Meng, C., Ermon, S.: Denoising diffusion implicit models. In: ICLR (2021)

\bibitem{song2020score}
Song, Y., Sohl-Dickstein, J., Kingma, D.P., Kumar, A., Ermon, S., Poole, B.: Score-based generative modeling through stochastic differential equations. In: ICLR (2021)

\bibitem{szegedy2013intriguing}
Szegedy, C., Zaremba, W., Sutskever, I., Bruna, J., Erhan, D., Goodfellow, I., Fergus, R.: Intriguing properties of neural networks. In: ICLR (2014)

\bibitem{tsipras2018robustness}
Tsipras, D., Santurkar, S., Engstrom, L., Turner, A., Madry, A.: Robustness may be at odds with accuracy (2019)

\bibitem{vaccari2020deepfakes}
Vaccari, C., Chadwick, A.: Deepfakes and disinformation: Exploring the impact of synthetic political video on deception, uncertainty, and trust in news. Social media + society  (2020)

\bibitem{vasilcoiu2025latte}
Vasilcoiu, A., Najdenkoska, I., Geradts, Z., Worring, M.: {LATTE:} latent trajectory embedding for diffusion-generated image detection. Preprint (2025)

\bibitem{wang2022out}
Wang, R., Yi, M., Chen, Z., Zhu, S.: Out-of-distribution generalization with causal invariant transformations. In: CVPR (2022)

\bibitem{wang2020cnn}
Wang, S.Y., Wang, O., Zhang, R., Owens, A., Efros, A.A.: Cnn-generated images are surprisingly easy to spot... for now. In: CVPR (2020)

\bibitem{wang2019improving}
Wang, Y., Zou, D., Yi, J., Bailey, J., Ma, X., Gu, Q.: Improving adversarial robustness requires revisiting misclassified examples. In: ICLR (2020)

\bibitem{wang2025improved}
Wang, Z., Yi, M., Xue, S., Li, Z., Liu, M., Qin, B., Ma, Z.M.: Improved diffusion-based generative model with better adversarial robustness. In: ICLR (2025)

\bibitem{wang2023dire}
Wang, Z., Bao, J., Zhou, W., Wang, W., Hu, H., Chen, H., Li, H.: {DIRE} for diffusion-generated image detection. In: ICCV (2023)

\bibitem{yi2021reweighting}
Yi, M., Hou, L., Shang, L., Jiang, X., Liu, Q., Ma, Z.M.: Reweighting augmented samples by minimizing the maximal expected loss. In: ICLR (2021)

\bibitem{yi2021improved}
Yi, M., Hou, L., Sun, J., Shang, L., Jiang, X., Liu, Q., Ma, Z.: Improved {OOD} generalization via adversarial training and pretraing. In: ICML (2021)

\bibitem{yi2024towards}
Yi, M., Li, A., Xin, Y., Li, Z.: Towards understanding the working mechanism of text-to-image diffusion model. In: NeurIPS (2024)

\bibitem{yi2023generalization}
Yi, M., Sun, J., Li, Z.: On the generalization of diffusion model. Preprint (2023)

\bibitem{yibreaking}
Yi, M., Wang, R., Sun, J., Li, Z., Ma, Z.M.: Breaking correlation shift via conditional invariant regularizer. In: ICLR (2023)

\bibitem{zhang2019theoretically}
Zhang, H., Yu, Y., Jiao, J., Xing, E., El~Ghaoui, L., Jordan, M.: Theoretically principled trade-off between robustness and accuracy. In: ICML (2019)

\bibitem{zhang2018unreasonable}
Zhang, R., Isola, P., Efros, A.A., Shechtman, E., Wang, O.: The unreasonable effectiveness of deep features as a perceptual metric. In: CVPR (2018)

\end{thebibliography}

%=====================================================================

\clearpage

%=====================================================================
\appendix
\section{Appendix}
%=====================================================================

\subsection{Details of Generative Models}
\label{sec:model_details}

To ensure the comprehensive diversity of our benchmark, we select four generative models that exemplify distinct evolutionary stages and architectural paradigms in image synthesis.

\begin{itemize}[leftmargin=*, noitemsep]
    \item \textbf{Ablated Diffusion Model (ADM)~\cite{dhariwal2021diffusion}:} 
    As a representative of foundational \textit{pixel-space} diffusion models, ADM operates directly on high-dimensional image pixels using a standard U-Net architecture. It serves as a baseline for uncompressed diffusion processes, generating images characterized by high-frequency noise residuals typical of pixel-level denoising operations.

    \item \textbf{Stable Diffusion v1.5 (SDv1.5)~\cite{rombach2022high}:} 
    Representing the widely adopted \textit{Latent Diffusion Model (LDM)} family, SDv1.5 utilizes a Variational Autoencoder (VAE) to compress images into a low-dimensional perceptual latent space before performing the diffusion process. This two-stage mechanism (compression followed by generation) introduces artifacts structurally distinct from pixel-space models, primarily originating from the VAE reconstruction process.

    \item \textbf{Vector Quantized Diffusion Model (VQDM)~\cite{gu2022vector}:} 
    VQDM integrates diffusion models with Vector Quantized Variational Autoencoders (VQ-VAE). Unlike continuous latent models, VQDM operates within a \textit{discrete} vector-quantized latent space. This quantization step introduces unique grid-like artifacts and discrete discontinuities, adding a challenging variation to our detection benchmark.

    \item \textbf{FLUX~\cite{blackforest2024flux}:} 
    Representing the state-of-the-art in generative capabilities, FLUX shifts from the traditional diffusion paradigm to \textit{Flow Matching} (specifically Rectified Flow). Architecturally, it replaces the conventional U-Net with a scalable \textbf{Transformer (DiT)} backbone. This fundamental structural shift results in a significantly different gradient landscape and image manifold compared to standard diffusion models, providing a rigorous test for the transferability of adversarial attacks across diverse generative architectures.
\end{itemize}

%=====================================================================

\subsection{Detection Performance Under Benign Conditions}
\label{sec:benign_performance}

Before analyzing adversarial vulnerabilities, we establish the baseline performance of the target detectors under non-adversarial (benign) settings. We evaluate DIRE~\cite{wang2023dire}, LaRE$^2$~\cite{luo2024lare}, and AEROBLADE~\cite{ricker2024aeroblade} on images generated by the four aforementioned models.

\paragraph{Experimental Setup.}
We construct a rigorous cross-generator benchmark to evaluate detection generalization:
\begin{itemize}[leftmargin=*, noitemsep]
    \item \textbf{Dataset Construction:} To ensure robust training and fair evaluation, we strictly partition our data into disjoint training and testing subsets. We compile a training set of 80,000 images (40,000 real images sourced from ImageNet and 40,000 generated) for each of the four generative models (ADM, SDv1.5, FLUX, VQDM). Performance is reported on a separate, held-out evaluation set comprising 10,000 images (5,000 real, 5,000 generated) per generator. All generated images utilize prompts derived from ImageNet class labels to maintain strict semantic alignment with the real samples.
    
    \item \textbf{Evaluation Protocol:} We report the detection accuracy (\%). The evaluation follows a cross-generator protocol where rows in the heatmaps denote the source generator used for training the detector, and columns denote the target generator used for testing.
\end{itemize}

\begin{figure}[t]
\centering
\begin{subfigure}{0.32\linewidth}
    \includegraphics[width=\linewidth]{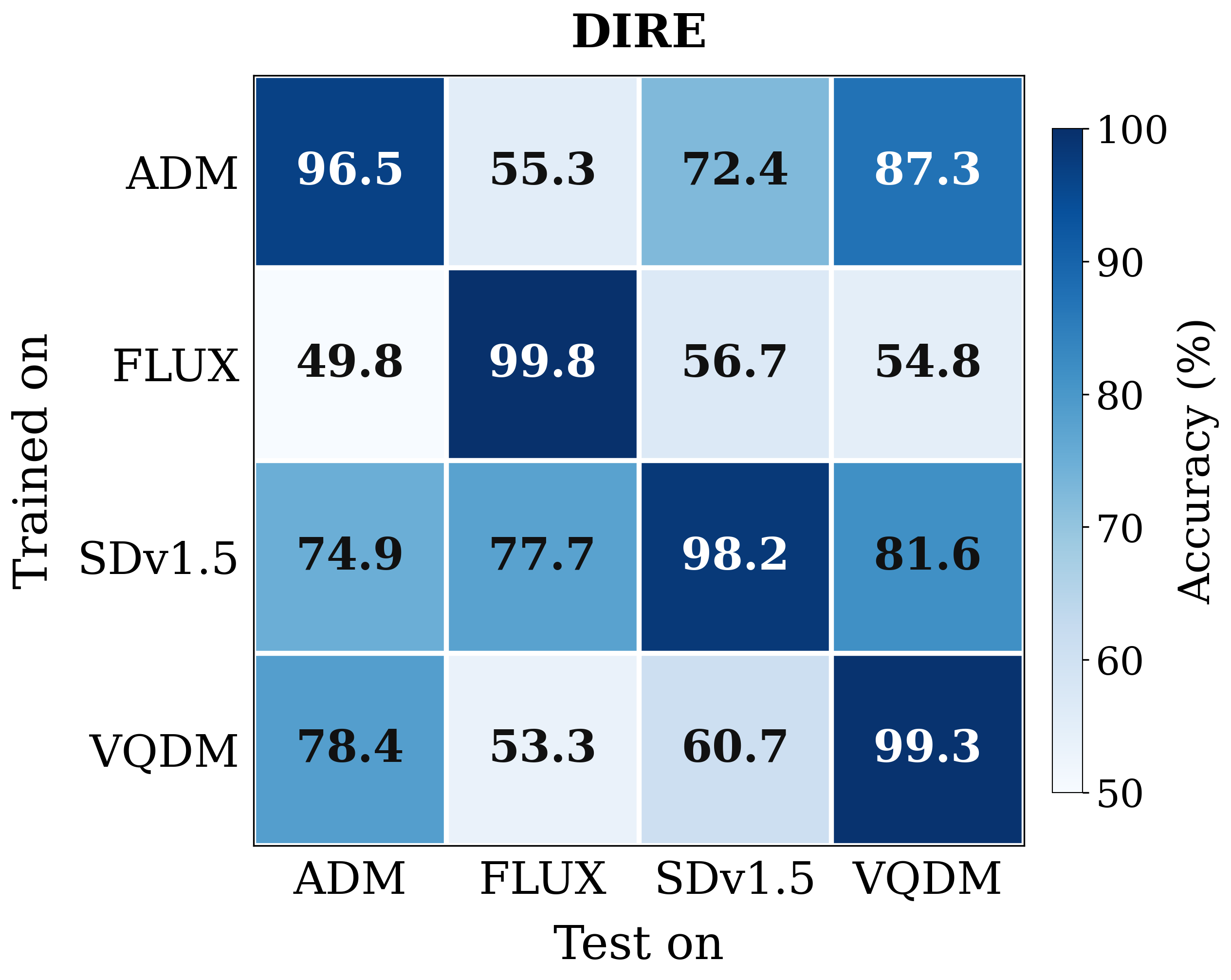}
    \caption{DIRE}
\end{subfigure}
\hfill
\begin{subfigure}{0.32\linewidth}
    \includegraphics[width=\linewidth]{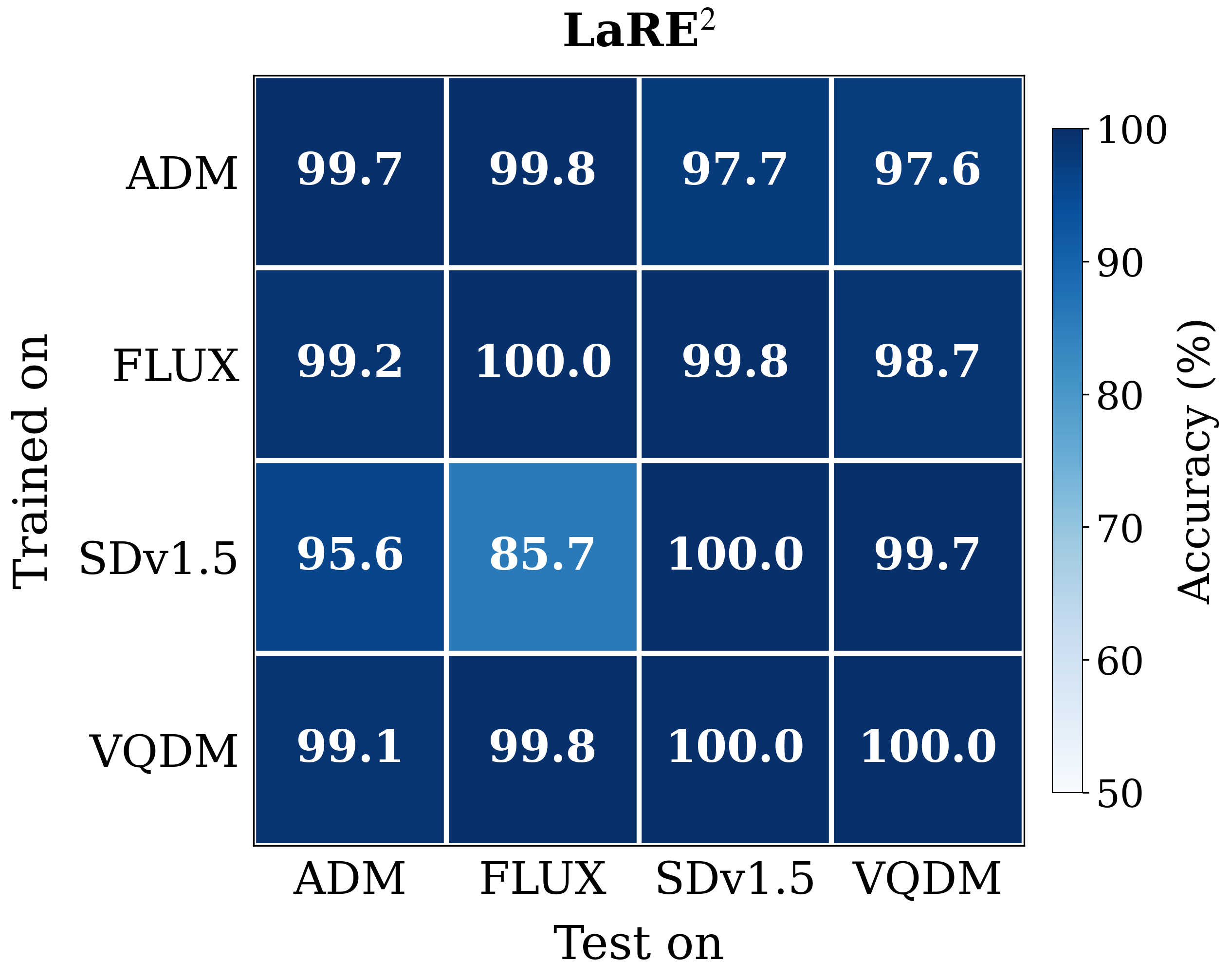}
    \caption{LaRE$^2$}
\end{subfigure}
\hfill
\begin{subfigure}{0.32\linewidth}
    \includegraphics[width=\linewidth]{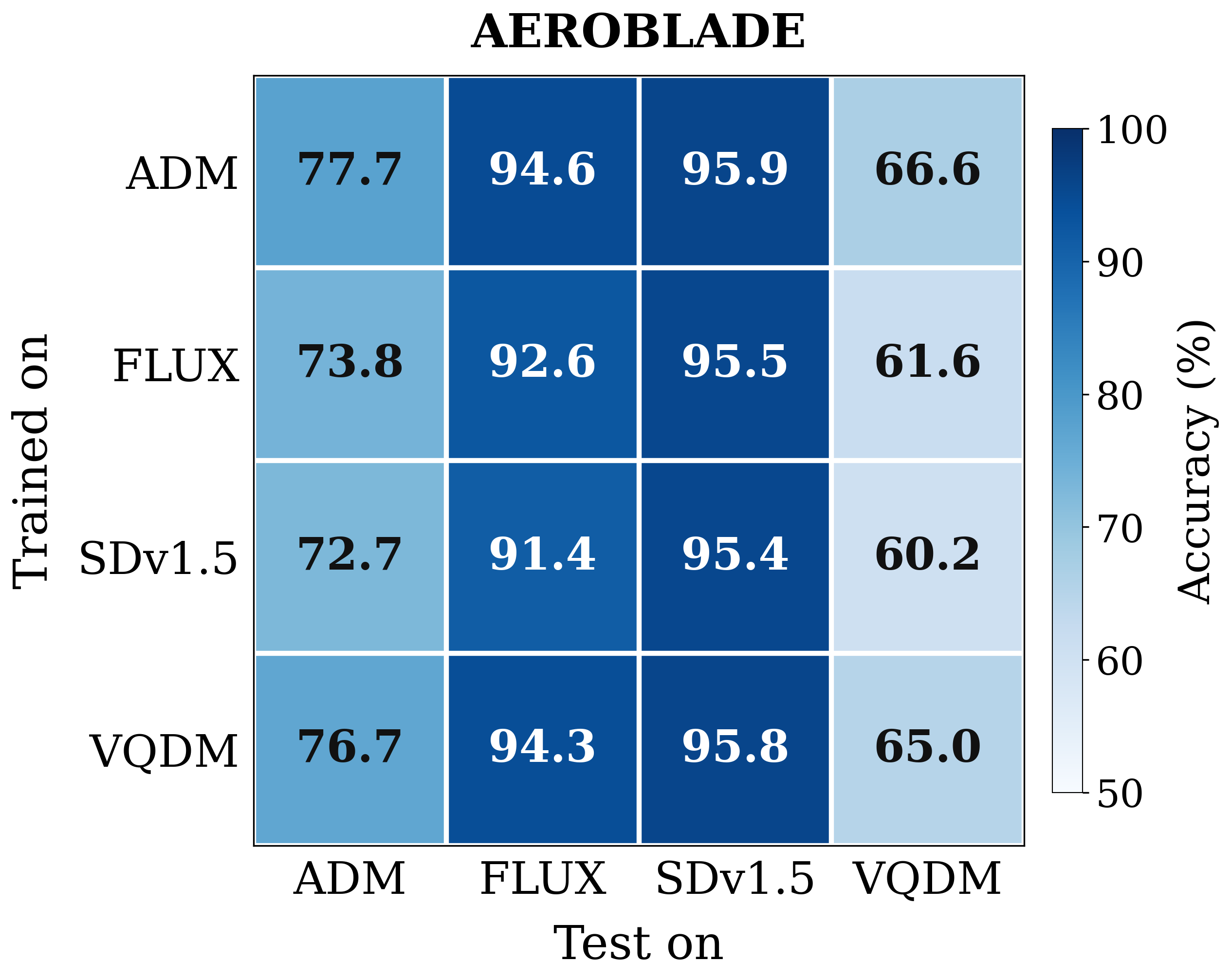}
    \caption{AEROBLADE}
\end{subfigure}
\caption{Detection performance under benign conditions. Each heatmap visualizes the accuracy (\%) when classifiers trained on one generator (rows) are evaluated on images from another (columns).}
\label{fig:detection_performance}
\end{figure}

As illustrated in Fig.~\ref{fig:detection_performance}, all three methods fundamentally rely on reconstruction error as the primary discriminative signal. Specifically, DIRE (Fig.~\ref{fig:detection_performance}a) achieves near-perfect accuracy on in-distribution data but suffers from significant degradation when encountering unseen architectures (e.g., FLUX), indicating a tendency to overfit to specific generative artifacts. In contrast, LaRE$^2$ (Fig.~\ref{fig:detection_performance}b) demonstrates superior generalization capabilities by extracting robust features from the latent space. AEROBLADE (Fig.~\ref{fig:detection_performance}c) maintains stable cross-generator performance due to its training-free nature. Overall, this heavy reliance on reconstruction consistency yields high accuracy on benign data but provides insufficient robustness when the reconstruction distance is intentionally manipulated, leading to the adversarial vulnerabilities analyzed in the main text.

% \subsection{Adversarial Attack Experimental Setup}
% \label{subsec:appendix_setup}

% To ensure reproducibility and a rigorous evaluation, we adhere to standard adversarial protocols with the following detailed configurations:

% \begin{itemize}[leftmargin=*, noitemsep]
%     \item \textbf{Dataset Construction:} We construct a comprehensive benchmark consisting of four distinct evaluation subsets. Each subset comprises 10,000 images, balanced between 5,000 real images (sourced from ImageNet~\cite{deng2009imagenet}) and 5,000 generated images sampled from one of the following four representative models: ADM~\cite{dhariwal2021diffusion}, SDv1.5~\cite{rombach2022high}, FLUX~\cite{blackforest2024flux}, and VQDM~\cite{gu2022vector}.

%     \item \textbf{Detection Models:} We evaluate three representative detectors: DIRE (pixel-space reconstruction), LaRE$^2$ (latent-space reconstruction), and AEROBLADE (training-free reconstruction). All models are rigorously calibrated to achieve optimal detection performance on benign data prior to adversarial evaluation.

%     \item \textbf{Attack Configuration:} We utilize the Auto-Projected Gradient Descent (APGD)~\cite{croce2020reliable} component from the AutoAttack suite, a gold standard for robustness evaluation. We enforce a strict $\ell_\infty$ constraint with a maximum perturbation budget of $\varepsilon = 8/255$. As a parameter-free method, APGD automatically adapts the step size over $100$ iterations, eliminating potential bias from manual hyperparameter tuning.
% \end{itemize}

\subsection{Extended Adversarial Robustness Analysis}
\label{sec:appendix_adv_extended}

To provide a more comprehensive evaluation of reconstruction-based detectors, we conduct additional ablation studies focusing on the sensitivity to attack hyperparameters.

\begin{figure}[t]
    \centering
    \begin{subfigure}{0.24\linewidth}
        \includegraphics[width=\linewidth]{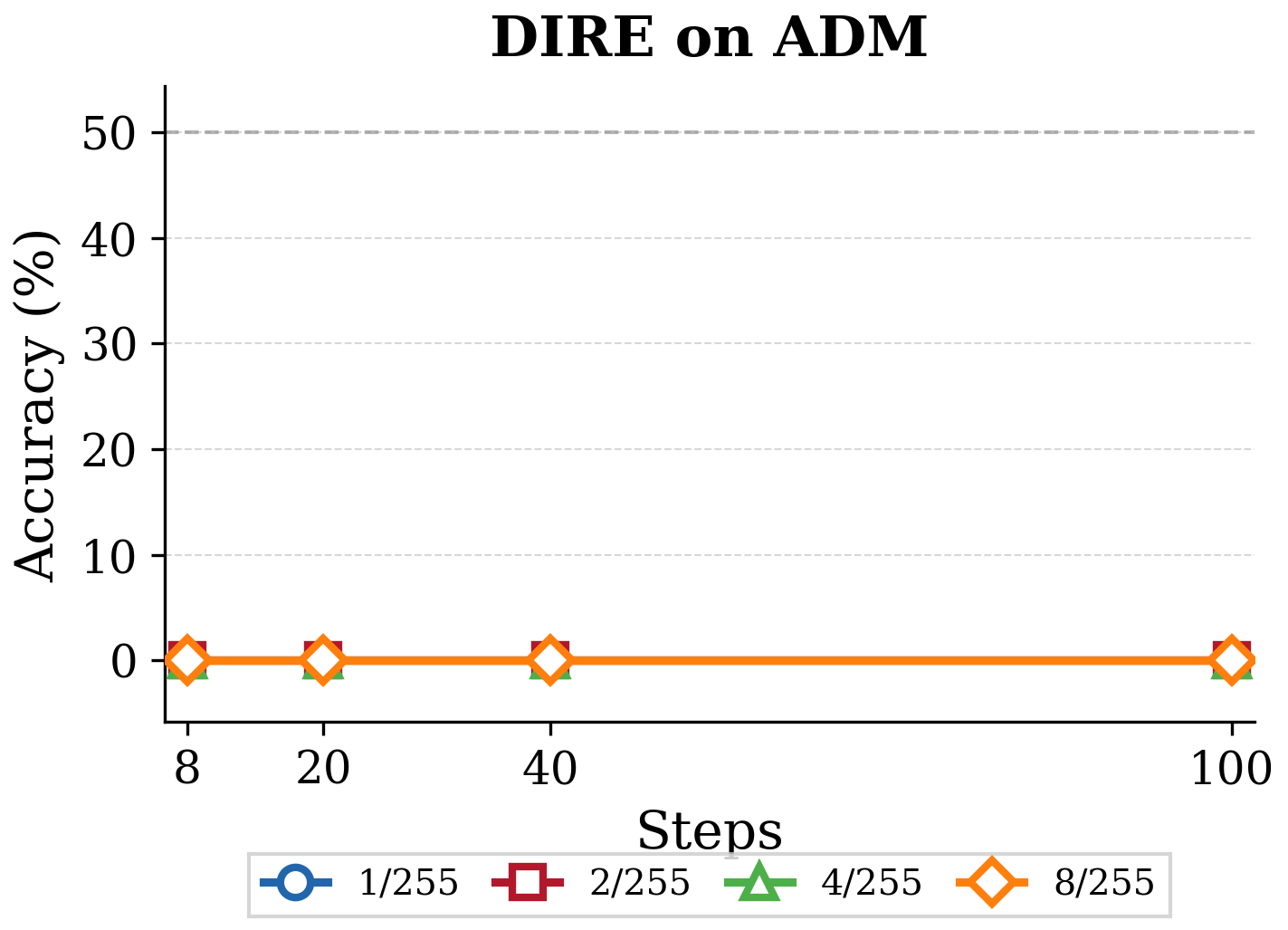} 
        \caption{ADM}
    \end{subfigure}
    \hfill
    \begin{subfigure}{0.24\linewidth}
        \includegraphics[width=\linewidth]{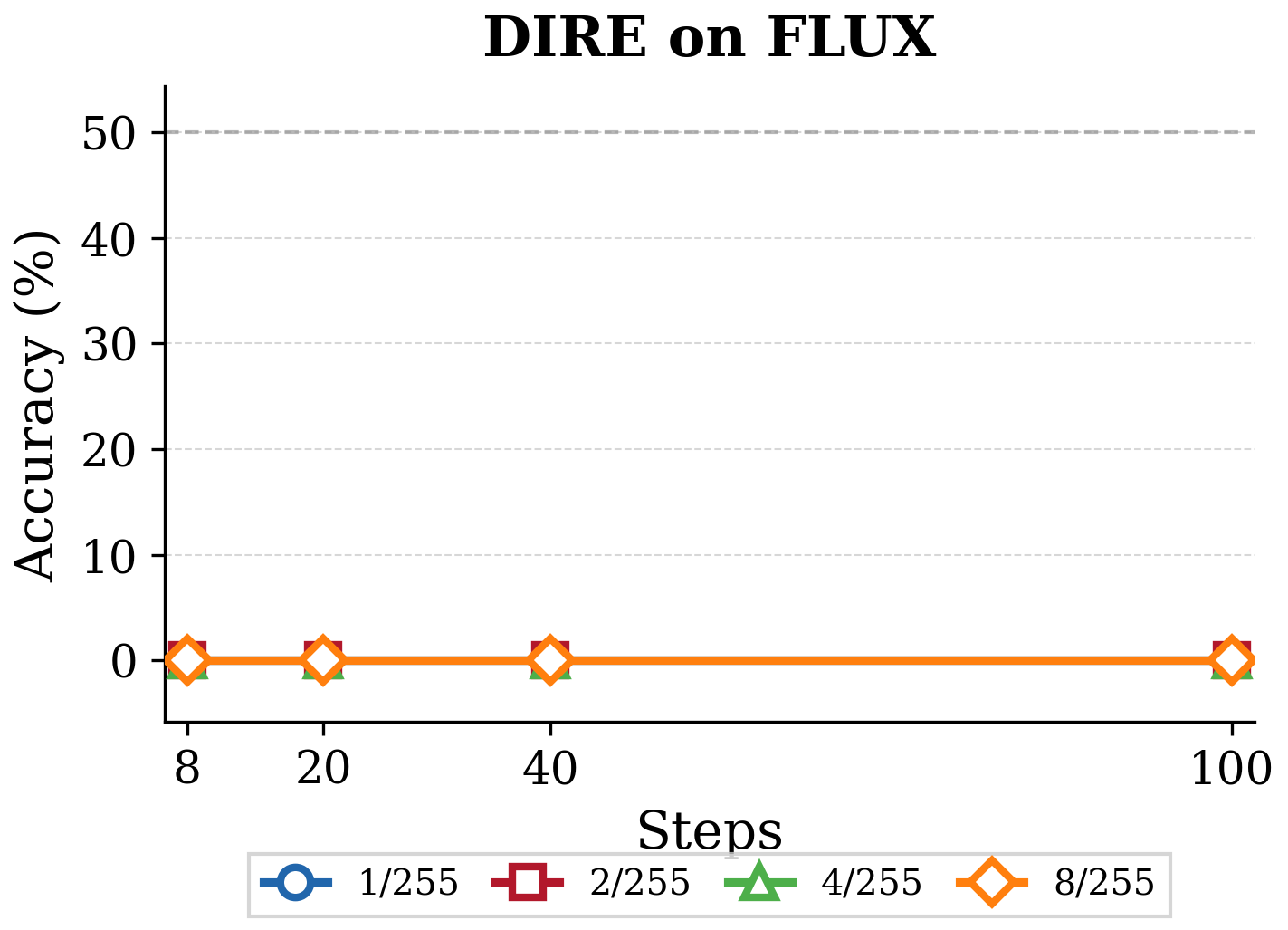} 
        \caption{FLUX}
    \end{subfigure}
    \hfill
    \begin{subfigure}{0.24\linewidth}
        \includegraphics[width=\linewidth]{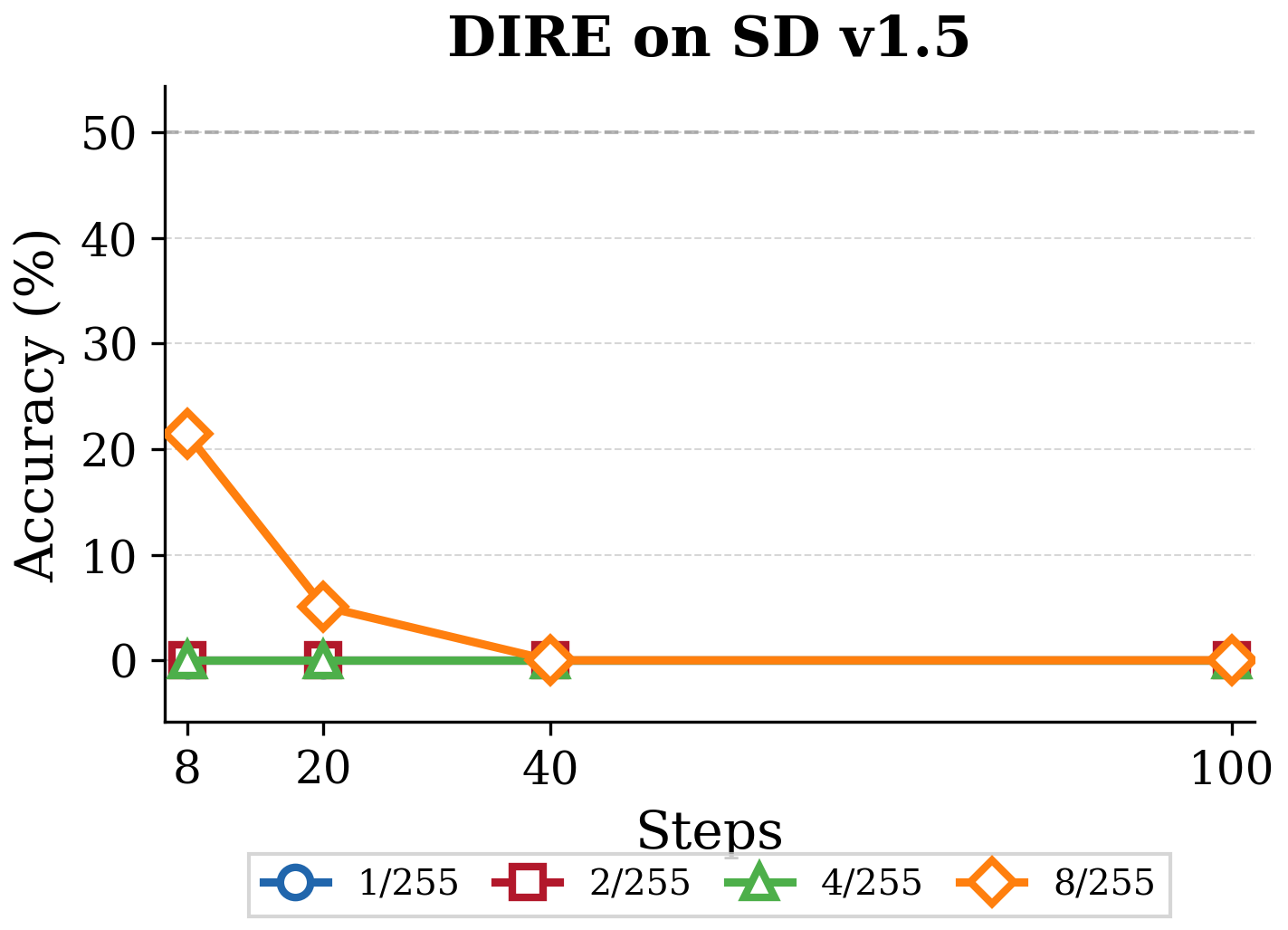} 
        \caption{SDv1.5}
    \end{subfigure}
    \hfill
    \begin{subfigure}{0.24\linewidth}
        \includegraphics[width=\linewidth]{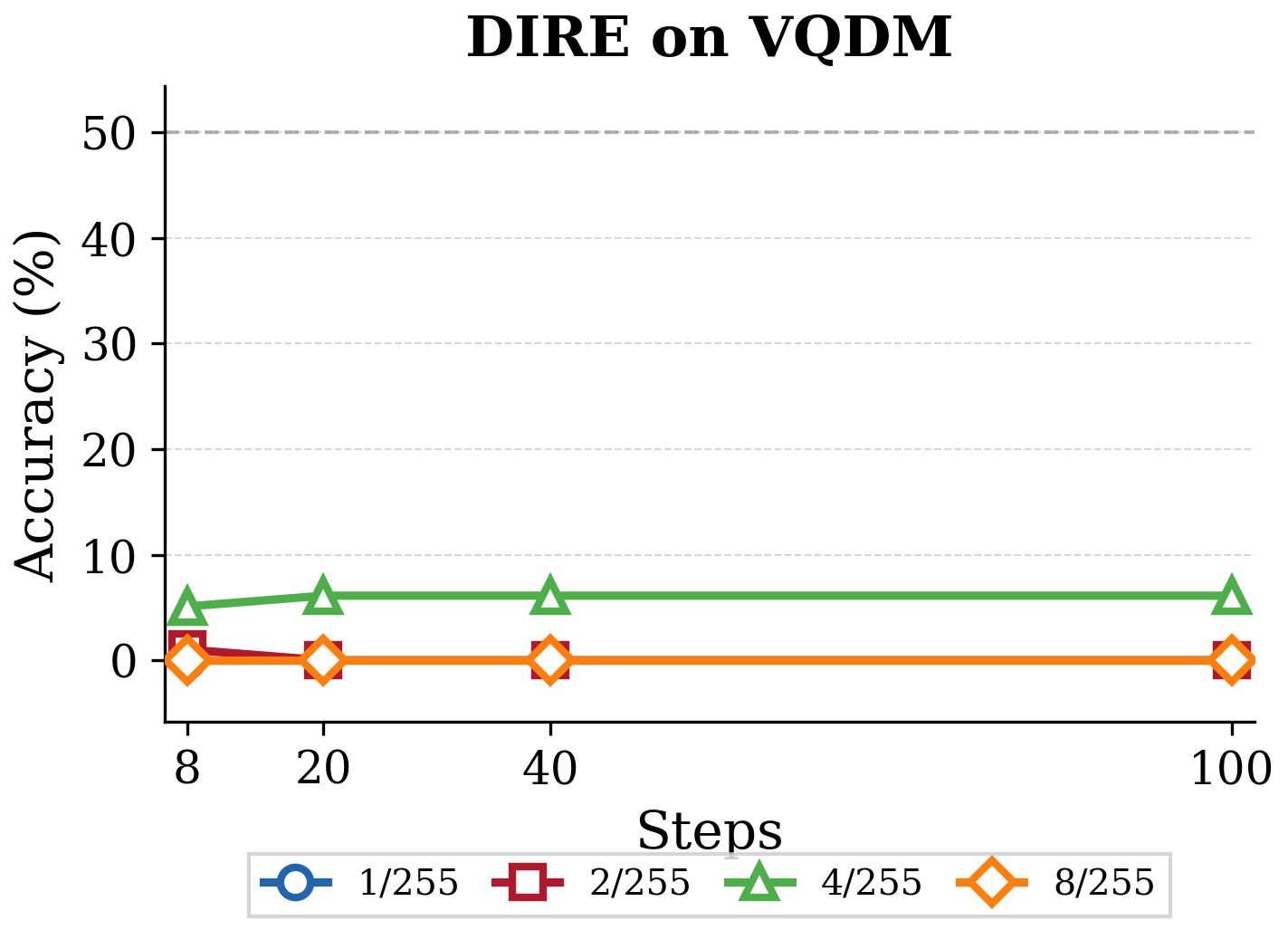} 
        \caption{VQDM}
    \end{subfigure}
    \caption{Ablation of attack hyperparameters on DIRE.}
    \label{fig:ablation_dire}
\end{figure}

\begin{figure}[t]
    \centering
    \begin{subfigure}{0.24\linewidth}
        \includegraphics[width=\linewidth]{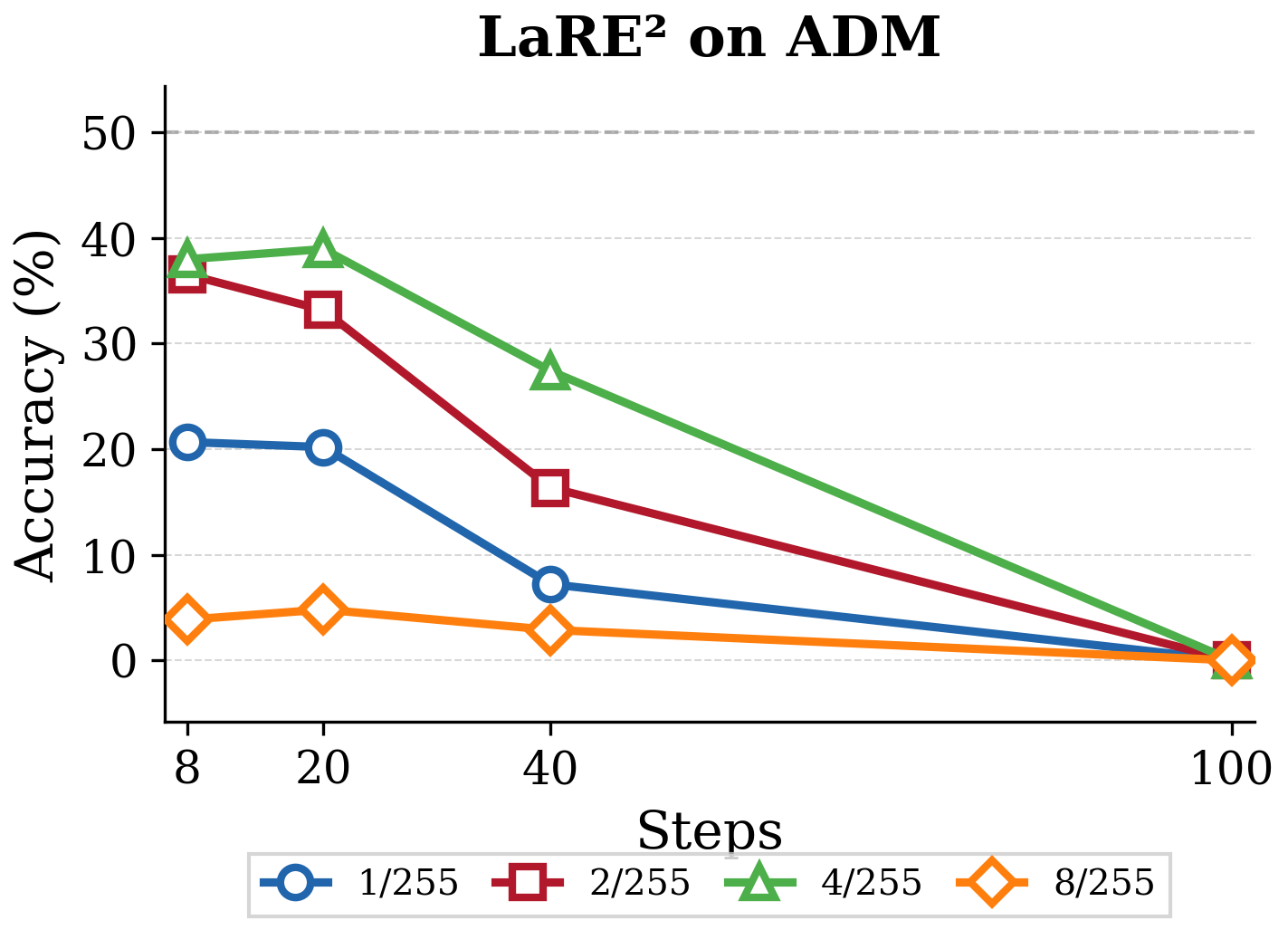}
        \caption{ADM}
    \end{subfigure}
    \hfill
    \begin{subfigure}{0.24\linewidth}
        \includegraphics[width=\linewidth]{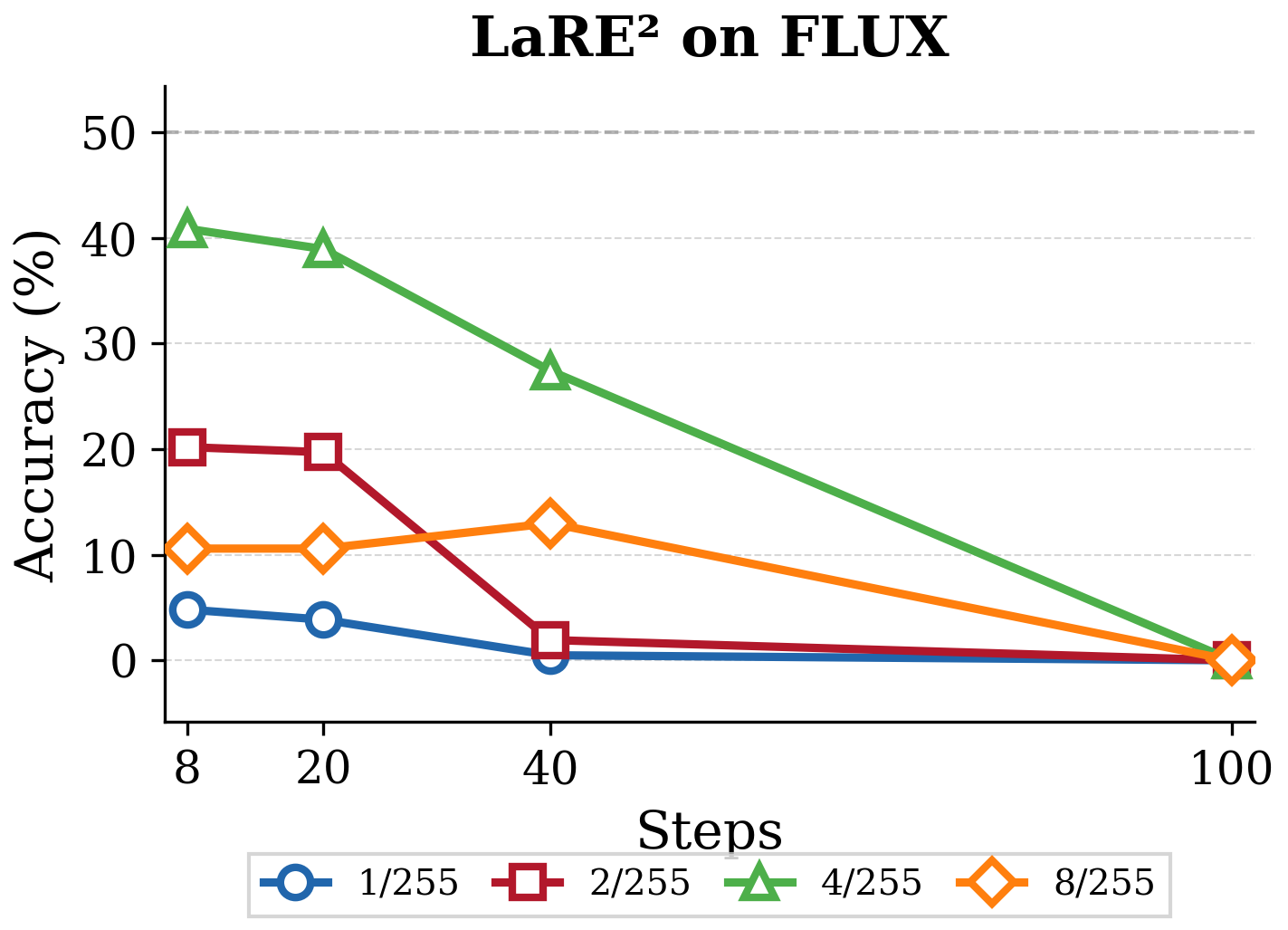}
        \caption{FLUX}
    \end{subfigure}
    \hfill
    \begin{subfigure}{0.24\linewidth}
        \includegraphics[width=\linewidth]{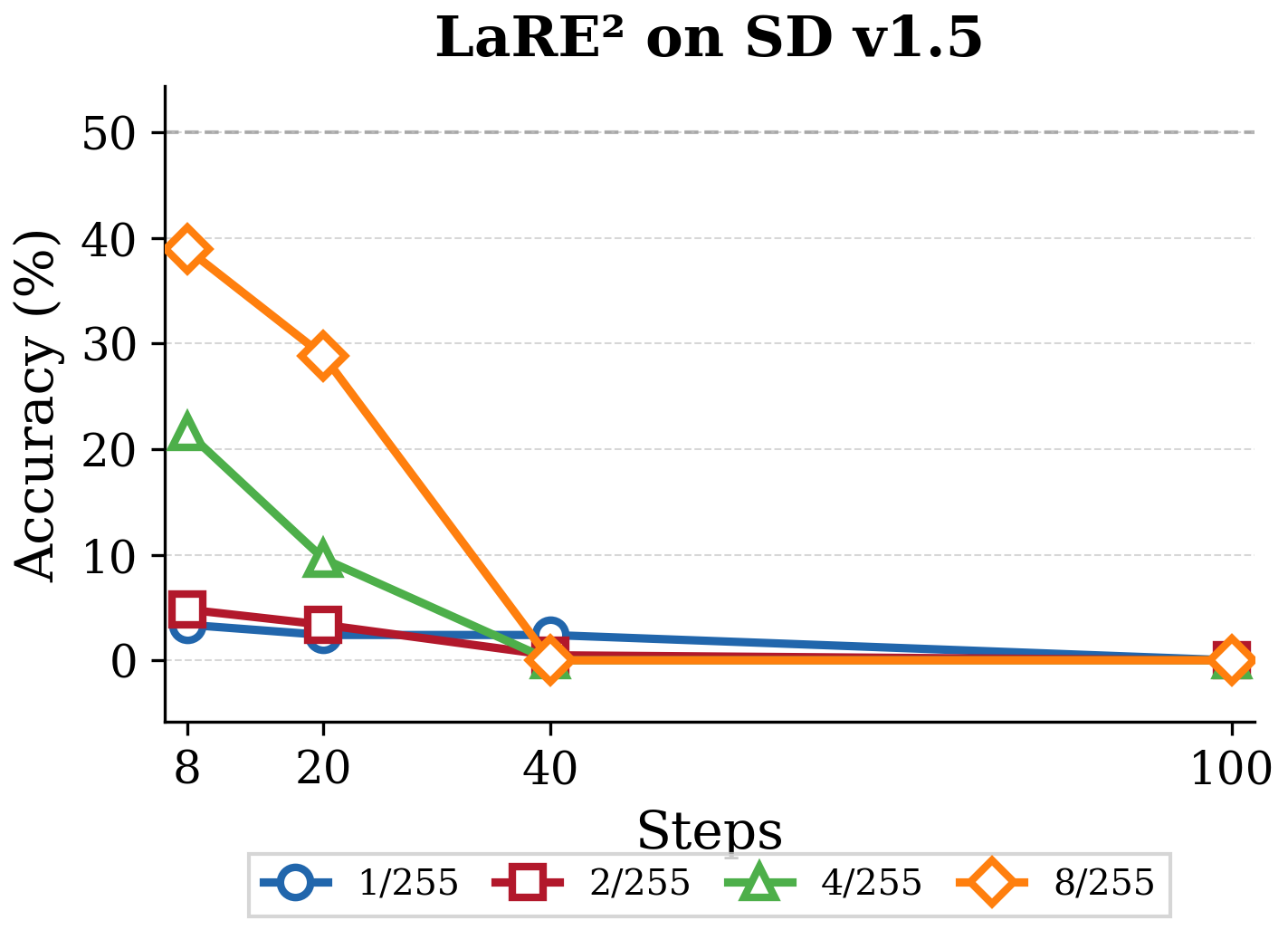}
        \caption{SDv1.5}
    \end{subfigure}
    \hfill
    \begin{subfigure}{0.24\linewidth}
        \includegraphics[width=\linewidth]{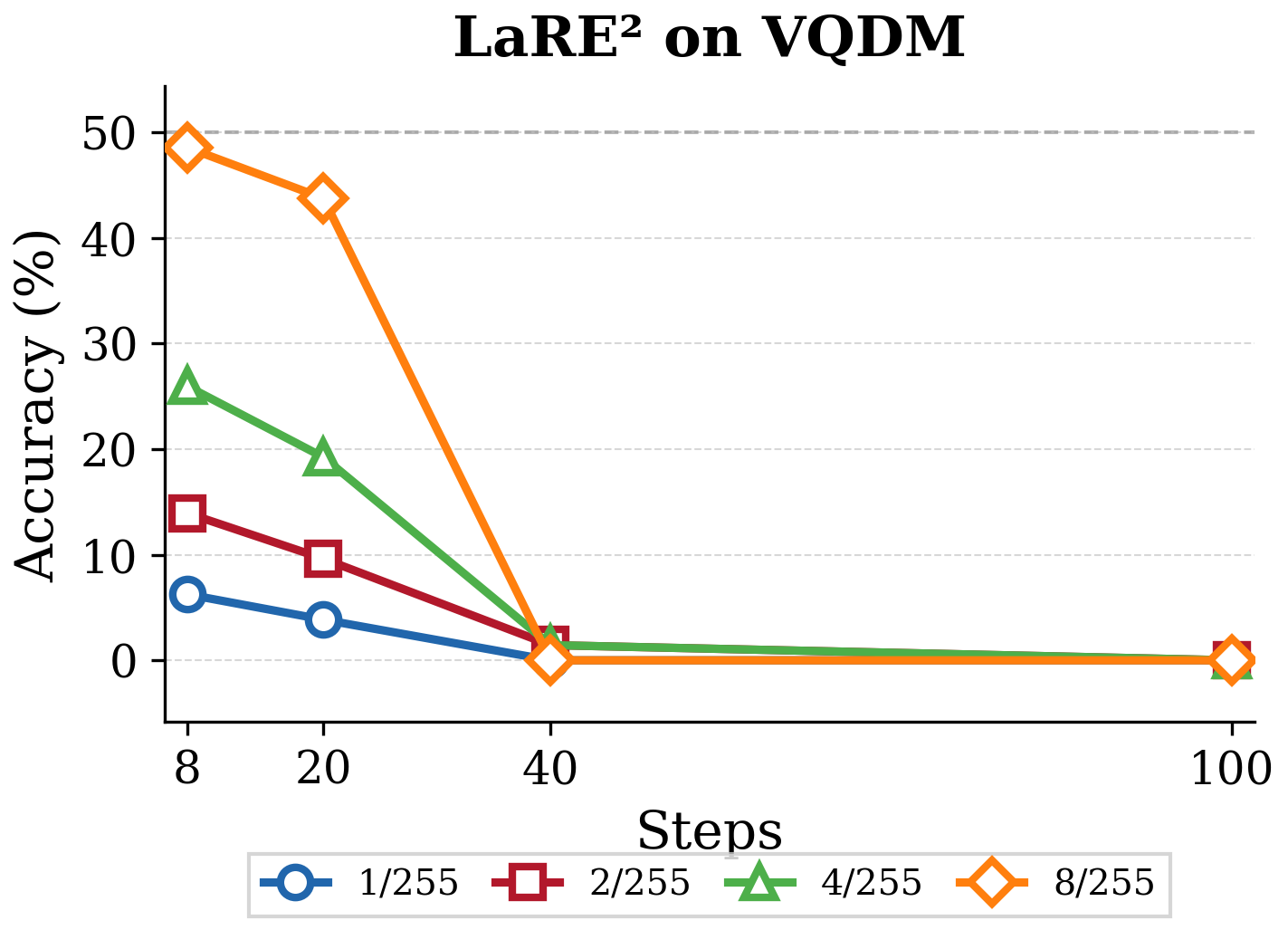}
        \caption{VQDM}
    \end{subfigure}
    \caption{Ablation of attack hyperparameters on LaRE$^2$.}
    \label{fig:ablation_lare}
\end{figure}

\begin{figure}[t]
    \centering
    \begin{subfigure}{0.24\linewidth}
        \includegraphics[width=\linewidth]{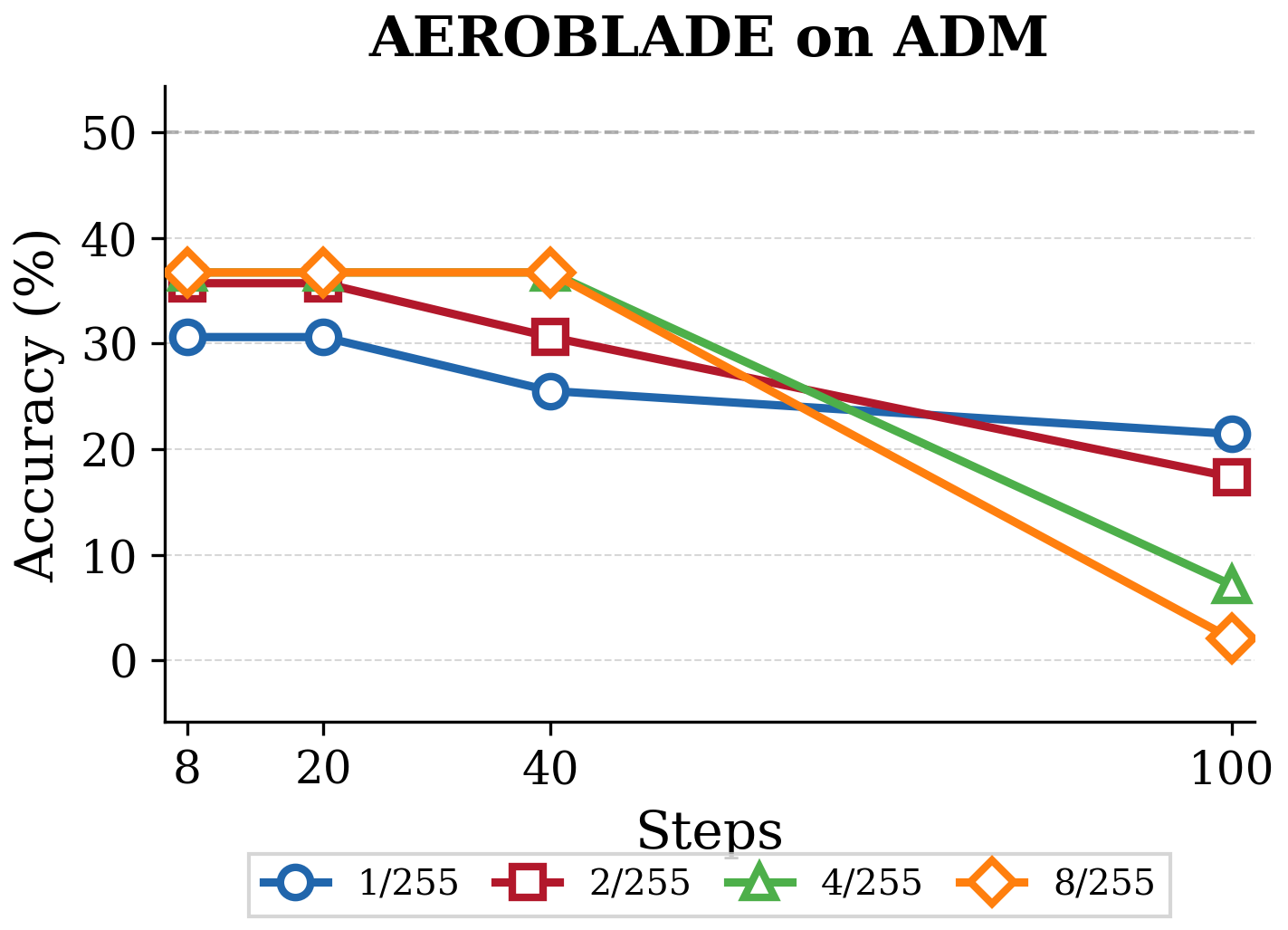} 
        \caption{ADM}
    \end{subfigure}
    \hfill
    \begin{subfigure}{0.24\linewidth}
        \includegraphics[width=\linewidth]{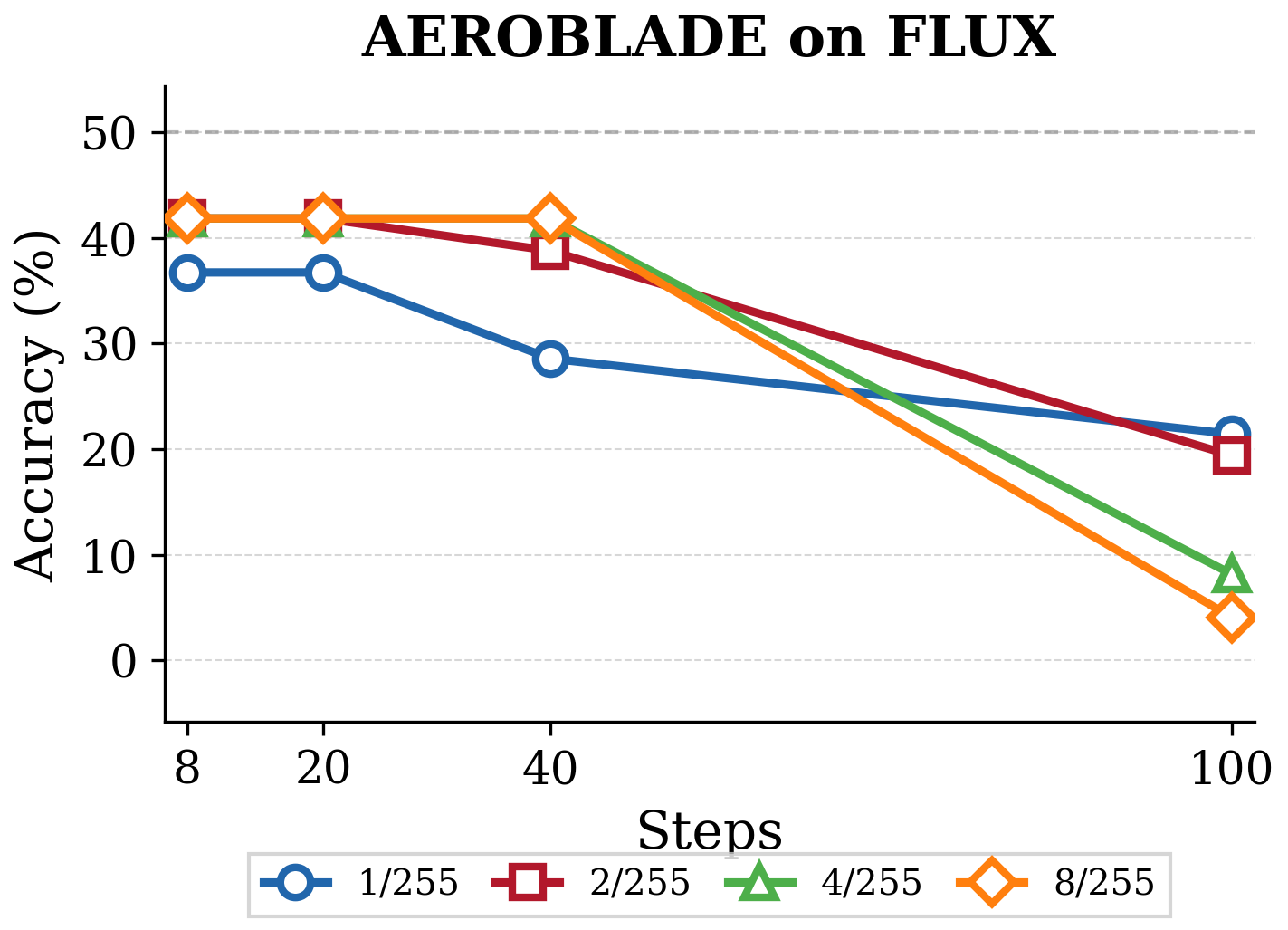} 
        \caption{FLUX}
    \end{subfigure}
    \hfill
    \begin{subfigure}{0.24\linewidth}
        \includegraphics[width=\linewidth]{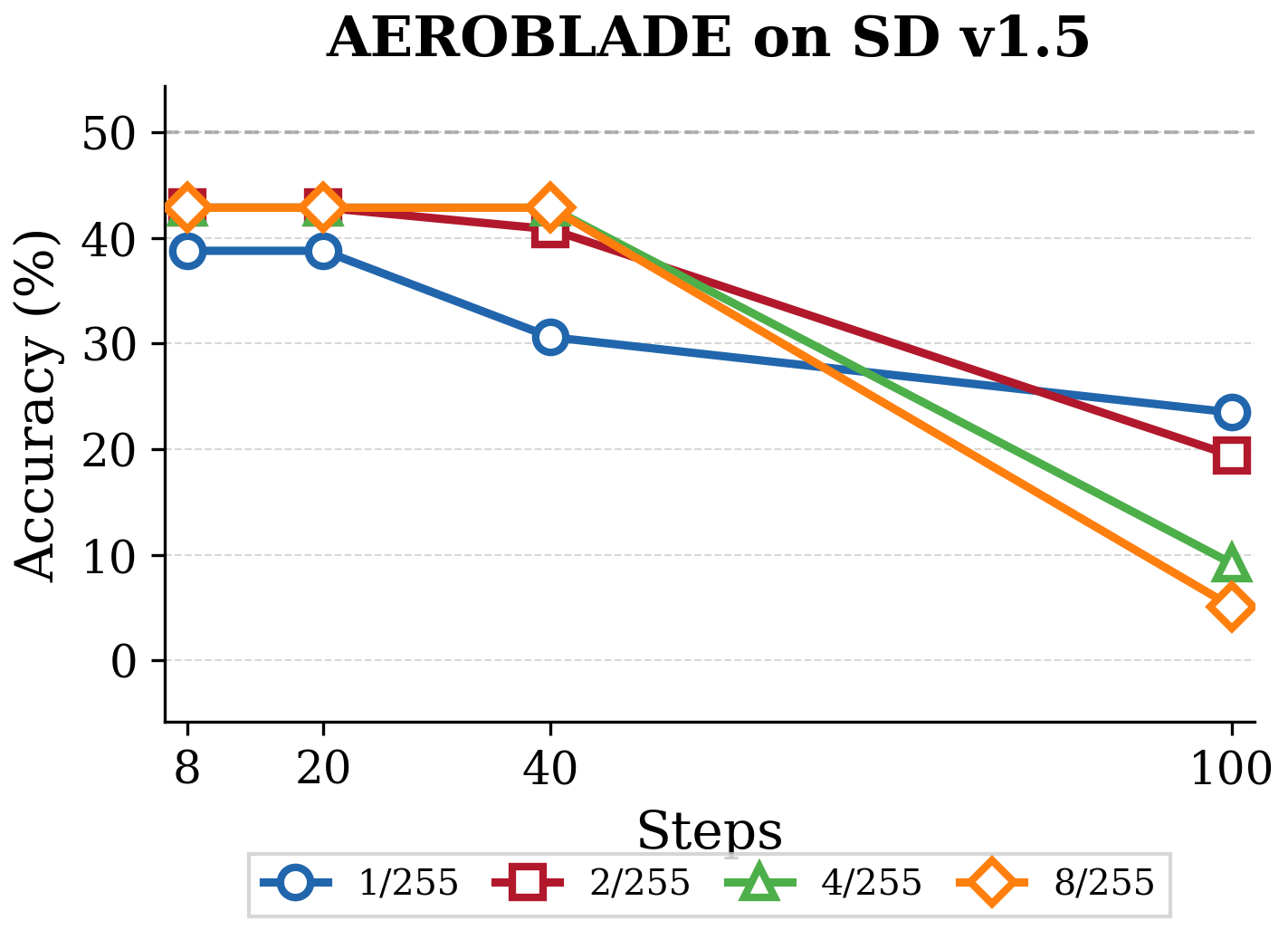} 
        \caption{SDv1.5}
    \end{subfigure}
    \hfill
    \begin{subfigure}{0.24\linewidth}
        \includegraphics[width=\linewidth]{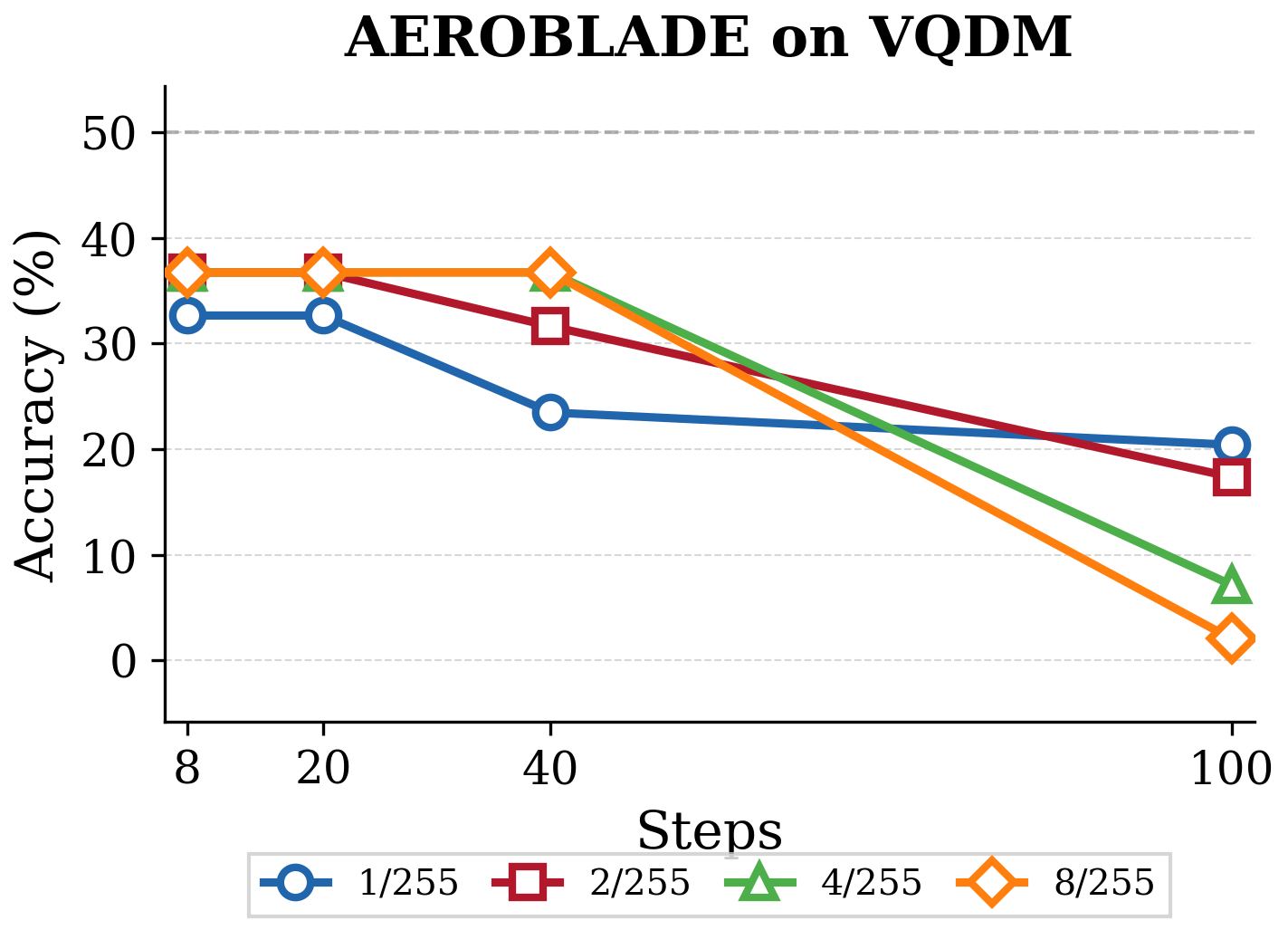} 
        \caption{VQDM}
    \end{subfigure}
    \caption{Ablation of attack hyperparameters on AEROBLADE.}
    \label{fig:ablation_aero}
\end{figure}

\paragraph{Impact of Attack Hyperparameters.}
We evaluate detection accuracy across various perturbation budgets $\varepsilon$ and optimization steps $T$. As shown in Fig.~\ref{fig:ablation_dire}--\ref{fig:ablation_aero}, the methods exhibit distinct behaviors. AEROBLADE demonstrates a \textit{monotonic decay}: while its performance drops under stronger attacks, it retains partial robustness ($\sim 20\%$) at lower budgets ($\varepsilon=1/255$) even after 100 steps. In contrast, LaRE$^2$ shows inconsistent robustness across datasets; while it offers \textit{transient resistance} on ADM and FLUX at lower budgets, it suffers rapid degradation on SDv1.5 and VQDM, eventually collapsing to near $0\%$ across all benchmarks as optimization proceeds. Finally, DIRE proves to be the most fragile, suffering a complete and instantaneous collapse upon attack initiation.

% --- DIRE Ablation ---

% --- LaRE^2 Ablation ---

% --- AeroBlade Ablation ---

\paragraph{Robustness: Transferred Attacks vs. Random Noise.}
\label{subsec:appendix_transfer_vs_random}

To contextualize the threat level of our proposed attacks, we compare the destructive capability of fully black-box transfer attacks (i.e., cross-generator and cross-method) against the injection of random Gaussian noise. 

\begin{figure}[t]
    \centering
    \begin{subfigure}{0.24\linewidth}
        \includegraphics[width=\linewidth]{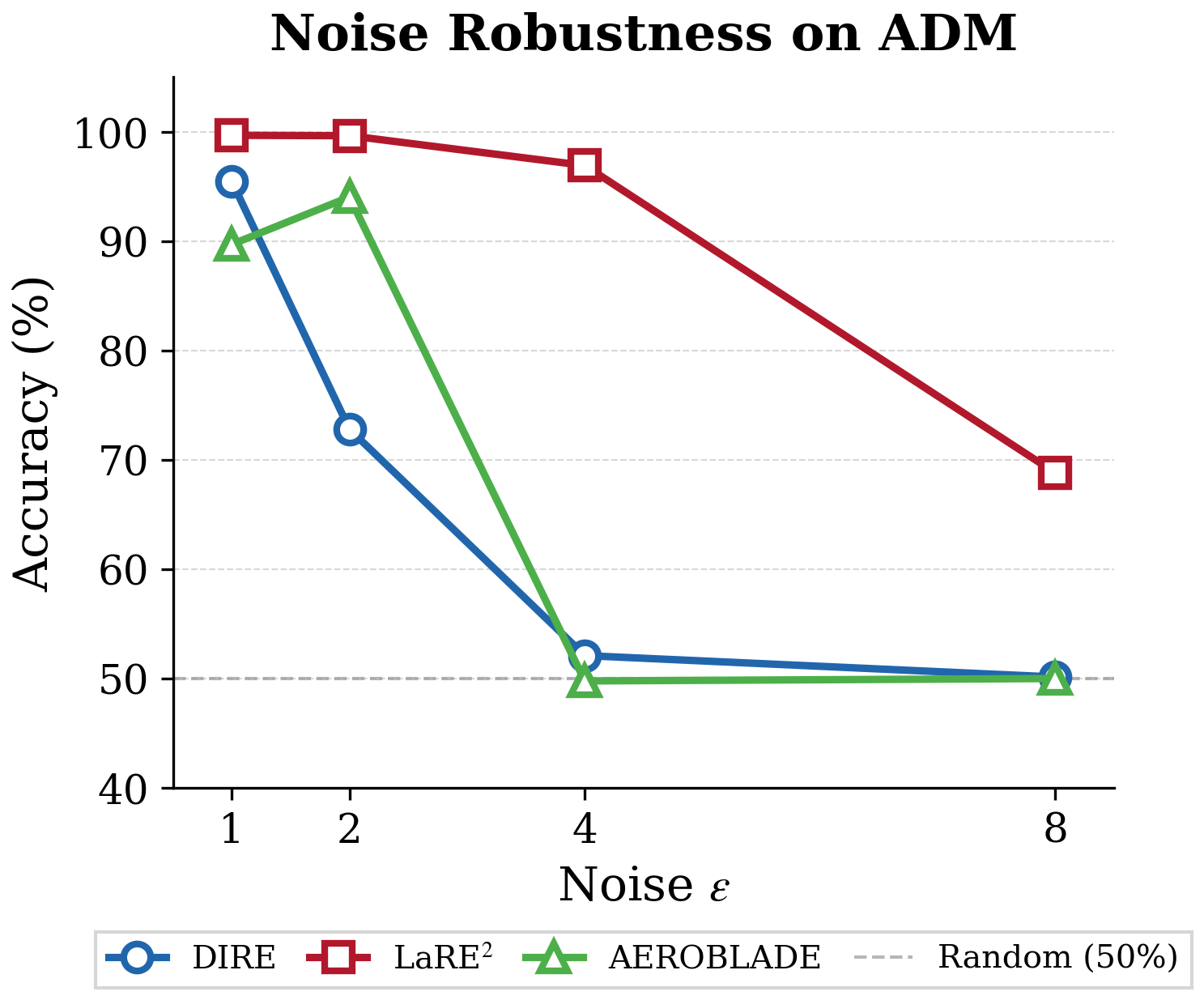} 
        \caption{ADM}
    \end{subfigure}
    \hfill
    \begin{subfigure}{0.24\linewidth}
        \includegraphics[width=\linewidth]{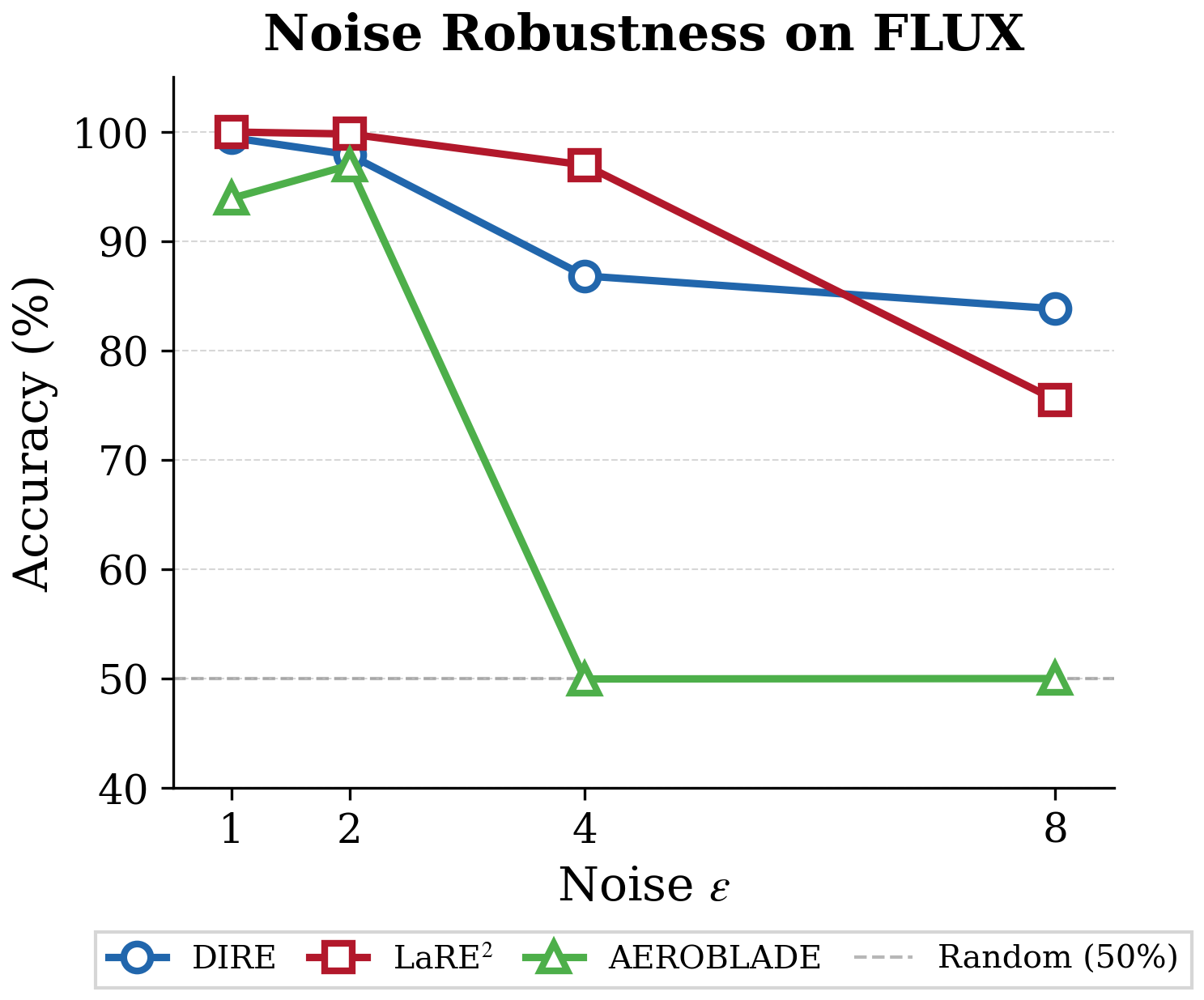} 
        \caption{FLUX}
    \end{subfigure}
    \hfill
    \begin{subfigure}{0.24\linewidth}
        \includegraphics[width=\linewidth]{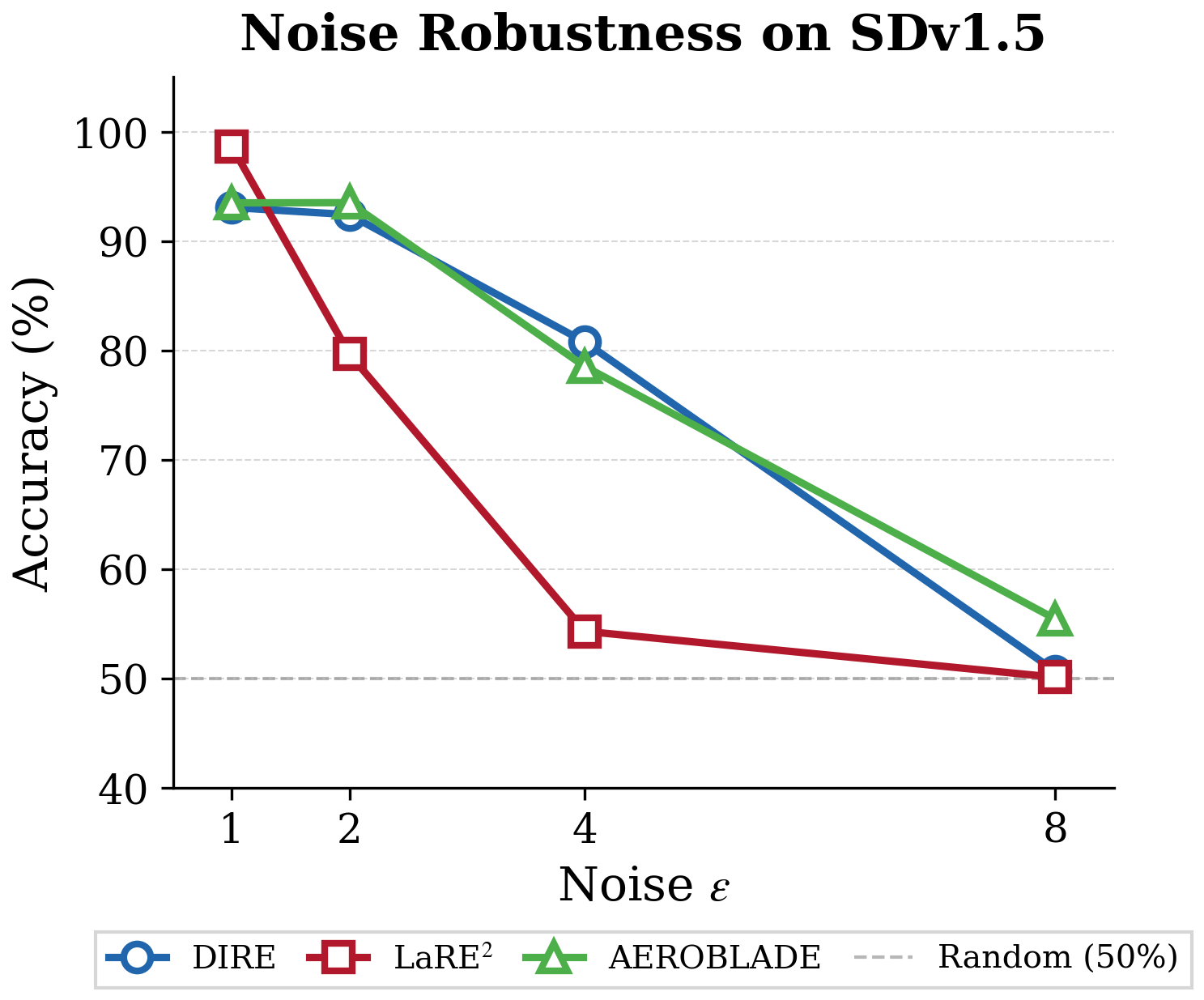} 
        \caption{SDv1.5}
    \end{subfigure}
    \hfill
    \begin{subfigure}{0.24\linewidth}
        \includegraphics[width=\linewidth]{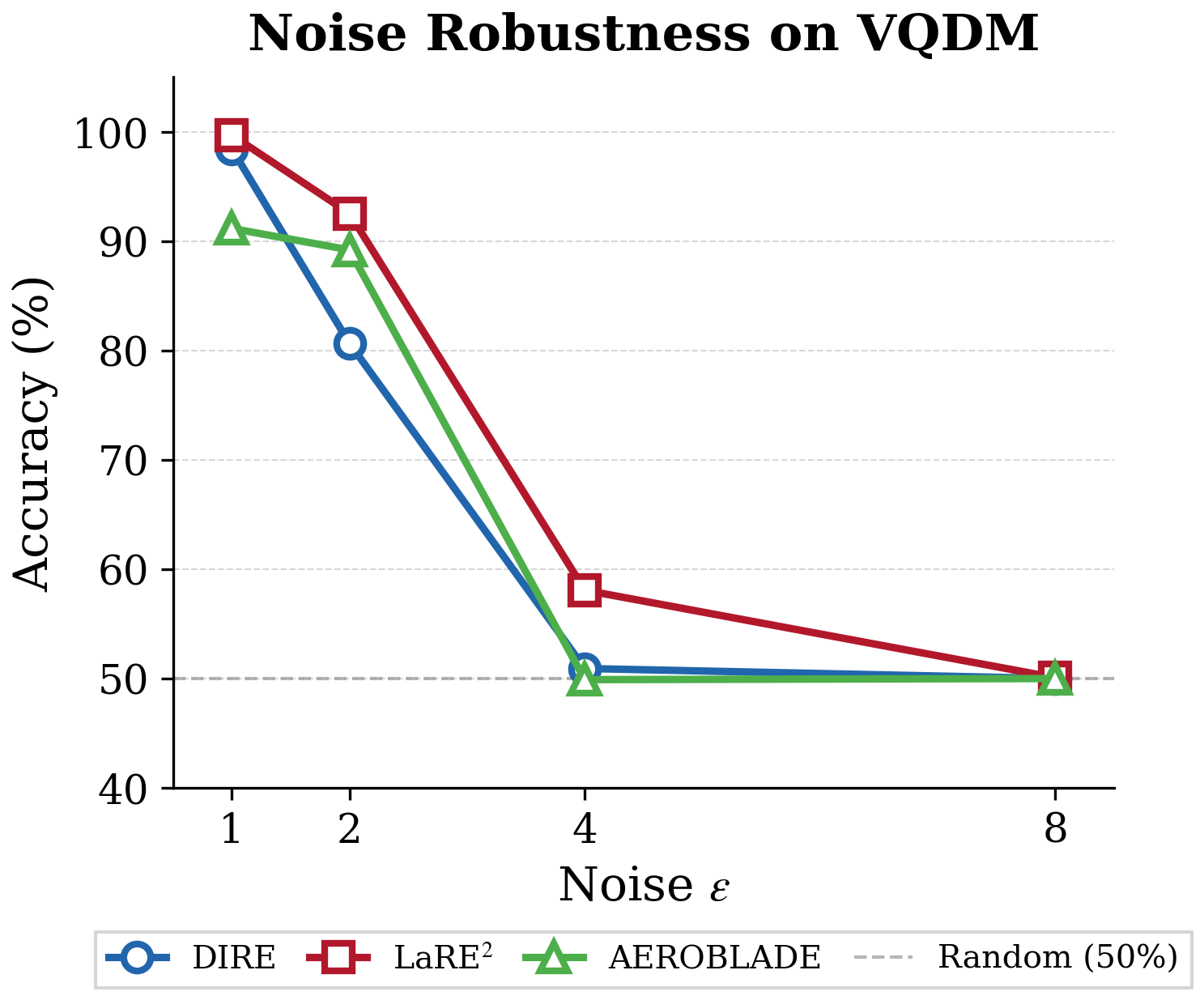} 
        \caption{VQDM}
    \end{subfigure}
    \caption{Detection accuracy under random Gaussian noise. }
\label{fig:random_noise}
\end{figure}

As illustrated in Fig.~\ref{fig:random_noise}, while random noise bounded by the maximum budget ($\varepsilon=8/255$) degrades detection performance, it frequently fails to completely neutralize the detectors. For instance, on the FLUX dataset, DIRE and LaRE$^2$ maintain significant discriminative capabilities, exhibiting robust accuracies of approximately $85\%$ and $75\%$, respectively, under maximum random noise. This indicates that random noise merely introduces arbitrary signal corruption, which advanced detectors can often partially withstand.

In stark contrast, our fully black-box transfer attacks (as previously shown in Table~\ref{tab:cross_both}) universally paralyze the target detectors across nearly all generator-method combinations, consistently driving the post-attack accuracy down to the random-guessing level of $\approx 50\%$. 

The fact that our transferred perturbations can easily neutralize even robust architectures—such as LaRE$^2$ on FLUX, where random noise has limited impact—provides crucial evidence for the efficacy of our attack. It demonstrates that our method does not merely degrade image quality with stochastic artifacts. Instead, the transferred perturbations \textit{directly target and dismantle} the underlying reconstruction consistency that these detectors fundamentally rely on. This targeted disruption of a shared mechanism is precisely what underpins the high transferability and potency of our attack.

\paragraph{Discussion.}
Collectively, these extended evaluations highlight a fundamental distinction between arbitrary signal corruption and targeted adversarial exploitation. The hyperparameter ablations reveal that while advanced architectures like LaRE$^2$ and AEROBLADE offer transient resistance at limited budgets, they inevitably succumb to sustained optimization, exposing the same underlying structural fragility as DIRE. More importantly, the stark contrast between our proposed attacks and random Gaussian noise confirms the targeted nature of our method. While random noise merely acts as a crude mask—often resisted by robust models or, at best, reducing performance to a random guess—our adversarial perturbations systematically dismantle the shared reconstructive priors. Ultimately, these results demonstrate that the vulnerability of reconstruction-based detectors is an inherent, universally exploitable flaw, rather than a simple sensitivity to input degradation.

\subsection{Full Results for Cross-Domain Black-Box Attacks}
\label{subsec:appendix_cross_both}
\begin{figure}[h]
\centering
\begin{subfigure}{0.32\linewidth}
    \includegraphics[width=\linewidth]{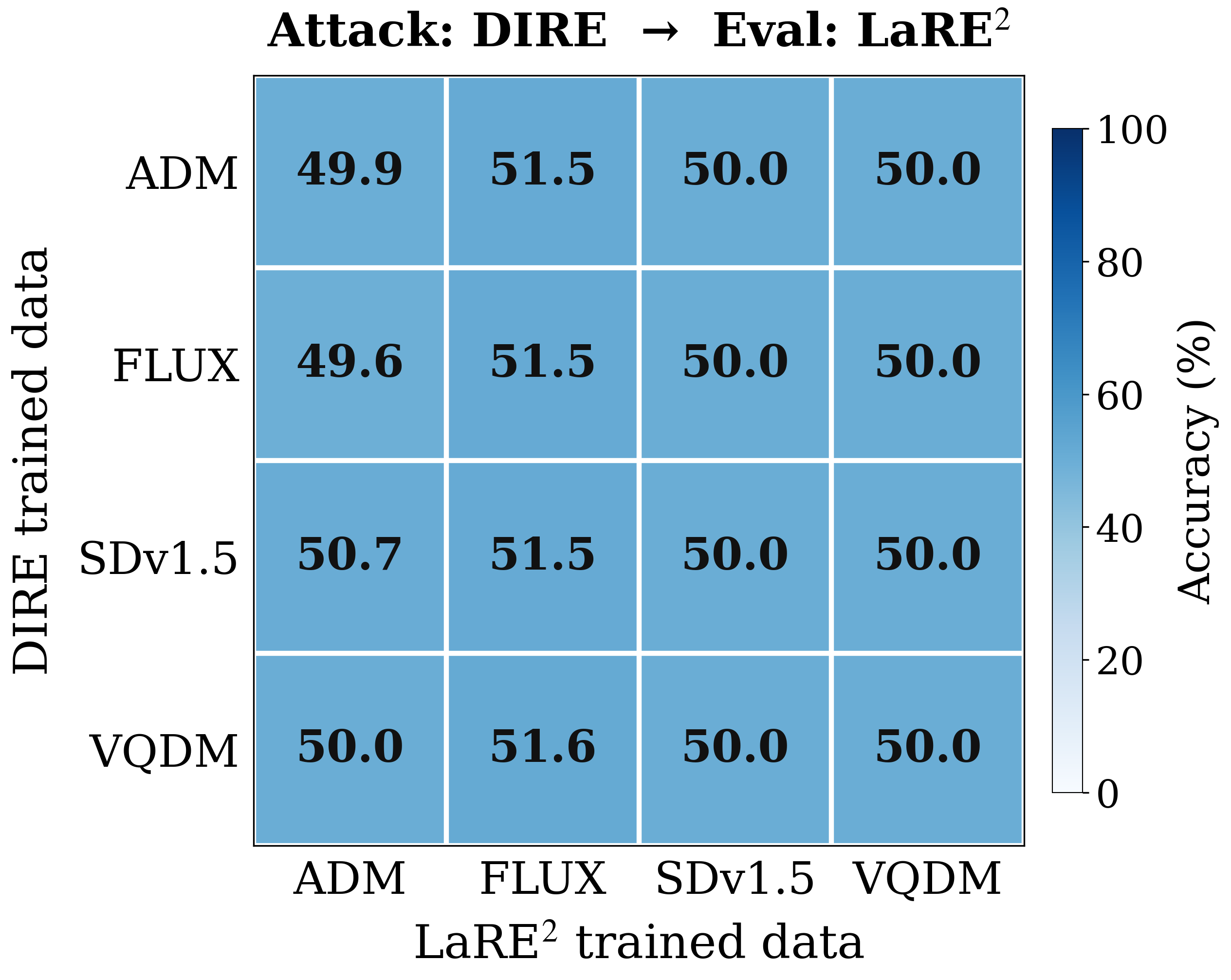}
    \caption{DIRE $\to$ LaRE$^2$}
\end{subfigure}
\hfill
\begin{subfigure}{0.32\linewidth}
    \includegraphics[width=\linewidth]{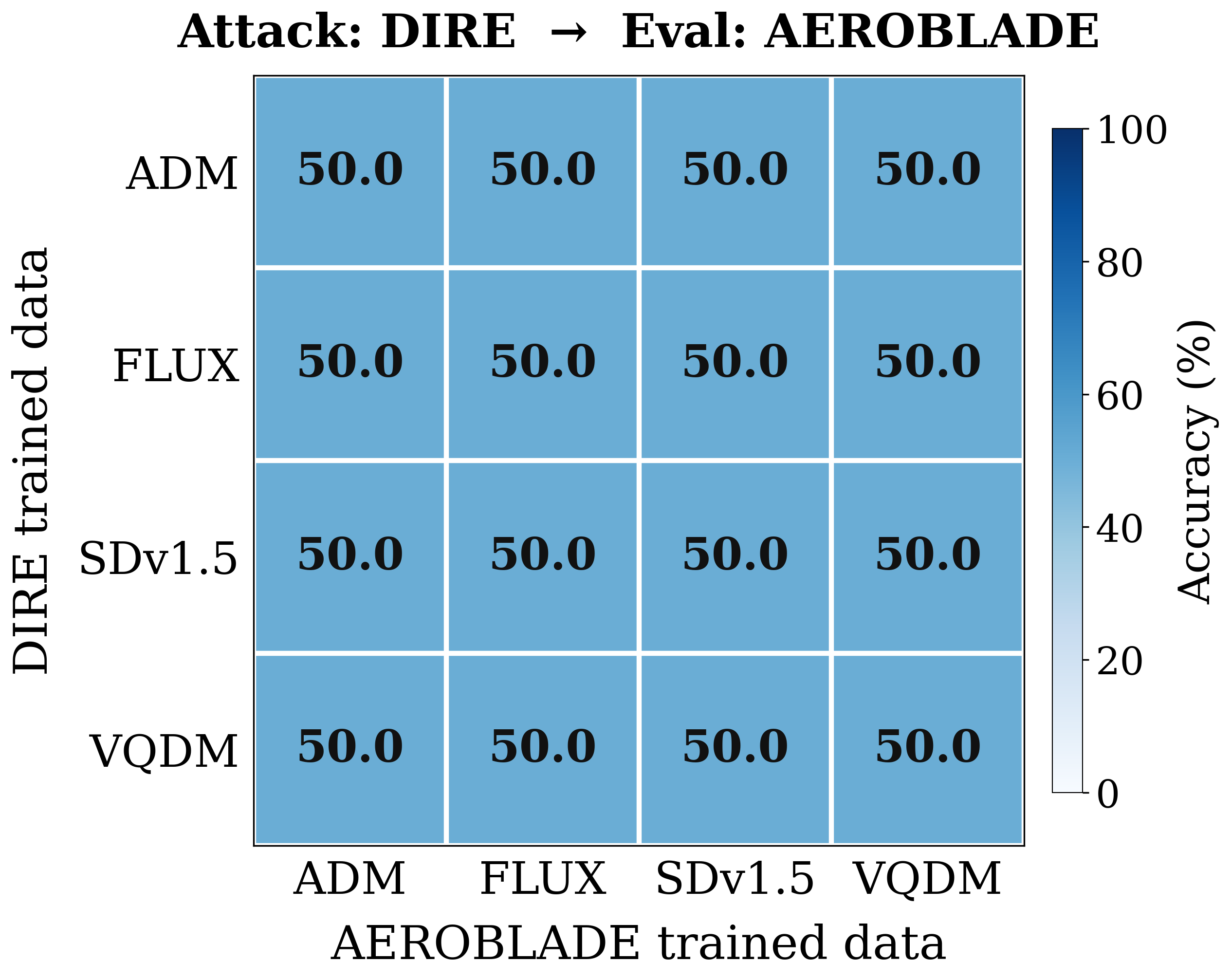}
    \caption{DIRE $\to$ AEROBLADE}
\end{subfigure}
\hfill
\begin{subfigure}{0.32\linewidth}
    \includegraphics[width=\linewidth]{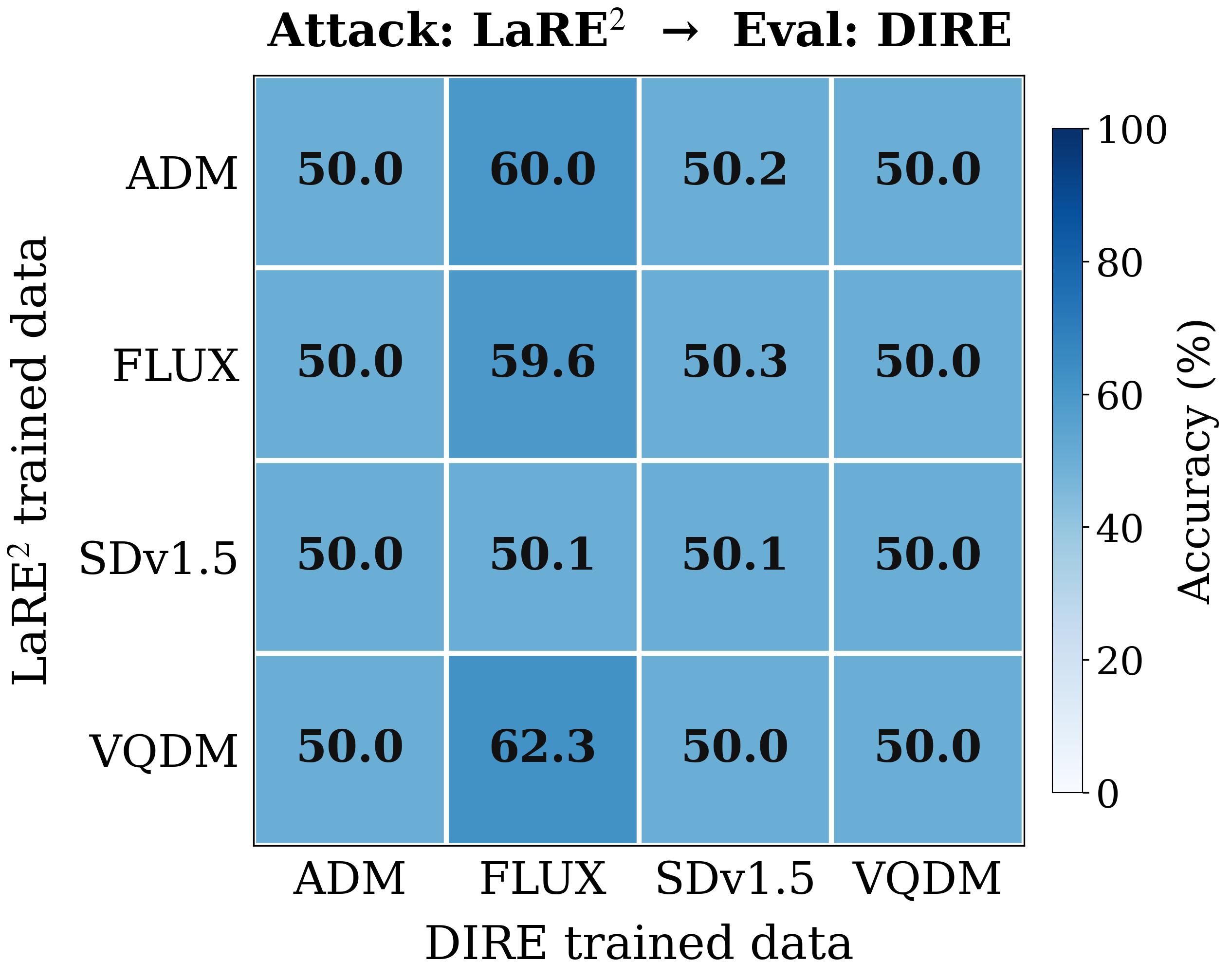}
    \caption{LaRE$^2$ $\to$ DIRE}
\end{subfigure}

\begin{subfigure}{0.32\linewidth}
    \includegraphics[width=\linewidth]{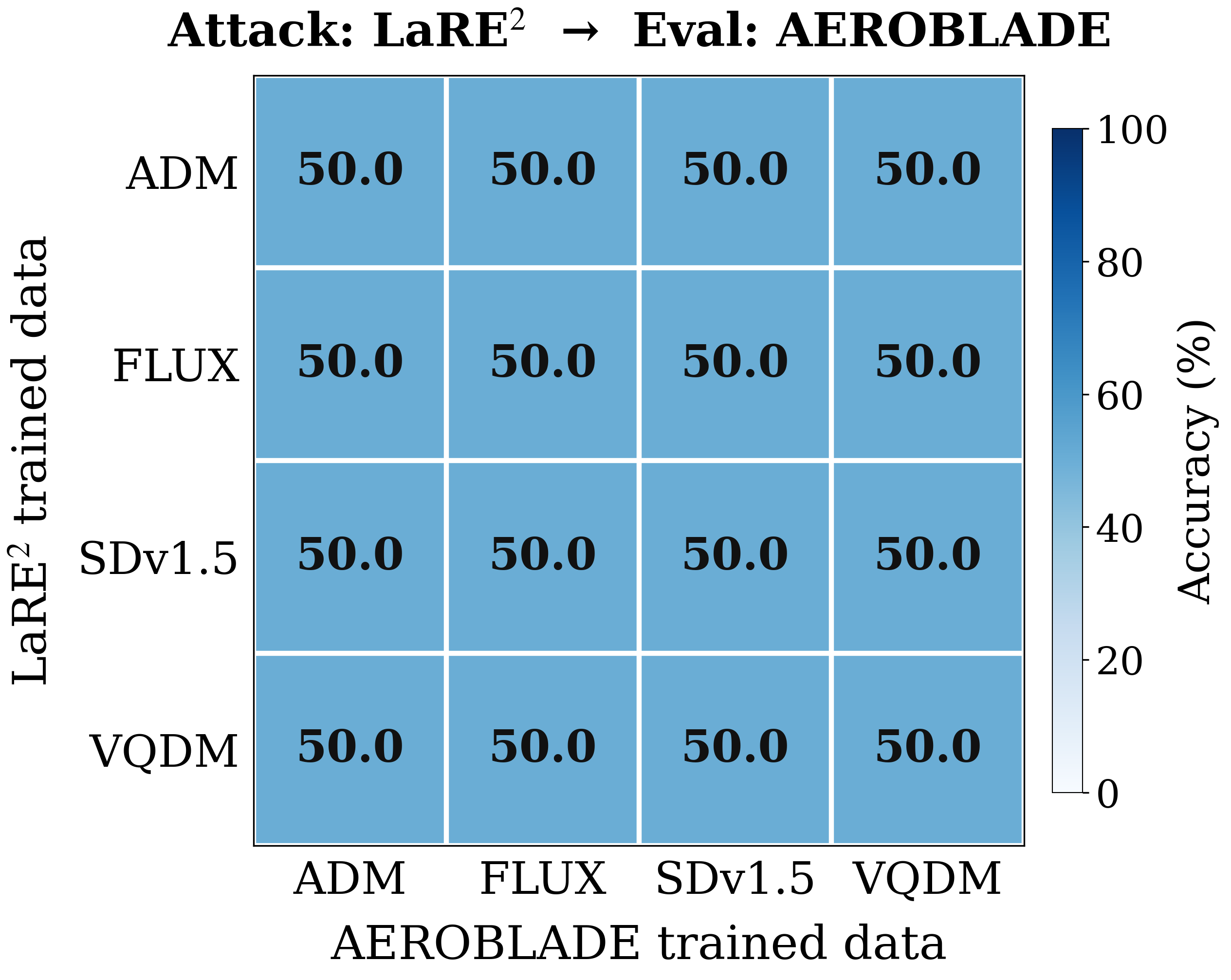}
    \caption{LaRE$^2$ $\to$ AEROBLADE}
\end{subfigure}
\hfill
\begin{subfigure}{0.32\linewidth}
    \includegraphics[width=\linewidth]{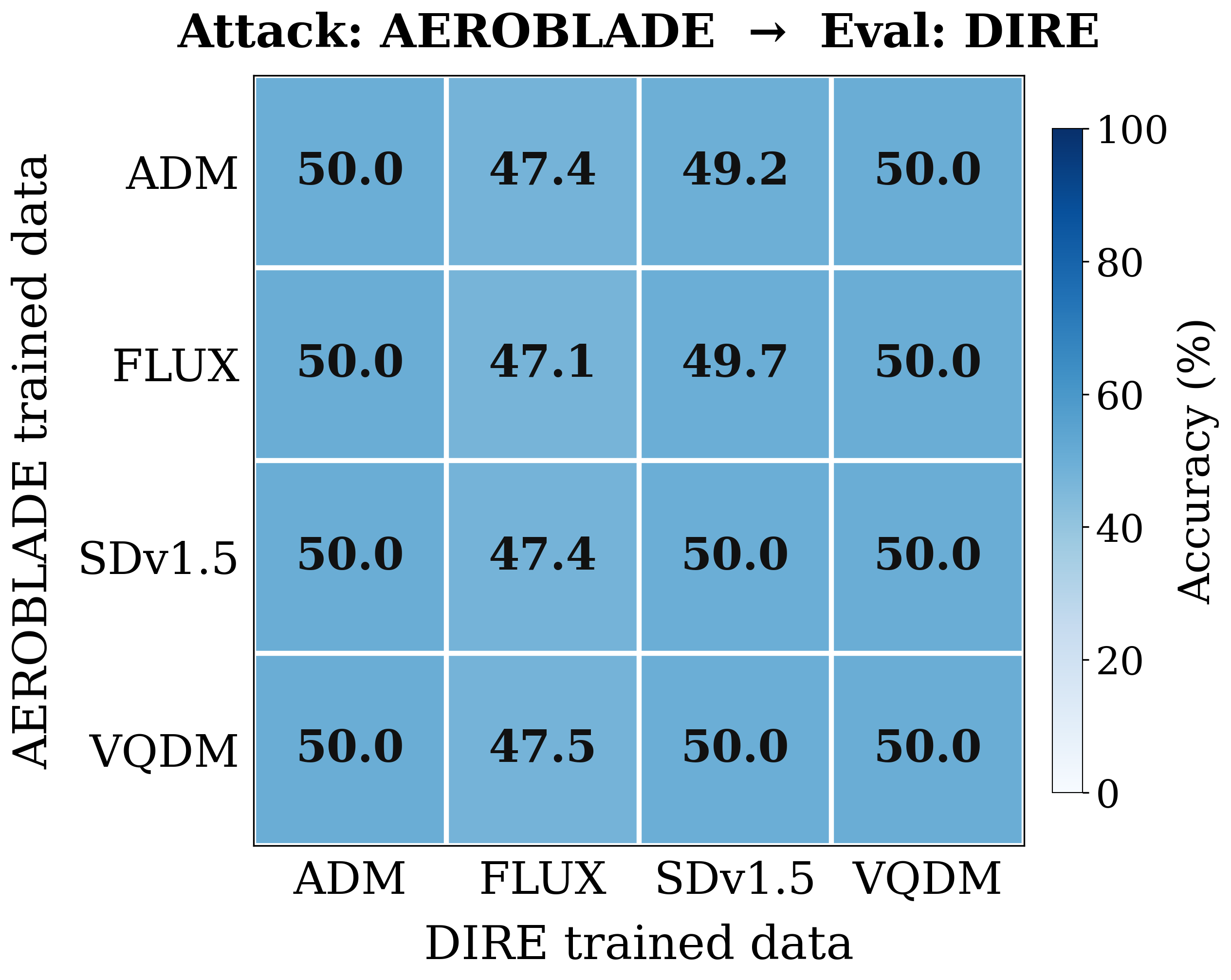}
    \caption{AEROBLADE $\to$ DIRE}
\end{subfigure}
\hfill
\begin{subfigure}{0.32\linewidth}
    \includegraphics[width=\linewidth]{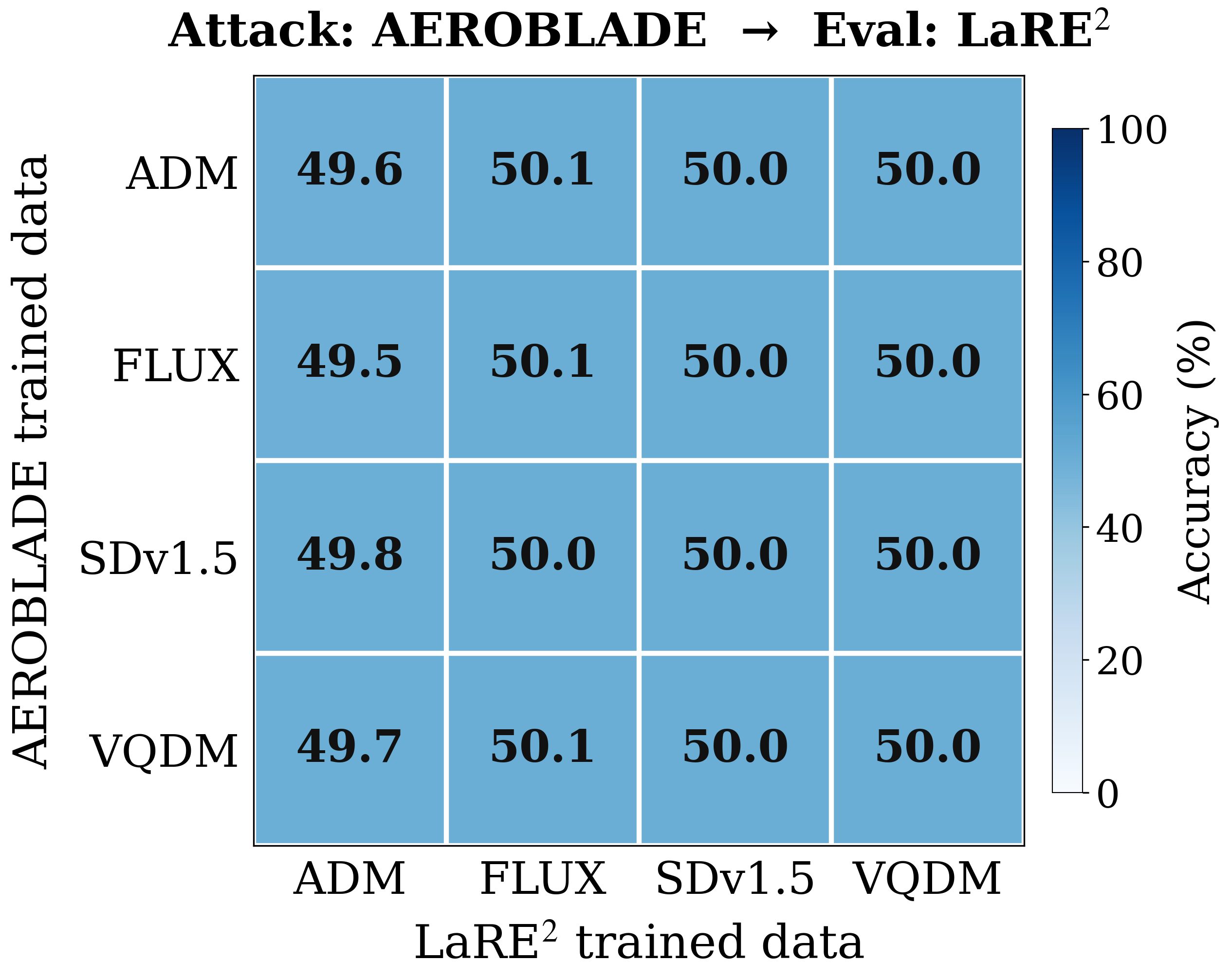}
    \caption{AEROBLADE $\to$ LaRE$^2$}
\end{subfigure}

\caption{Fully black-box transfer attack results. Each subplot shows transferability from one detection method (rows: surrogate) to another (columns: target), with generators varying across cells. Values denote post-attack accuracy (\%); diagonal entries share the same generator (cross-method only), off-diagonal entries represent the fully black-box setting.}
\label{fig:cross_both}
\vspace{-3mm}
\end{figure}

Building upon the summary provided in Section~\ref{subsec:cross_both} of the main text, Fig.~\ref{fig:cross_both} presents the comprehensive, cell-by-cell robust accuracy for the fully black-box transfer attacks across all generator-detector combinations.

\subsection{Collapse Analysis}
\label{subsec:collapse_analysis}

In the main text, we observed that successful transfer attacks frequently degrade the target detector's accuracy to approximately $50\%$. As discussed, this is not due to a symmetric flipping of labels, but rather a deterministic collapse of the classifier's decision boundary. To empirically validate this mechanism, we investigate the inherent prior bias of these classifiers when presented with completely unrecognizable, out-of-distribution (OOD) signals. 

\begin{figure}[h]
    \centering
    \includegraphics[width=0.7\linewidth]{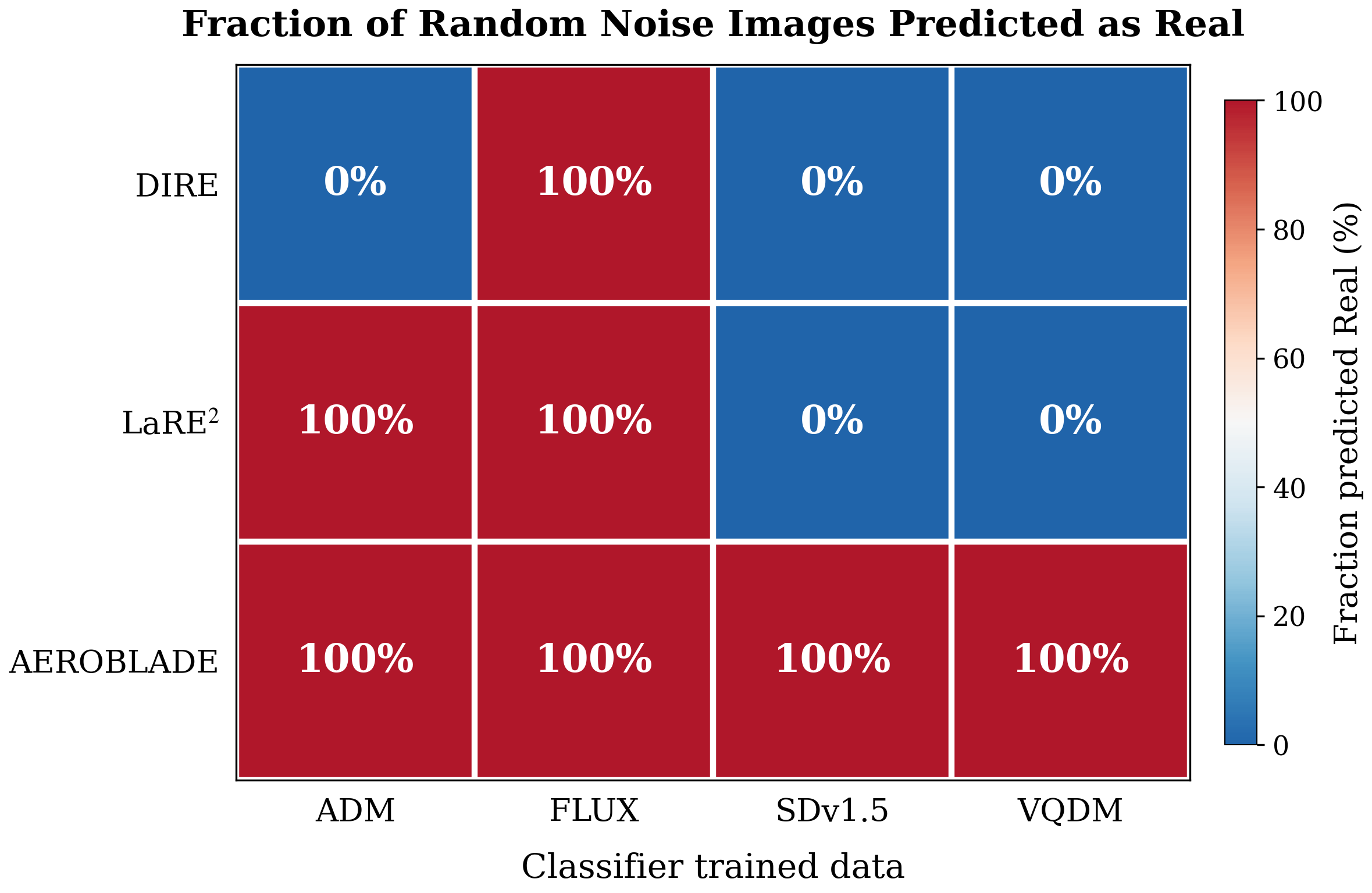}
    \caption{The fraction of pure random noise images predicted as ``Real'' by various classifiers.}
    \label{fig:appendix_collapse}
\end{figure}

Specifically, we feed pure random noise images into the trained classifiers across all combinations of detection methods and generators, recording the fraction of outputs predicted as ``Real''. The results are visualized in Fig.~\ref{fig:appendix_collapse}.

As shown in Fig.~\ref{fig:appendix_collapse}, when classifiers face inputs devoid of familiar semantic features, their predictions do not oscillate around $50\%$. Instead, they completely collapse into a rigid, unilateral default state determined by their architecture and training data. For instance, DIRE trained on ADM defaults to predicting $100\%$ ``Fake'' ($0\%$ Real), whereas AEROBLADE uniformly defaults to predicting $100\%$ ``Real'' across all generators.

This finding perfectly elucidates the $50\%$ post-attack accuracy. A highly successful adversarial attack effectively functions as a catastrophic feature disruptor. By overwhelming the intrinsic reconstructive cues with adversarial artifacts, the attack reduces the input to an ``unrecognized'' state from the target classifier's perspective. Consequently, the classifier falls back on its inherent prior bias, unilaterally classifying all test images into a single category. Given that our evaluation datasets are perfectly balanced ($50\%$ Real and $50\%$ Fake), this deterministic unilateral prediction mathematically guarantees a stable $50\%$ accuracy, confirming that the adversarial artifact has completely dominated the model's discriminative capability.

\subsection{Extended Analysis of Feature Perturbation Variations}
\label{sec:appendix_ratio}

This section provides a more comprehensive analysis of the relative perturbation variation $\rho$, extending the discussions presented in Section~\ref{subsec:snr_analysis}. Specifically, we provide the full distribution of $\rho$ across all remaining generators and compare the feature displacement caused by adversarial attacks with that caused by random noise across multiple detection methods.

\begin{figure*}[t]
    \centering
    % Row 1: DIRE
    \begin{subfigure}{0.24\linewidth}
        \includegraphics[width=\linewidth]{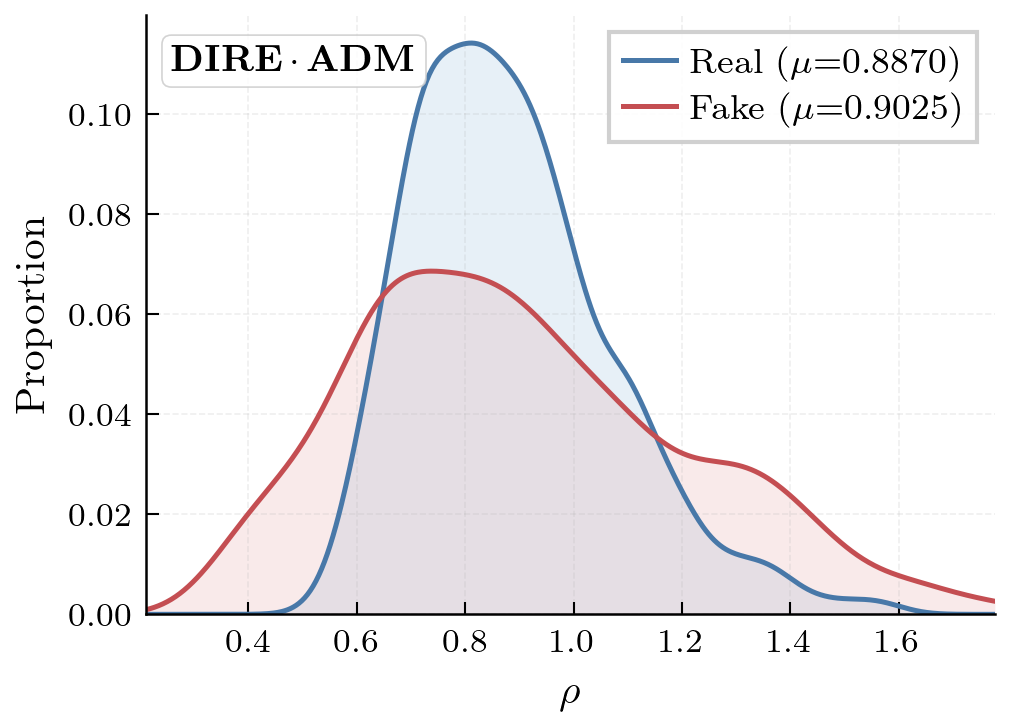}
        \caption{DIRE$\cdot$ADM}
    \end{subfigure}
    \hfill
    \begin{subfigure}{0.24\linewidth}
        \includegraphics[width=\linewidth]{image/SNR/dire/flux_ratio_l2.png}
        \caption{DIRE$\cdot$FLUX}
    \end{subfigure}
    \hfill
    \begin{subfigure}{0.24\linewidth}
        \includegraphics[width=\linewidth]{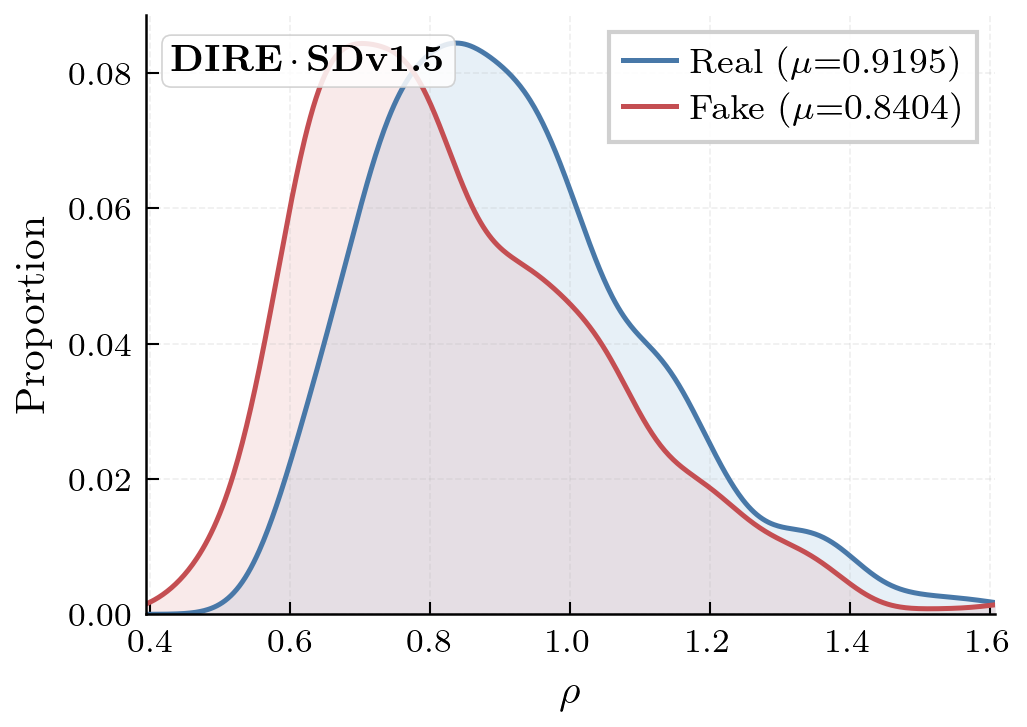}
        \caption{DIRE$\cdot$SDv1.5}
    \end{subfigure}
    \hfill
    \begin{subfigure}{0.24\linewidth}
        \includegraphics[width=\linewidth]{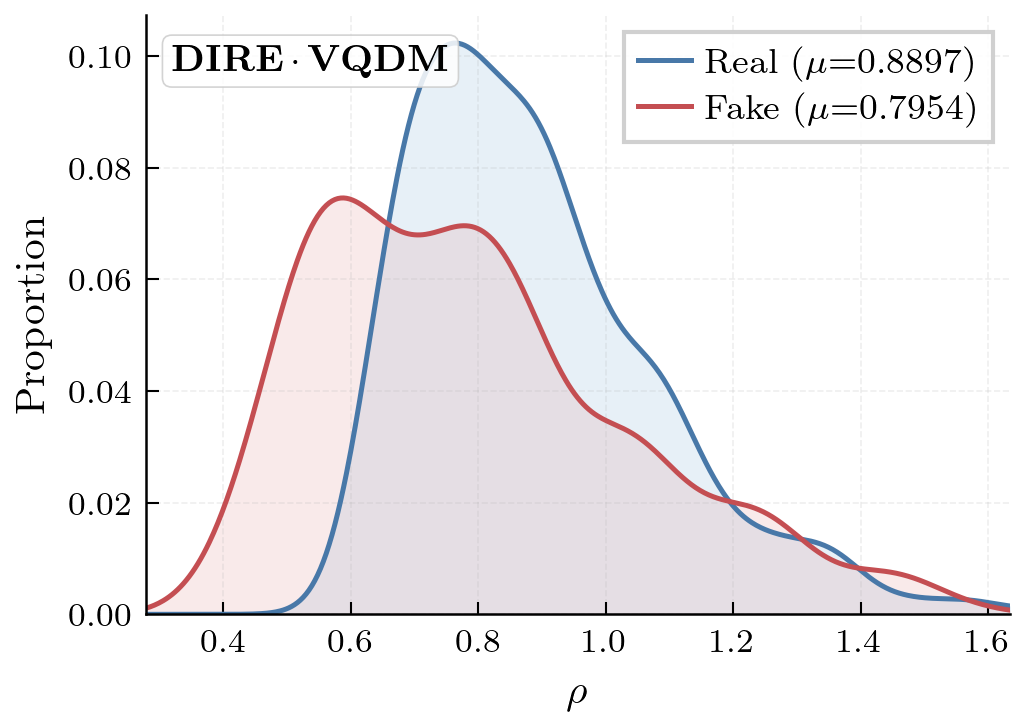}
        \caption{DIRE$\cdot$VQDM}
    \end{subfigure}
    
    \begin{subfigure}{0.24\linewidth}
        \includegraphics[width=\linewidth]{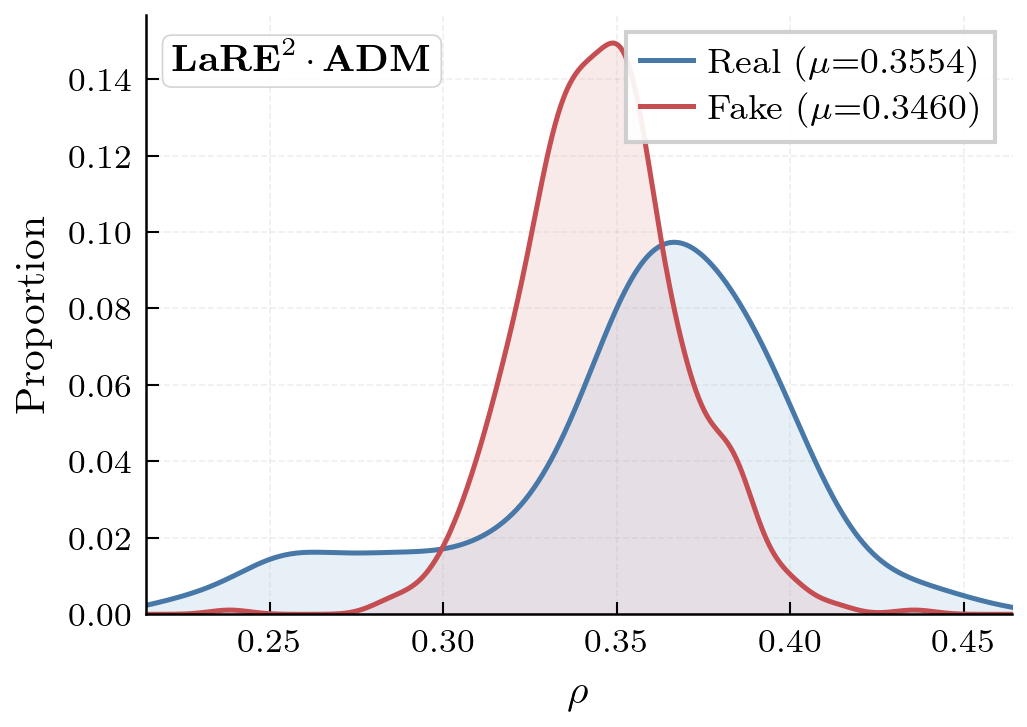}
        \caption{LaRE$^2\cdot$ADM}
    \end{subfigure}
    \hfill
    \begin{subfigure}{0.24\linewidth}
        \includegraphics[width=\linewidth]{image/SNR/lare_decoder/flux_decoded_ratio_l2.png}
        \caption{LaRE$^2\cdot$FLUX}
    \end{subfigure}
    \hfill
    \begin{subfigure}{0.24\linewidth}
        \includegraphics[width=\linewidth]{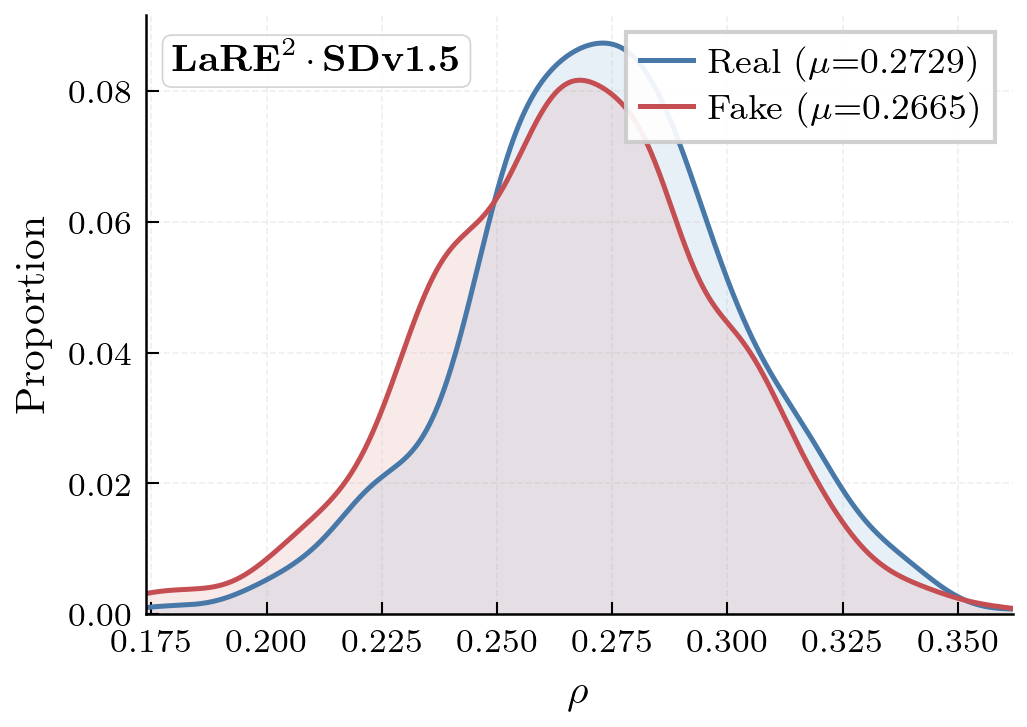}
        \caption{LaRE$^2\cdot$SDv1.5}
    \end{subfigure}
    \hfill
    \begin{subfigure}{0.24\linewidth}
        \includegraphics[width=\linewidth]{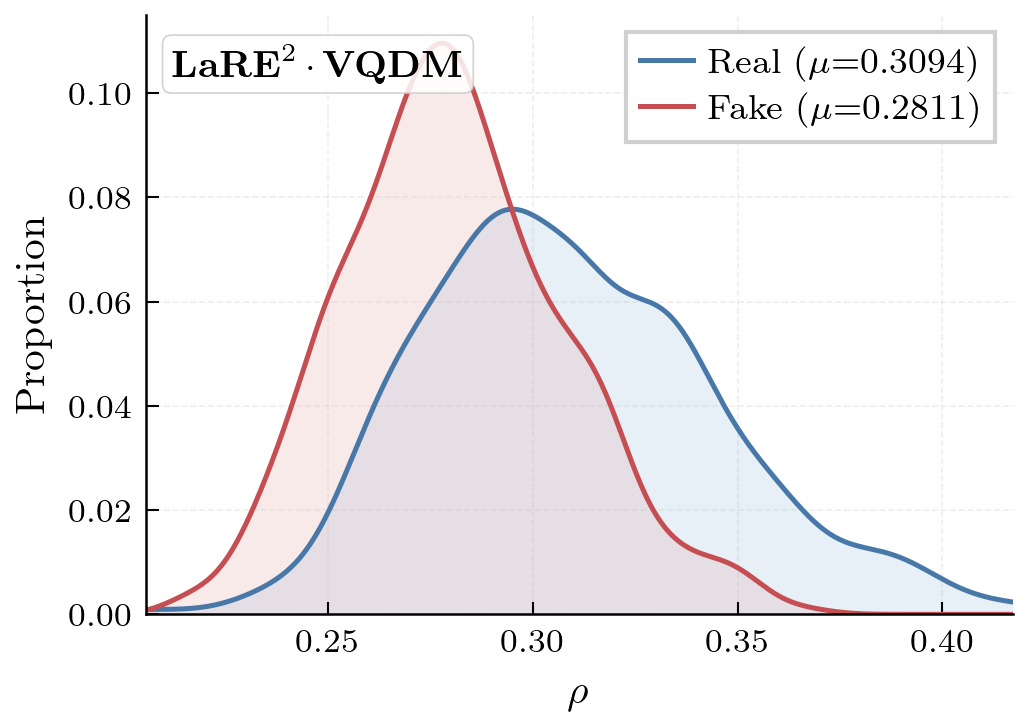}
        \caption{LaRE$^2\cdot$VQDM}
    \end{subfigure}
    
    % Row 3: AeroBlade
    \begin{subfigure}{0.48\linewidth}
        \includegraphics[width=\linewidth]{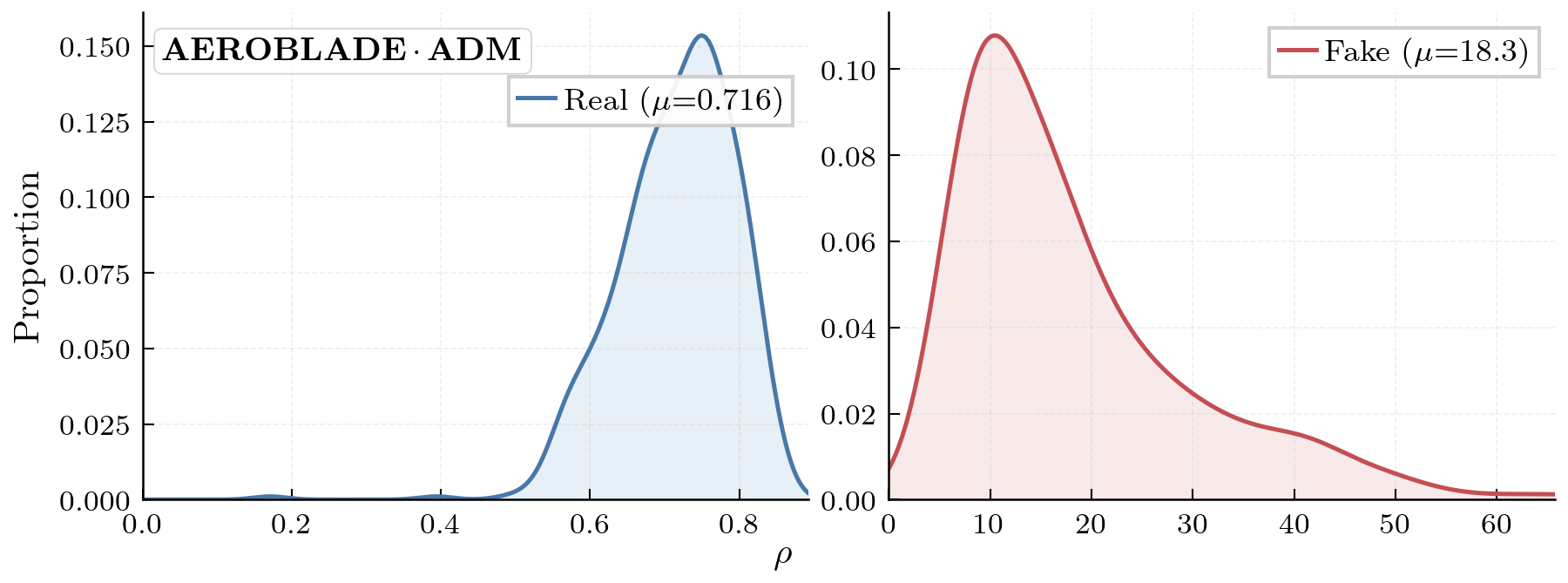}
        \caption{AEROBLADE$\cdot$ADM}
    \end{subfigure}
    \hfill
    \begin{subfigure}{0.48\linewidth}
        \includegraphics[width=\linewidth]{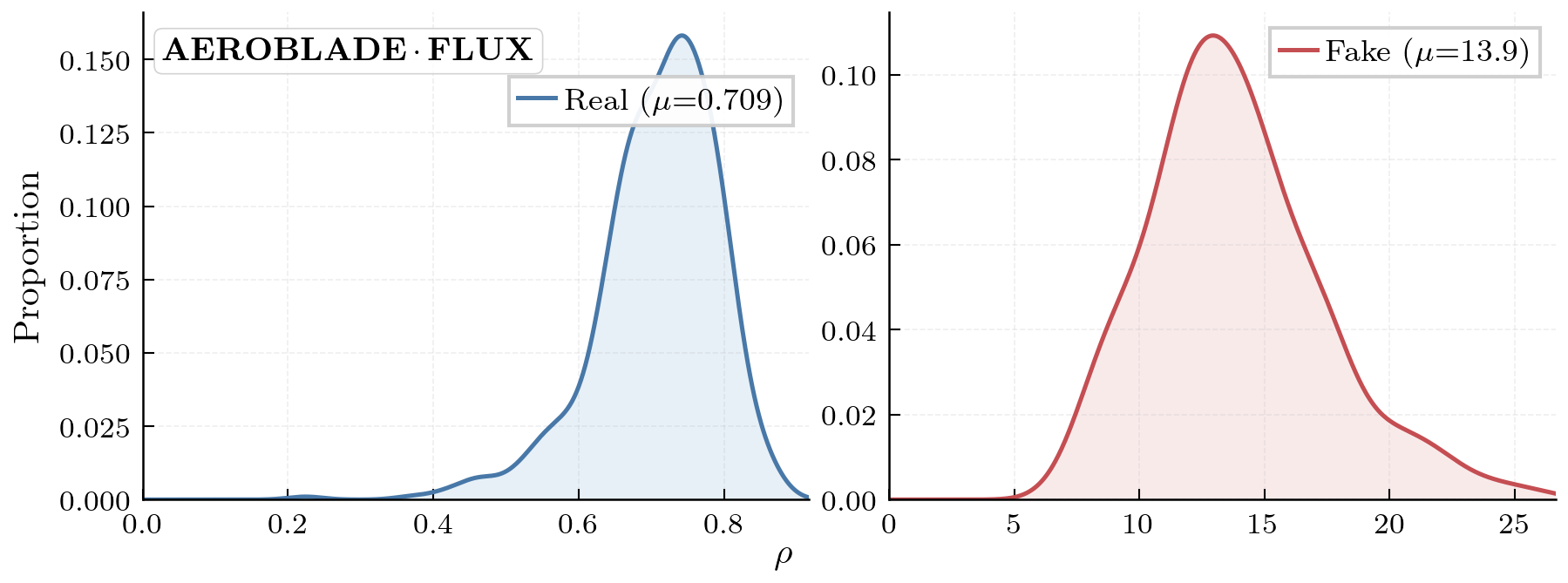}
        \caption{AEROBLADE$\cdot$FLUX}
    \end{subfigure}
    \hfill
    \begin{subfigure}{0.48\linewidth}
        \includegraphics[width=\linewidth]{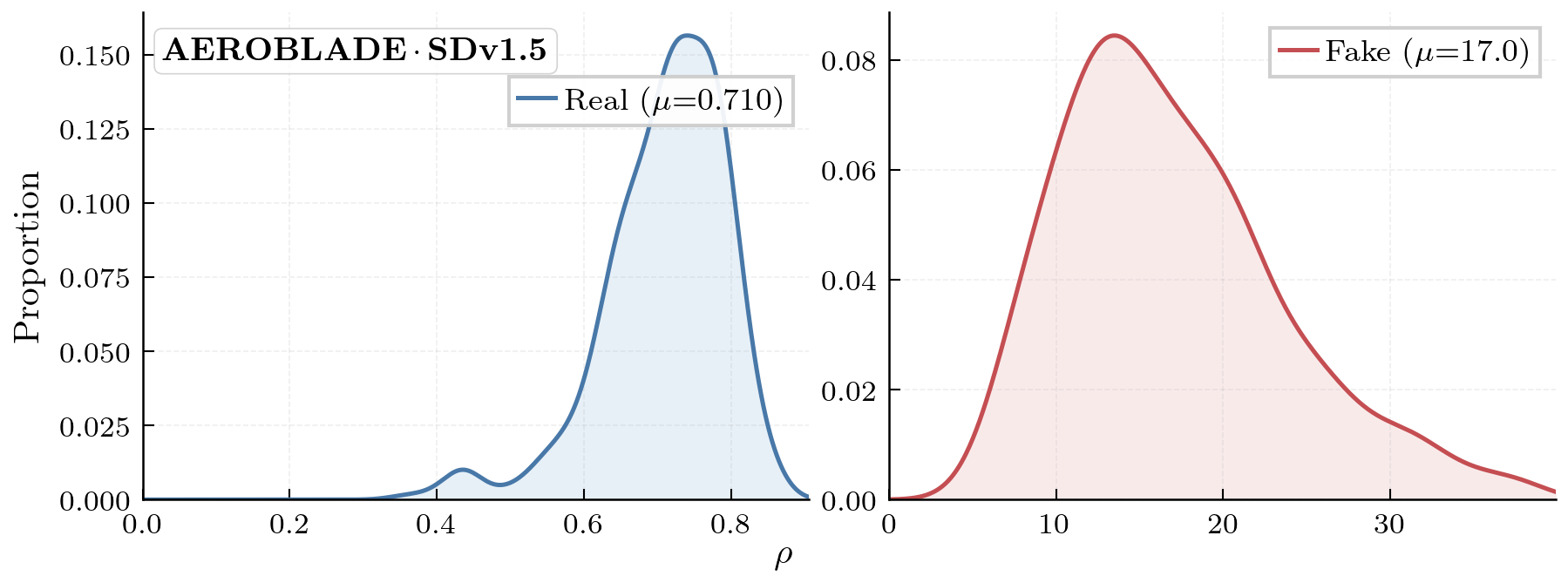}
        \caption{AEROBLADE$\cdot$SDv1.5}
    \end{subfigure}
    \hfill
    \begin{subfigure}{0.48\linewidth}
        \includegraphics[width=\linewidth]{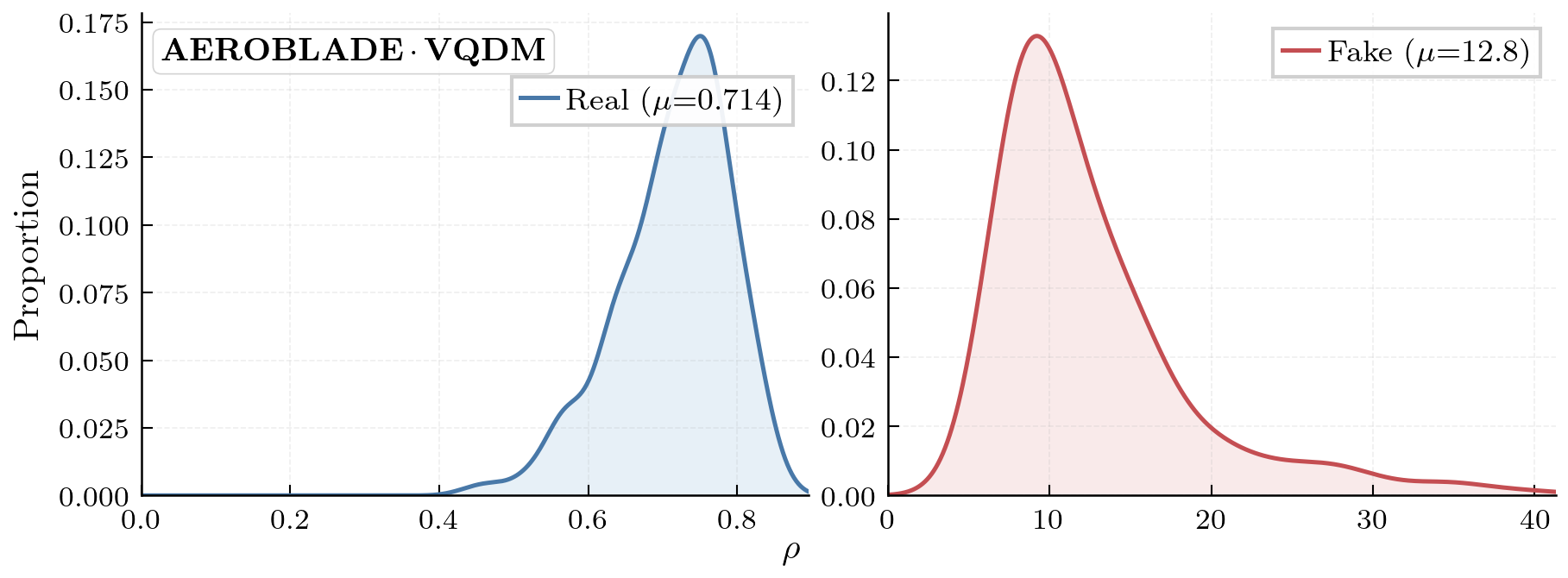}
        \caption{AEROBLADE$\cdot$VQDM}
    \end{subfigure}

    \caption{Extended distributions of relative perturbation variations $\rho$ across four generators and three detectors.}
    \label{fig:appendix_snr_extended}
\end{figure*}

\paragraph{Distributions Across Diverse Generators and Detectors.}
In the main text, we demonstrated the distributions of $\rho$ on the ADM and FLUX datasets. Fig.~\ref{fig:appendix_snr_extended} extends this analysis to a comprehensive evaluation across four generators and three detection methods. 

Consistent with earlier observations, DIRE exhibits extreme perturbation saturation ($\mu \approx 0.9$) for both Real and Fake distributions across all generators. 

Interestingly, AEROBLADE exhibits severe asymmetry. This stems directly from the inherent bounds of its distance-based detection. Since original perceptual distances are naturally small (typically $< 1$), minimizing the distance for Real images is strictly lower-bounded by zero, restricting the maximum feature deviation. Conversely, maximizing the distance for Fake images has no theoretical upper bound. This unconstrained optimization allows the attack to unboundedly enlarge the distance, causing $\rho$ to explode.

In sharp contrast, LaRE$^2$ maintains relatively lower $\rho$ values for both distributions across all architectures. This consistent behavior further validates our core conclusion: preserving feature integrity under perturbation is an architectural prerequisite for robust detection.

\paragraph{Comparative Impact: Adversarial vs. Random Perturbations.}

\begin{figure*}[t]
    \centering
    \includegraphics[width=\linewidth]{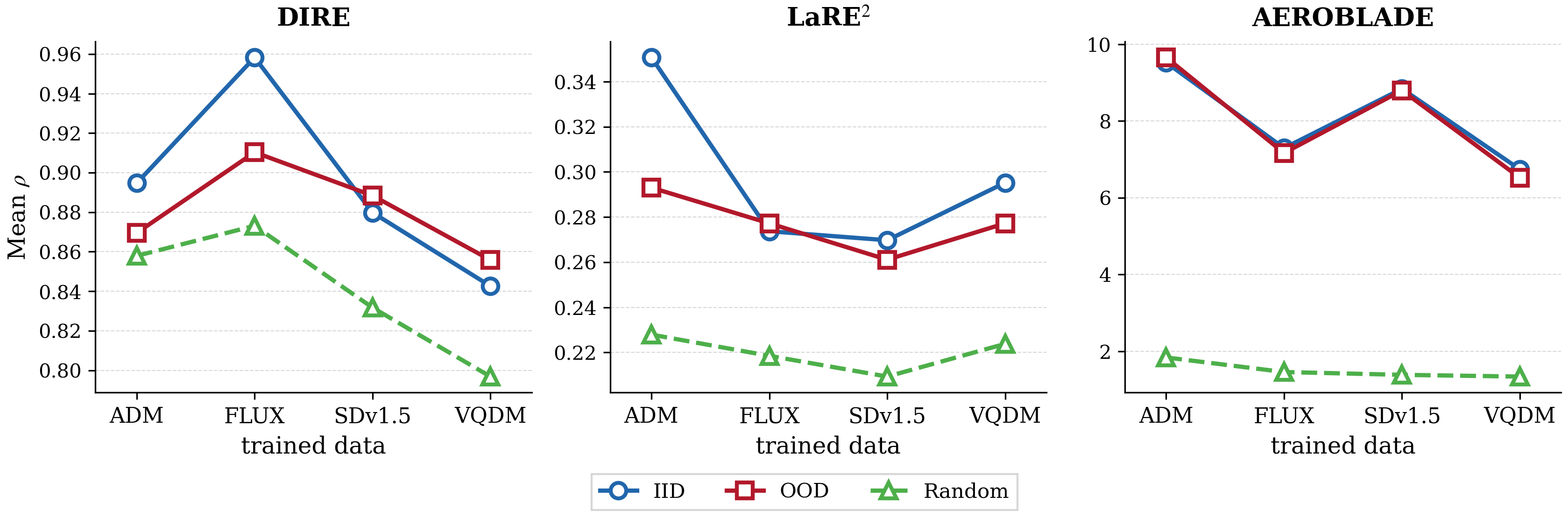} 
    \caption{\textbf{Mean Relative Perturbation ($\mu_\rho$) across detectors and datasets.} We compare the perturbation magnitude induced by White-box attacks (IID), Cross-generator Transfer attacks (OOD), and Random Gaussian noise.}
    \label{fig:rho_mean_summary}
\end{figure*}

\begin{figure*}[t]
    \centering
    % Row 1: DIRE (IID/OOD/Random)
    \begin{subfigure}{0.24\linewidth}
        \includegraphics[width=\linewidth]{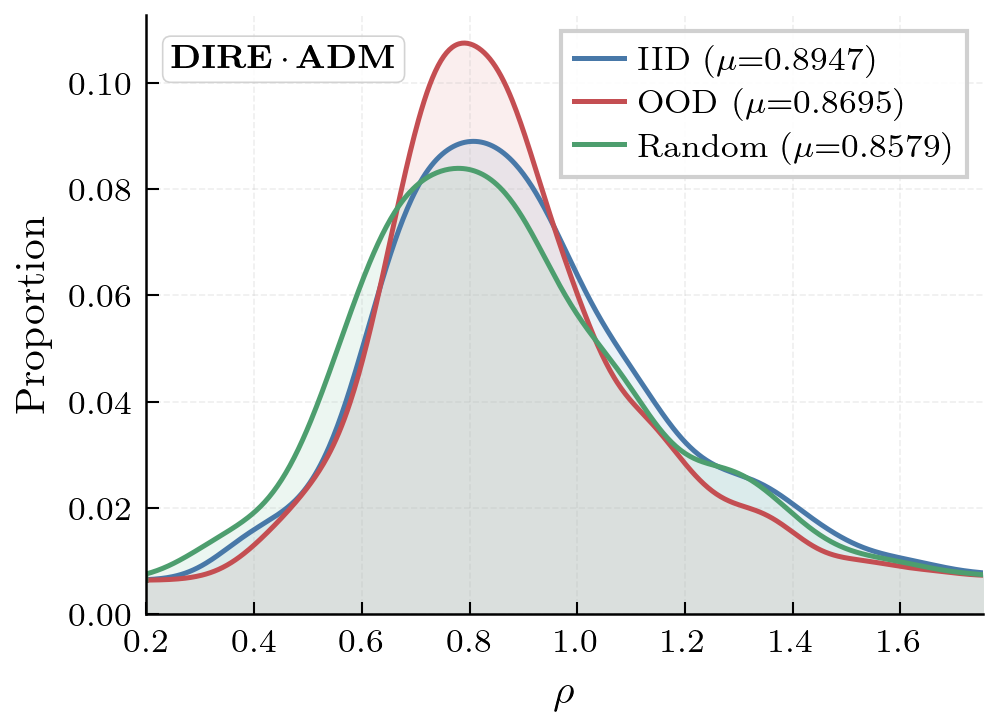}
        \caption{DIRE$\cdot$ADM}
    \end{subfigure}
    \hfill
    \begin{subfigure}{0.24\linewidth}
        \includegraphics[width=\linewidth]{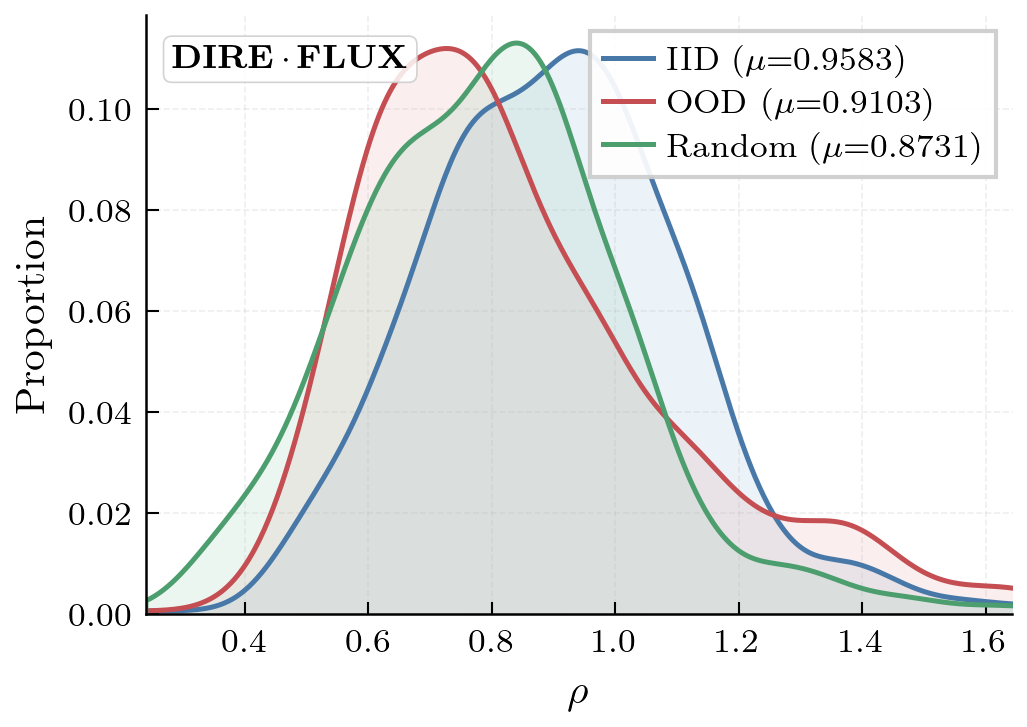}
        \caption{DIRE$\cdot$FLUX}
    \end{subfigure}
    \hfill
    \begin{subfigure}{0.24\linewidth}
        \includegraphics[width=\linewidth]{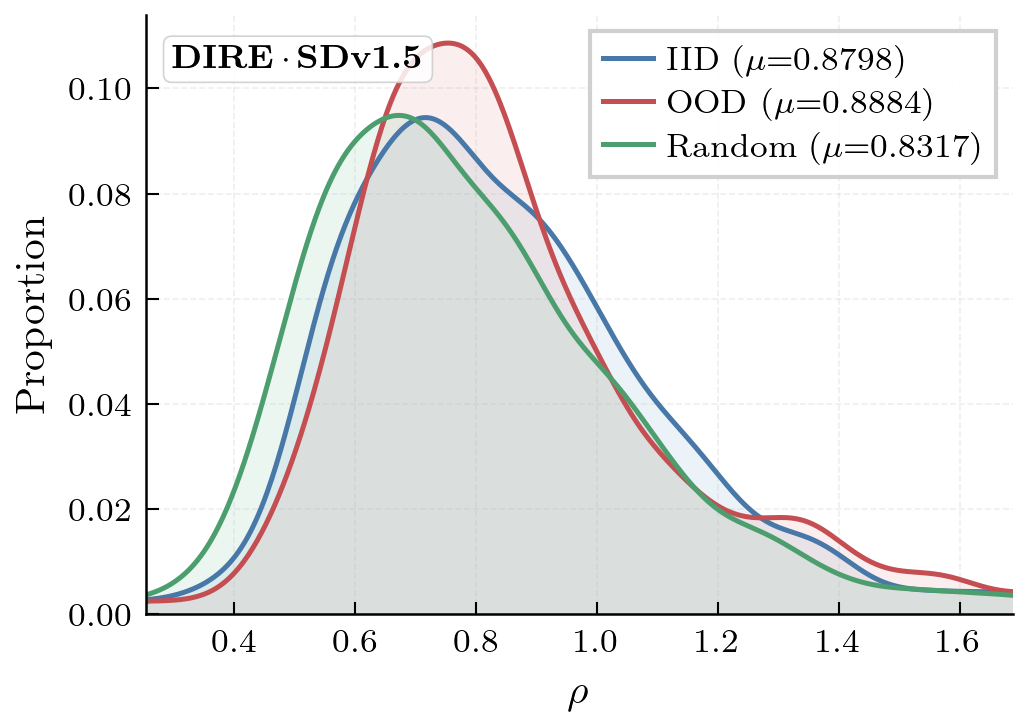}
        \caption{DIRE$\cdot$SDv1.5}
    \end{subfigure}
    \hfill
    \begin{subfigure}{0.24\linewidth}
        \includegraphics[width=\linewidth]{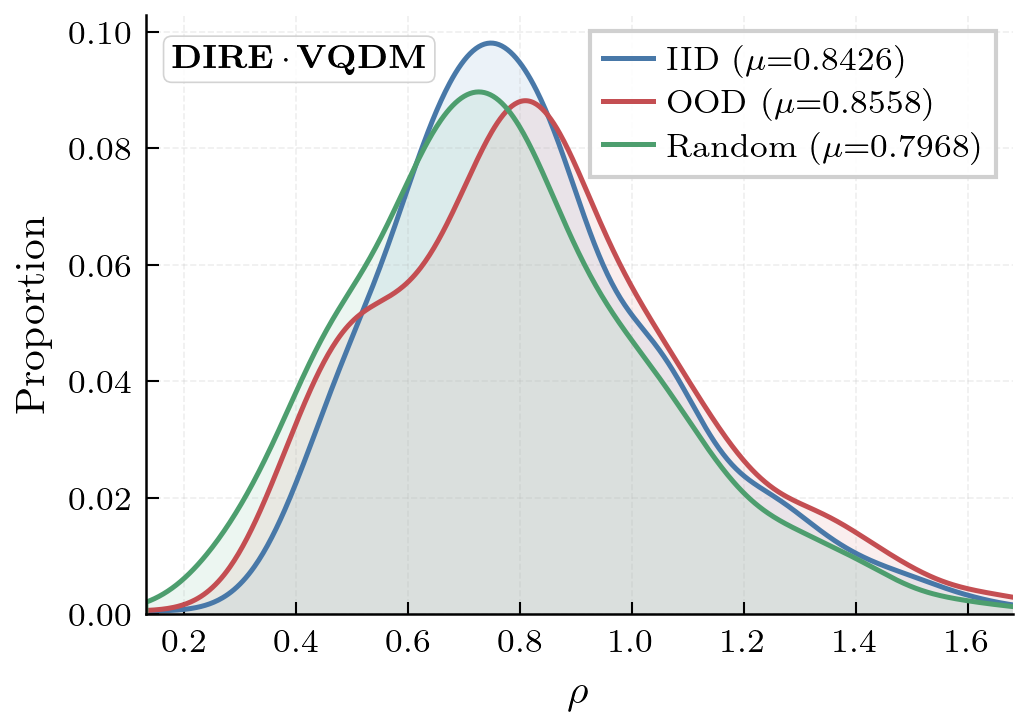}
        \caption{DIRE$\cdot$VQDM}
    \end{subfigure}
    
    % Row 2: LaRE^2 (IID/OOD/Random)
    \begin{subfigure}{0.24\linewidth}
        \includegraphics[width=\linewidth]{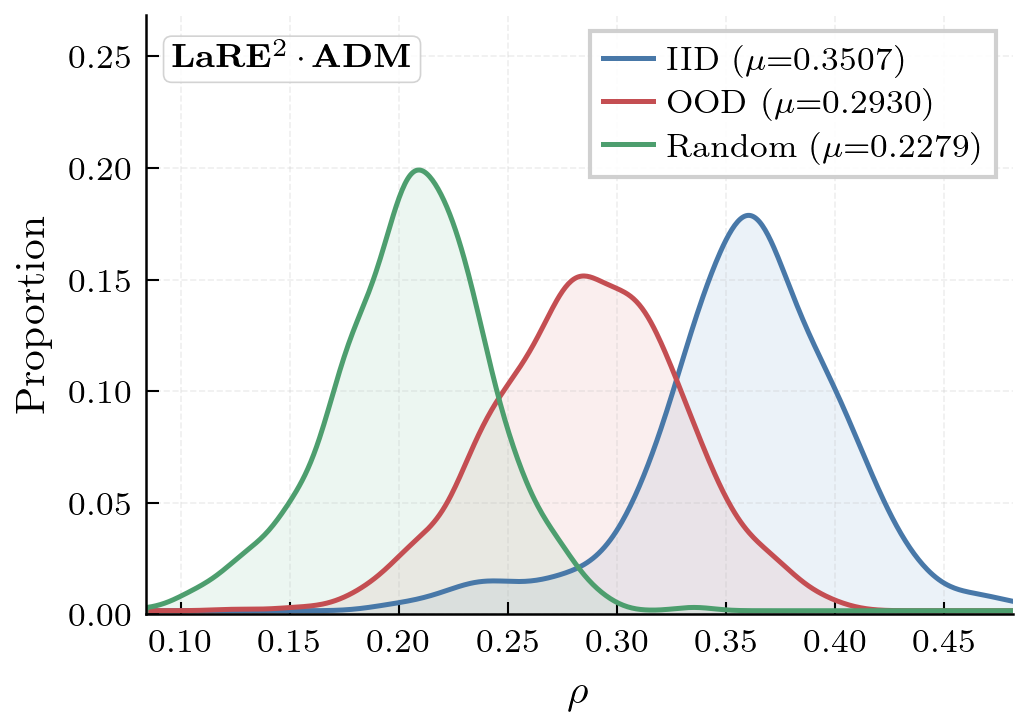}
        \caption{LaRE$^2\cdot$ADM}
    \end{subfigure}
    \hfill
    \begin{subfigure}{0.24\linewidth}
        \includegraphics[width=\linewidth]{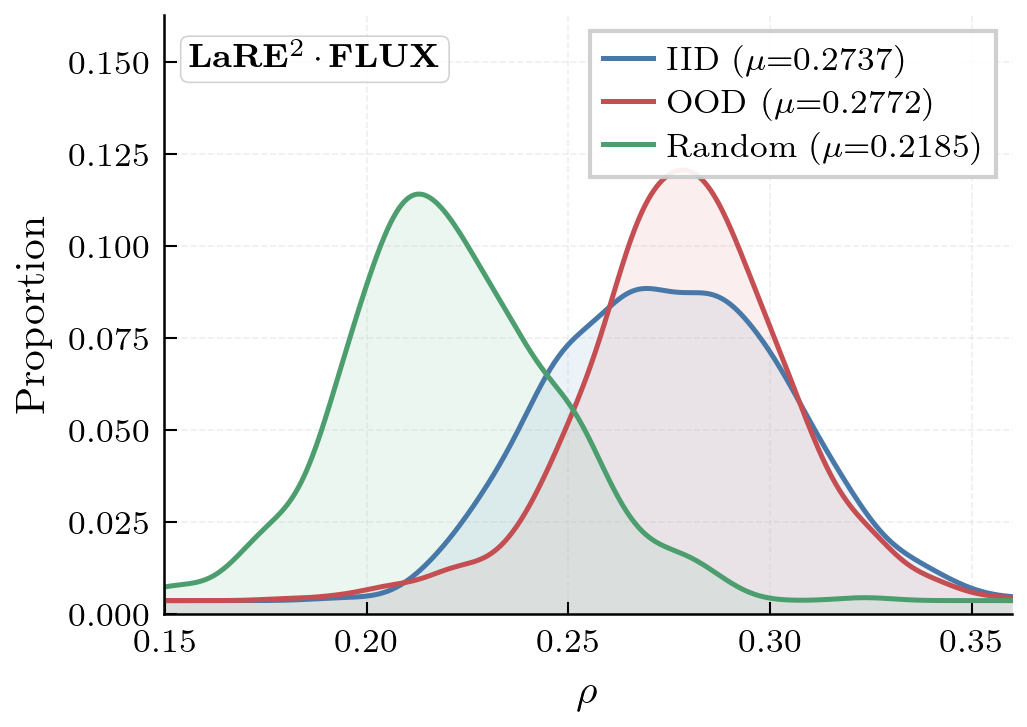}
        \caption{LaRE$^2\cdot$FLUX}
    \end{subfigure}
    \hfill
    \begin{subfigure}{0.24\linewidth}
        \includegraphics[width=\linewidth]{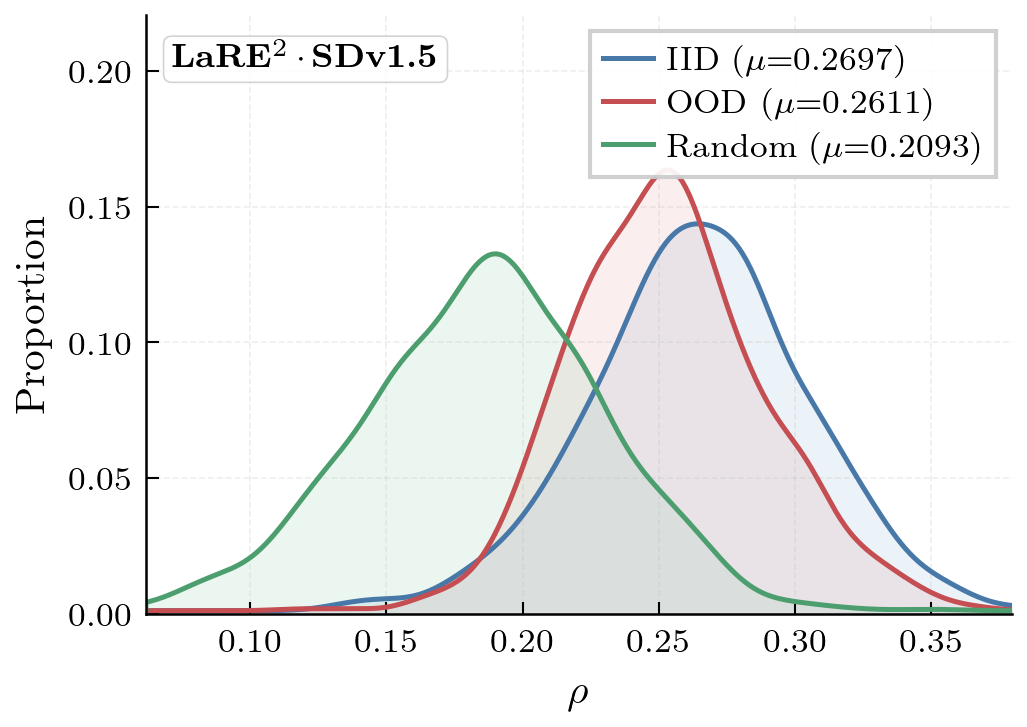}
        \caption{LaRE$^2\cdot$SDv1.5}
    \end{subfigure}
    \hfill
    \begin{subfigure}{0.24\linewidth}
        \includegraphics[width=\linewidth]{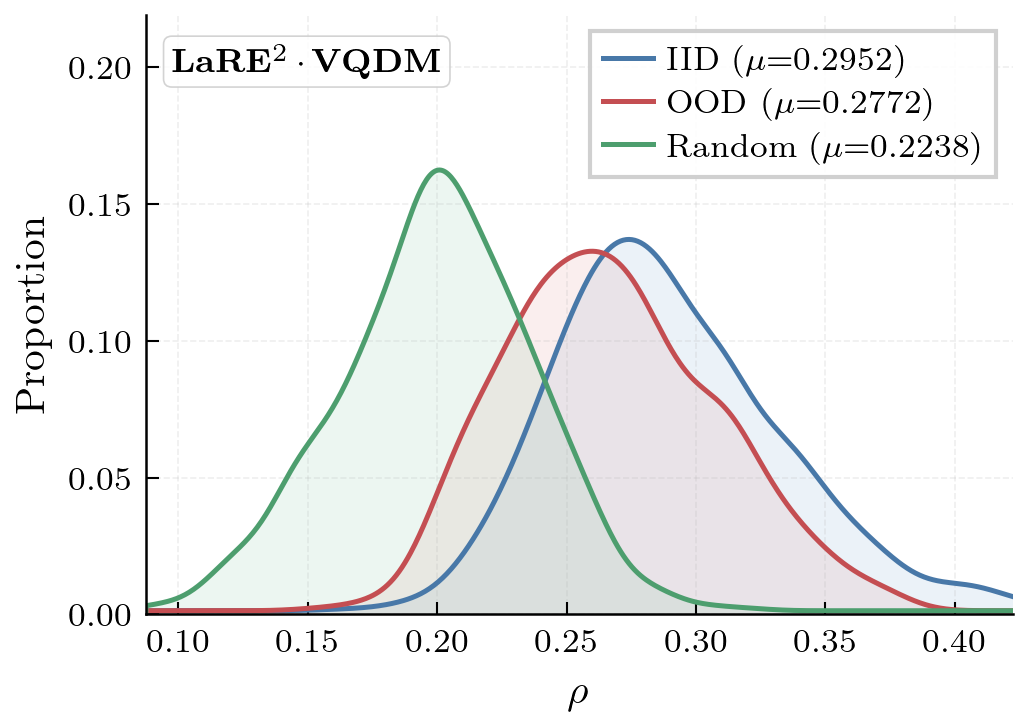}
        \caption{LaRE$^2\cdot$VQDM}
    \end{subfigure}

    % Row 3: AeroBlade (IID/OOD/Random)
    \begin{subfigure}{0.24\linewidth}
        \includegraphics[width=\linewidth]{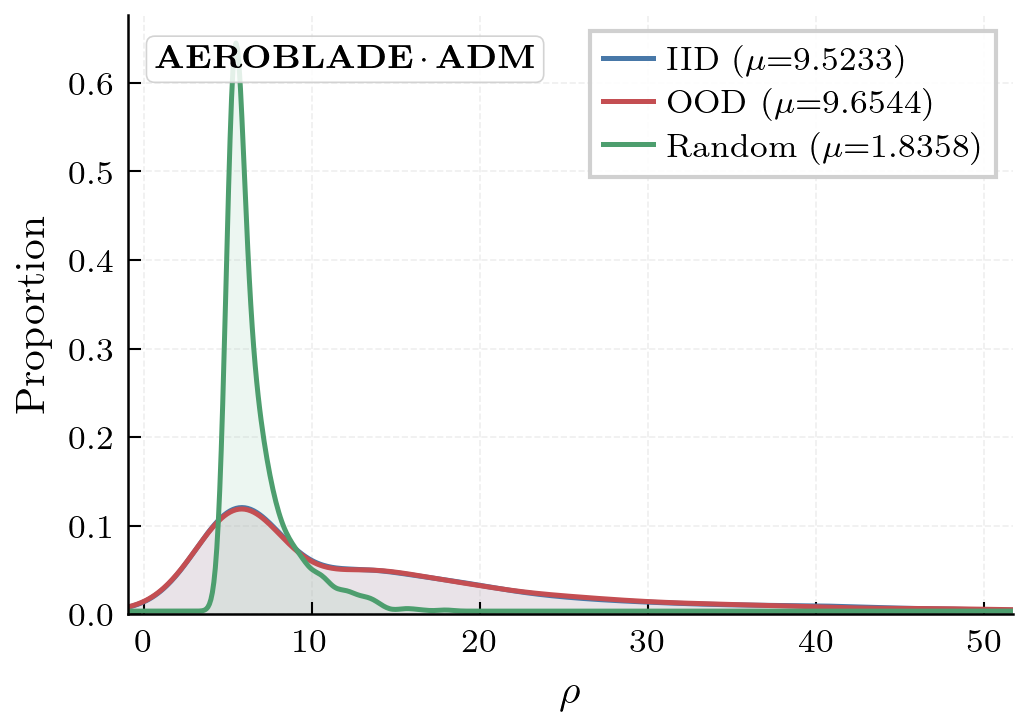}
        \caption{AEROBLADE$\cdot$ADM}
    \end{subfigure}
    \hfill
    \begin{subfigure}{0.24\linewidth}
        \includegraphics[width=\linewidth]{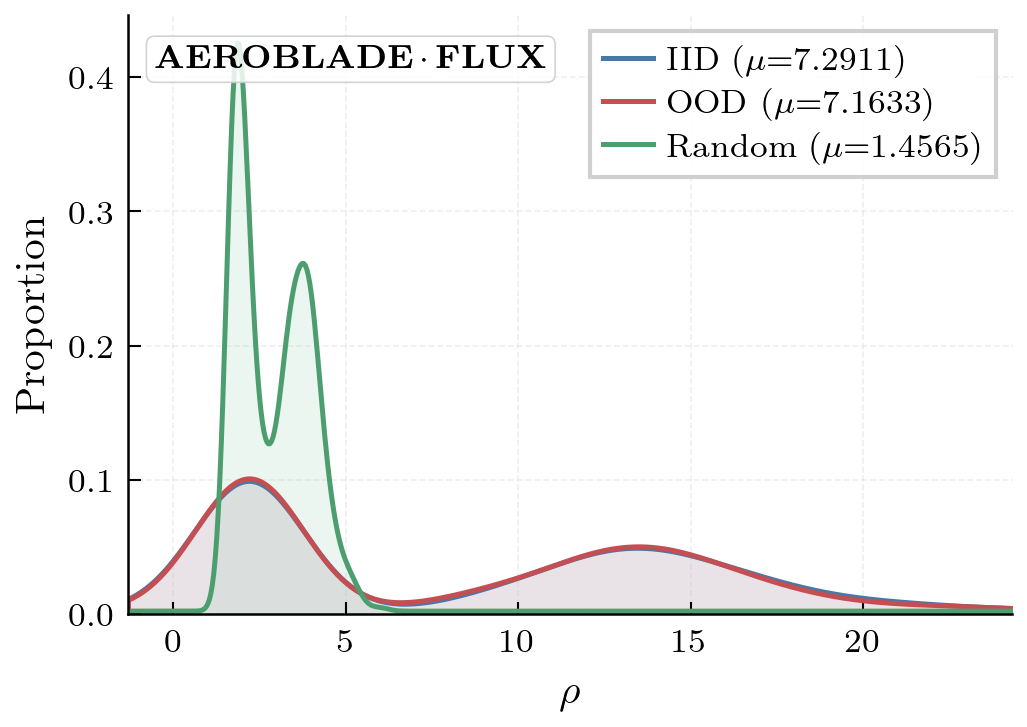}
        \caption{AEROBLADE$\cdot$FLUX}
    \end{subfigure}
    \hfill
    \begin{subfigure}{0.24\linewidth}
        \includegraphics[width=\linewidth]{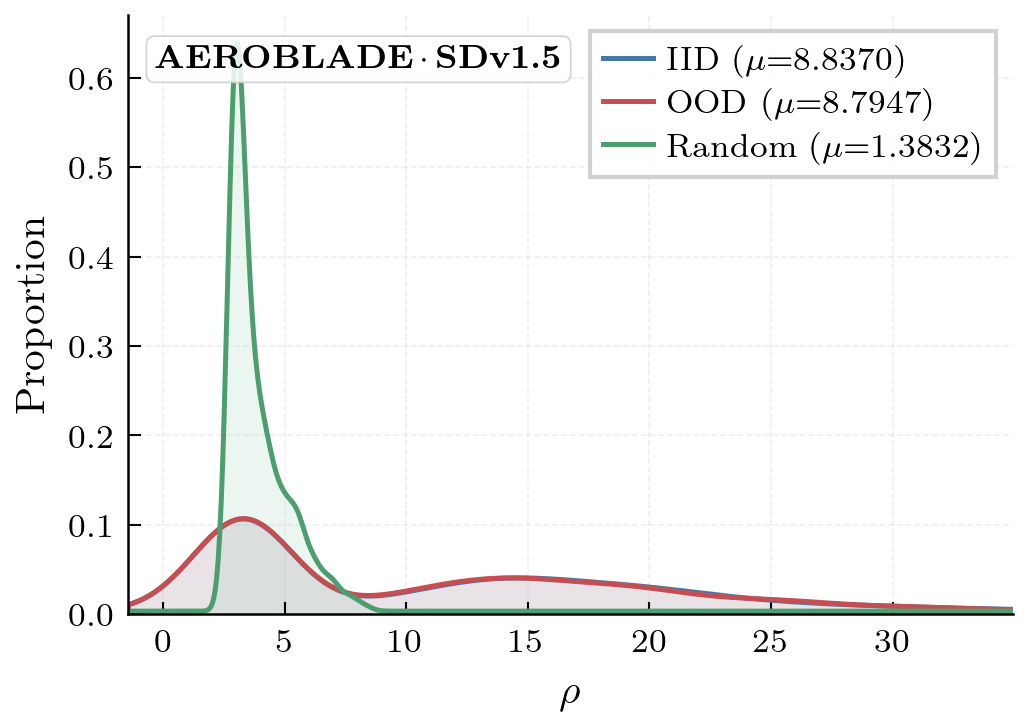}
        \caption{AEROBLADE$\cdot$SDv1.5}
    \end{subfigure}
    \hfill
    \begin{subfigure}{0.24\linewidth}
        \includegraphics[width=\linewidth]{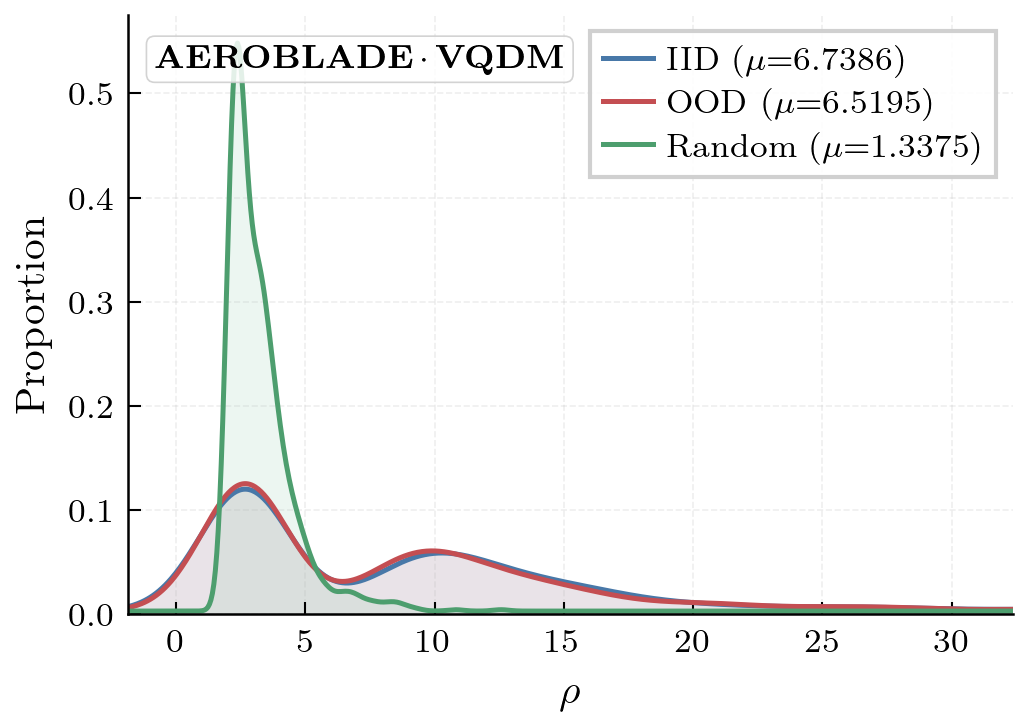}
        \caption{AEROBLADE$\cdot$VQDM}
    \end{subfigure}

    \caption{Density distributions of $\rho$ comparing IID attacks, OOD transfer attacks, and Random Gaussian noise ($\varepsilon=8/255$).}
    \label{fig:appendix_snr_comparison}
\end{figure*}

To further validate that our adversarial perturbations exert a more profound and targeted impact on the detection feature space than simple signal corruption, we compare the relative perturbation variations induced by different noise types. Figures~\ref{fig:rho_mean_summary} and \ref{fig:appendix_snr_comparison} illustrate the density of $\rho$ across the 12 method-generator combinations under three distinct settings: an IID attack (white-box), an OOD transfer attack (cross-generator), and random Gaussian noise bounded by the same maximum budget $\varepsilon = 8/255$.

Across the vast majority of settings, we observe a clear and logical hierarchy in the magnitude of feature displacement: $\mu_{\text{IID}} > \mu_{\text{OOD}} > \mu_{\text{Random}}$. While random noise naturally introduces some degree of feature deviation, it is fundamentally untargeted. In contrast, the OOD transfer attack achieves a higher deviation, and the IID attack maximizes this disruption. This consistent hierarchy provides crucial evidence that our attack methodology does not merely degrade the image with arbitrary noise. Instead, it successfully identifies and transfers optimized adversarial directions that push the latent representations significantly further away from their clean states, thereby demonstrating a superior and targeted disruptive capability.

\subsection{Adversarial Purification Setup}
\label{subsec:purification_setup}

To evaluate the robustness of detectors against post-hoc defenses, we establish a rigorous adversarial purification protocol.

\begin{itemize}[leftmargin=*, noitemsep]
    \item \textbf{Evaluation Setup:} We construct a comprehensive purification benchmark consisting of 12 distinct evaluation subsets, spanning all combinations of the 4 generative models (ADM, SDv1.5, FLUX, VQDM) and 3 target detectors (DIRE, AEROBLADE, LaRE$^2$). Each subset comprises 2,000 adversarial examples, strictly balanced between 1,000 real and 1,000 fake images, all of which are adversarially perturbed against the target detector.
    
    \item \textbf{Purification Model:} We strictly employ a source-matched purification strategy. For each subset, the adversarial examples are purified using the exact same generative model that was used for the original synthesis (e.g., using FLUX to purify FLUX-generated attacks). This ensures the defense utilizes the optimal prior distribution.
    
    \item \textbf{Purification Strength:} To systematically evaluate the defense, we sweep the noise level using a ratio $t/T \in \{0.01, 0.02, 0.03, 0.05, 0.1\}$ relative to the total diffusion schedule. For discrete architectures like VQDM, this continuous ratio is explicitly mapped to the nearest integer step. We report the accuracy trajectory to illustrate how the defense effectiveness changes under different purification strengths.
\end{itemize}

%=====================================================================
\subsection{Adversarial Training Setup}
\label{subsec:adv_training_efficient}

Standard adversarial training in the pixel space is computationally prohibitive due to the expensive gradient backpropagation through the diffusion backbone. To address this, we adopt an efficient optimization strategy tailored to each detector's architecture.

For DIRE, we treat the reconstruction process as a fixed pre-processing step $\phi_{\text{DIRE}}(x)$ and perform attacks directly on the residual features $r = \phi_{\text{DIRE}}(x)$. The training objective is:
\begin{equation}
    \min_\theta \mathbb{E}_{(x,y) \sim \mathcal{D}} \left[ \max_{\|\delta\|_\infty \leq \varepsilon_{\text{feat}}} \mathcal{L}_{\text{ce}}(f_\theta(r + \delta), y) \right].
\end{equation}
In practice, we calibrate the pixel-level budgets $\varepsilon \in \{1/255, 2/255, 4/255, 8/255\}$ to their feature-space equivalents (e.g., $8/255 \to 0.0089$) and solve the inner maximization using PGD-$K$ with varying steps $K \in \{1, 2, 4, 8\}$ and a step size of $\alpha = \varepsilon_{\text{feat}}/4$.

For LaRE$^2$, the classifier takes both the image $x$ and the estimated error map $e = \phi_{\text{LaRE$^2$}}(x)$. We apply the perturbation $\delta$ in the pixel space but detach the gradient flow through the error extraction module to bypass the diffusion backbone. The objective is:
\begin{equation}
    \min_\theta \mathbb{E}_{(x,y) \sim \mathcal{D}} \left[ \max_{\|\delta\|_\infty \leq \varepsilon} \mathcal{L}_{\text{ce}}\left(f_\theta(x + \delta, \mathrm{sg}[\phi_{\text{LaRE$^2$}}(x + \delta)]), y\right) \right],
\end{equation}
where $\mathrm{sg}[\cdot]$ denotes the stop-gradient operation. We apply perturbations directly in the pixel space with $\varepsilon \in \{1/255, 2/255, 4/255, 8/255\}$ and a step size of $\alpha = \varepsilon/4$, using the same PGD steps $K \in \{1, 2, 4, 8\}$.

To ensure a fair comparison, we train across all aforementioned configurations ($K$ steps and $\varepsilon$ budgets) and select the specific checkpoint that achieves the highest robust accuracy on the held-out validation set. For the final reporting, we evaluate \textit{clean accuracy} on unperturbed images and \textit{robust accuracy} under a rigorous APGD attack (100 steps, $\varepsilon = 8/255$) to strictly upper-bound the defense performance. The procedure is summarized in Algorithm~\ref{alg:efficient_adv_train}.

% Furthermore, we empirically observe a severe trade-off when exploring larger $K$: as the number of PGD steps increases during training, the clean accuracy suffers a continuous degradation, ultimately causing the optimization to collapse and the training to fail. This phenomenon justifies our configuration bounds and further underscores the inherent incompatibility between standard adversarial training and fragile reconstruction-based features.

\begin{algorithm}[h]
\caption{Adversarial Training}
\label{alg:efficient_adv_train}
\begin{algorithmic}[1]
\STATE \textbf{Input:} Training data $\mathcal{D}$, feature extractors $\phi_{\mathrm{DIRE}}$ and $\phi_{\mathrm{LaRE}^2}$, initialized classifier $f_\theta$, budgets $\varepsilon$ and $\varepsilon_{\mathrm{feat}}$, attack steps $K$, step size $\alpha$.
\STATE \textbf{Output:} Trained classifier parameters $\theta$.
\FOR{each minibatch $\{(x_i, y_i)\}_{i=1}^B \sim \mathcal{D}$}
    \IF{DIRE}
        \STATE $r_i \leftarrow \phi_{\mathrm{DIRE}}(x_i)$ \hfill $\triangleright$ Pre-compute reconstruction residual
        \STATE $\delta_i \sim \mathcal{U}(-\varepsilon_{\mathrm{feat}},\, \varepsilon_{\mathrm{feat}})$
        \FOR{$t = 1$ \TO $K$}
            \STATE $\delta_i \leftarrow \Pi_{\|\cdot\|_\infty \le\varepsilon_{\mathrm{feat}}}\Big(\delta_i + \alpha \cdot \mathrm{sign}\big(\nabla_{\delta_i} \mathcal{L}_{\mathrm{ce}}(f_\theta(r_i + \delta_i),\, y_i)\big)\Big)$
        \ENDFOR
        \STATE Update $\theta$ to minimize $\mathcal{L}_{\mathrm{ce}}\big(f_\theta(r_i + \delta_i),\, y_i\big)$
    \ELSIF{LaRE$^2$}
        \STATE $\delta_i \sim \mathcal{U}(-\varepsilon,\, \varepsilon)$
        \FOR{$t = 1$ \TO $K$}
            \STATE $e_i \leftarrow \mathrm{sg}\big[\phi_{\mathrm{LaRE}^2}(x_i + \delta_i)\big]$ \hfill $\triangleright$ Stop-gradient on $\phi$
            \STATE $\delta_i \leftarrow \Pi_{\|\cdot\|_\infty \le \varepsilon}\Big(\delta_i + \alpha \cdot \mathrm{sign}\big(\nabla_{\delta_i} \mathcal{L}_{\mathrm{ce}}(f_\theta(x_i + \delta_i,\, e_i),\, y_i)\big)\Big)$
        \ENDFOR
        \STATE Update $\theta$ to minimize $\mathcal{L}_{\mathrm{ce}}\big(f_\theta(x_i + \delta_i,\, e_i,\, y_i\big)$
    \ENDIF
\ENDFOR
\end{algorithmic}
\end{algorithm}

\subsection{Visualizations of Adversarial Examples}
\label{subsec:Visualizations}

\begin{figure*}[t]
    \centering
    % 1. Real Images
    \begin{subfigure}{\linewidth}
        \centering
        \includegraphics[width=\linewidth]{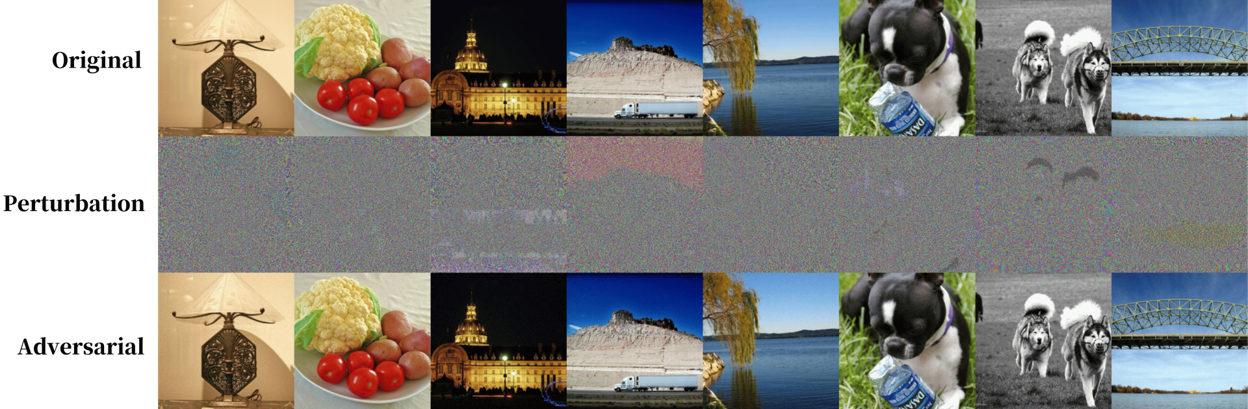} 
        \caption{Real Images (ImageNet)}
    \end{subfigure}
    \par\medskip
    
    % 2. IID Synthetic Images
    \begin{subfigure}{\linewidth}
        \centering
        \includegraphics[width=\linewidth]{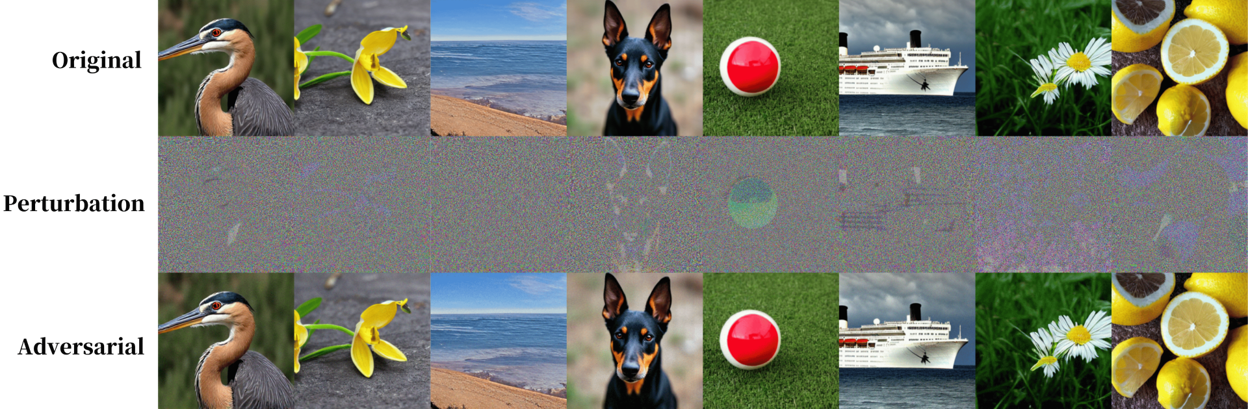}
        \caption{IID Synthetic Images (White-box attacks on images from the detector's known generative source)}
    \end{subfigure}
    \par\medskip
    
    % 3. OOD Synthetic Images
    \begin{subfigure}{\linewidth}
        \centering
        \includegraphics[width=\linewidth]{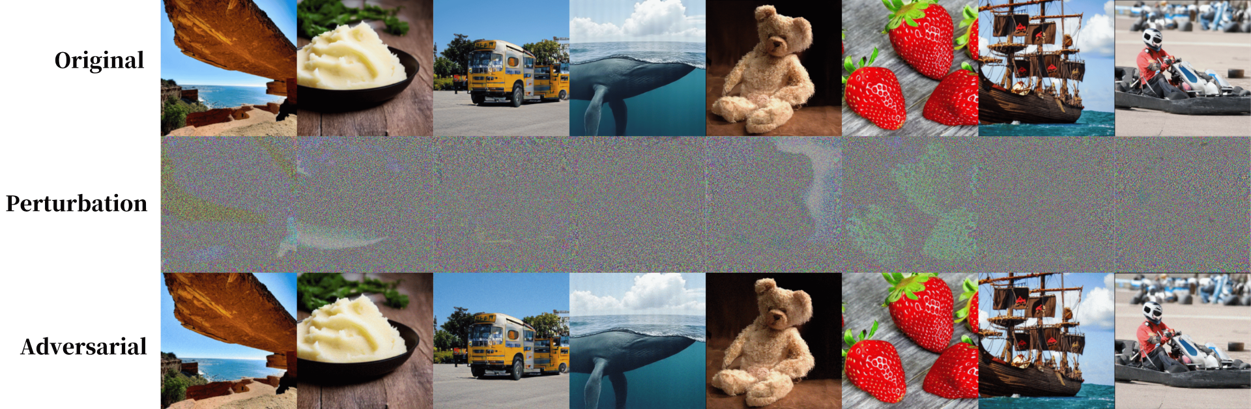}
        \caption{OOD Synthetic Images (White-box attacks on images from an unseen generative source)}
    \end{subfigure}
    
    \caption{\textbf{Visualizations of Adversarial Attacks against DIRE.} 
    The rows in each subfigure from top to bottom represent: Original Image, Adversarial Perturbation ($\times 15$ magnified), and the resulting Adversarial Example.}
    \label{fig:vis_dire}
\end{figure*}

\begin{figure*}[t]
    \centering
    % 1. Real Images
    \begin{subfigure}{\linewidth}
        \centering
        \includegraphics[width=\linewidth]{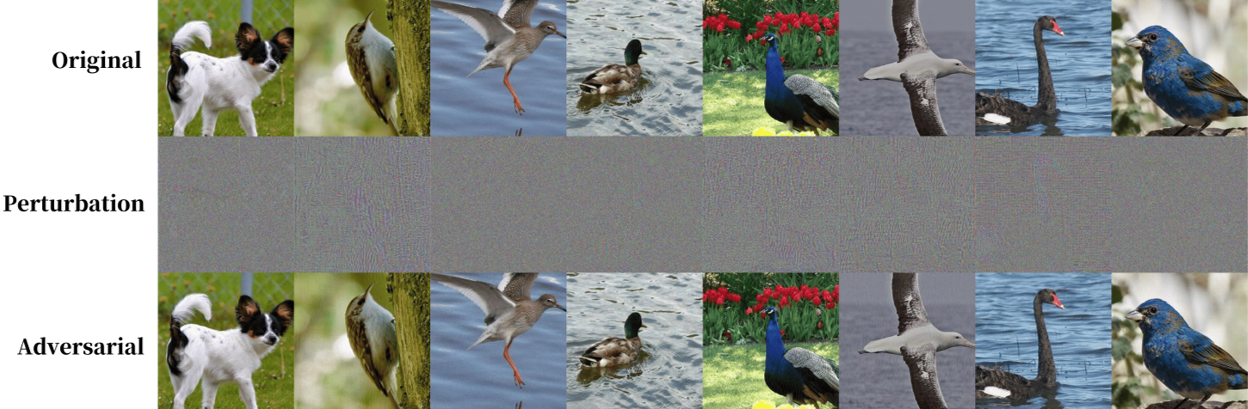}
        \caption{Real Images (ImageNet)}
    \end{subfigure}
    \par\medskip
    
    % 2. IID Synthetic Images
    \begin{subfigure}{\linewidth}
        \centering
        \includegraphics[width=\linewidth]{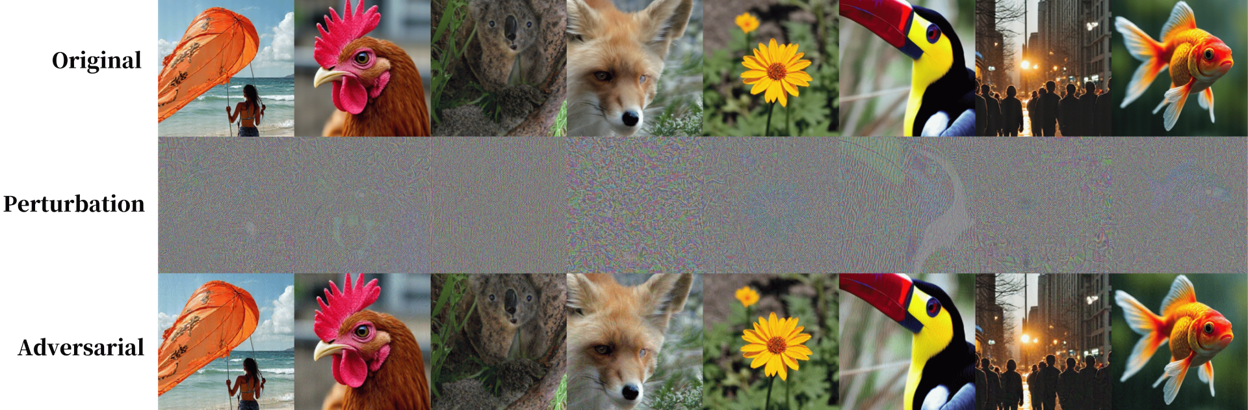}
        \caption{IID Synthetic Images (White-box attacks on images from the detector's known generative source)}
    \end{subfigure}
    \par\medskip
    
    % 3. OOD Synthetic Images
    \begin{subfigure}{\linewidth}
        \centering
        \includegraphics[width=\linewidth]{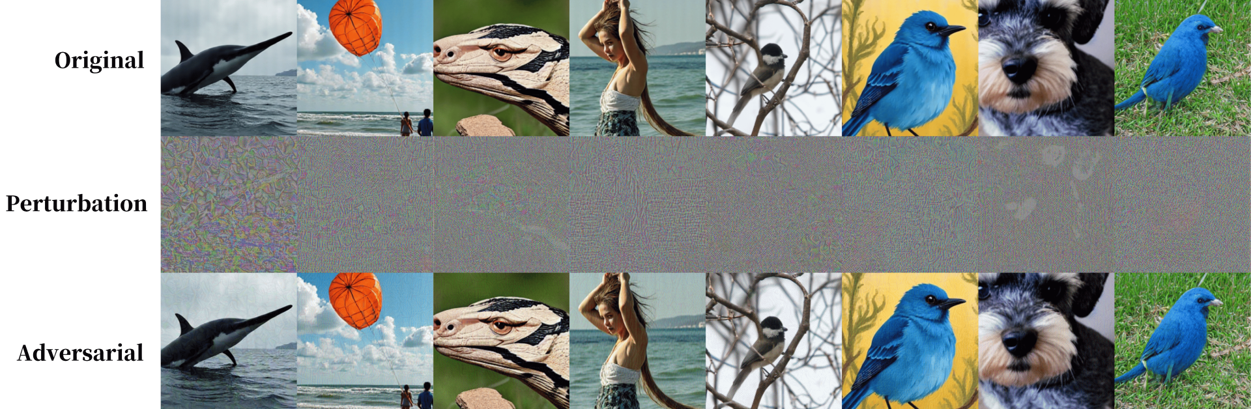}
        \caption{OOD Synthetic Images (White-box attacks on images from an unseen generative source)}
    \end{subfigure}
    
    \caption{\textbf{Visualizations of Adversarial Attacks against LaRE$^2$.} 
    The rows in each subfigure from top to bottom represent: Original Image, Adversarial Perturbation ($\times 15$ magnified), and the resulting Adversarial Example.}
    \label{fig:vis_lare}
\end{figure*}

\begin{figure*}[t]
    \centering
    % 1. Real Images
    \begin{subfigure}{\linewidth}
        \centering
        \includegraphics[width=\linewidth]{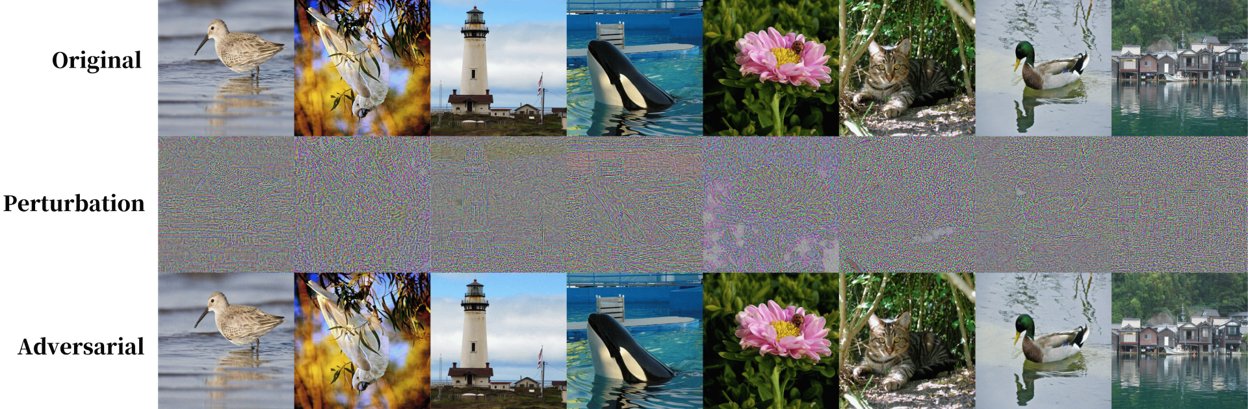}
        \caption{Real Images (ImageNet)}
    \end{subfigure}
    \par\medskip
    
    % 2. IID Synthetic Images
    \begin{subfigure}{\linewidth}
        \centering
        \includegraphics[width=\linewidth]{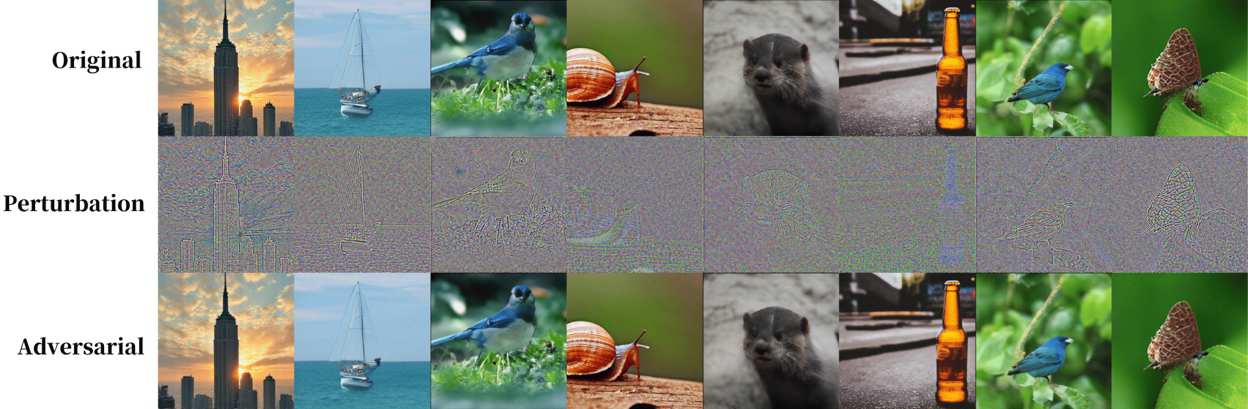}
        \caption{IID Synthetic Images (White-box attacks on images from the detector's known generative source)}
    \end{subfigure}
    \par\medskip
    
    % 3. OOD Synthetic Images
    \begin{subfigure}{\linewidth}
        \centering
        \includegraphics[width=\linewidth]{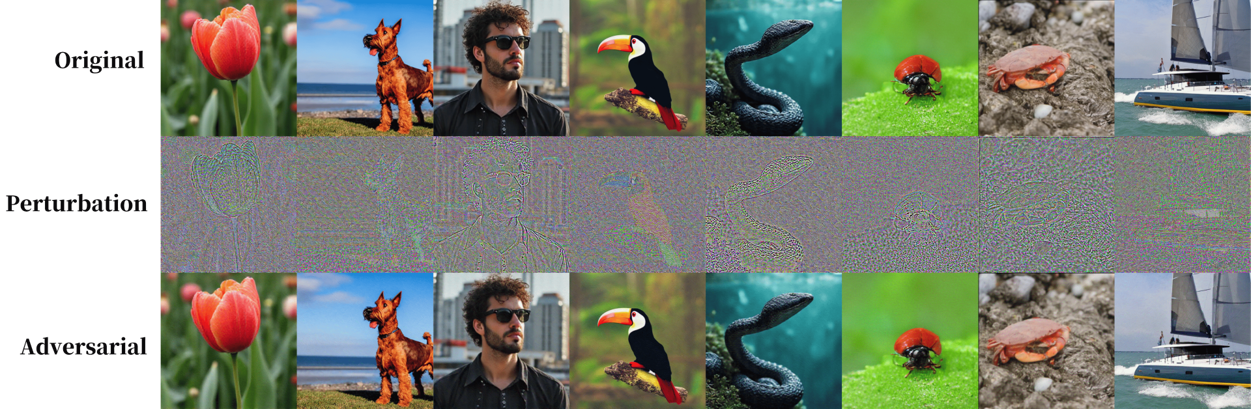}
        \caption{OOD Synthetic Images (White-box attacks on images from an unseen generative source)}
    \end{subfigure}
    
    \caption{\textbf{Visualizations of Adversarial Attacks against AEROBLADE.} 
    The rows in each subfigure from top to bottom represent: Original Image, Adversarial Perturbation ($\times 15$ magnified), and the resulting Adversarial Example.}
    \label{fig:vis_aeroblade}
\end{figure*}

We provide additional visualizations of original images, amplified adversarial perturbations, and the resulting adversarial examples for real images, as well as synthetic images under both IID and OOD settings, in Figures~\ref{fig:vis_dire}, \ref{fig:vis_lare}, and \ref{fig:vis_aeroblade}.

% ==========================================
% Figure 1: DIRE Visualization
% ==========================================

% ==========================================
% Figure 2: LaRE^2 Visualization
% ==========================================

% ==========================================
% Figure 3: AeroBlade Visualization
% ==========================================

\end{document}